\documentclass{article} 
\usepackage{iclr2027_conference,times}

\usepackage{amsmath,amsfonts,bm}

\def\eqref#1{equation~\ref{#1}}

\def\1{\bm{1}}

\DeclareMathAlphabet{\mathsfit}{\encodingdefault}{\sfdefault}{m}{sl}
\SetMathAlphabet{\mathsfit}{bold}{\encodingdefault}{\sfdefault}{bx}{n}

\usepackage{hyperref}
\usepackage{enumitem}
\usepackage{url}
\usepackage{comment}
\usepackage{booktabs}
\usepackage{multirow}
\usepackage{algorithm}
\usepackage{algorithmic}
\usepackage{graphicx}
\usepackage[capitalize]{cleveref}
\usepackage[dvipsnames,table]{xcolor}
\usepackage[section,above,below]{placeins}
\usepackage{float}
\usepackage{wrapfig}

\crefname{section}{Sec.}{Secs.}
\Crefname{section}{Section}{Sections}
\crefname{table}{Tab.}{Tabs.}
\Crefname{table}{Table}{Tables}

\title{Rethinking Object-Centric Representations for Video Dynamics}

\author{Amaury Wei, Ismail Nejjar, Olga Fink \\
Intelligent Maintenance and Operation Systems (IMOS) laboratory\\
Ecole Polytechnique Fédérale de Lausanne (EPFL)\\
Lausanne, CH-1015, Switzerland \\
\texttt{\{first.last\}@epfl.ch}
}

\iclrfinalcopy 
\begin{document}

\maketitle

\begin{figure}[h]
    \centering
    \vspace{-0.8cm}
    \includegraphics[width=1\linewidth]{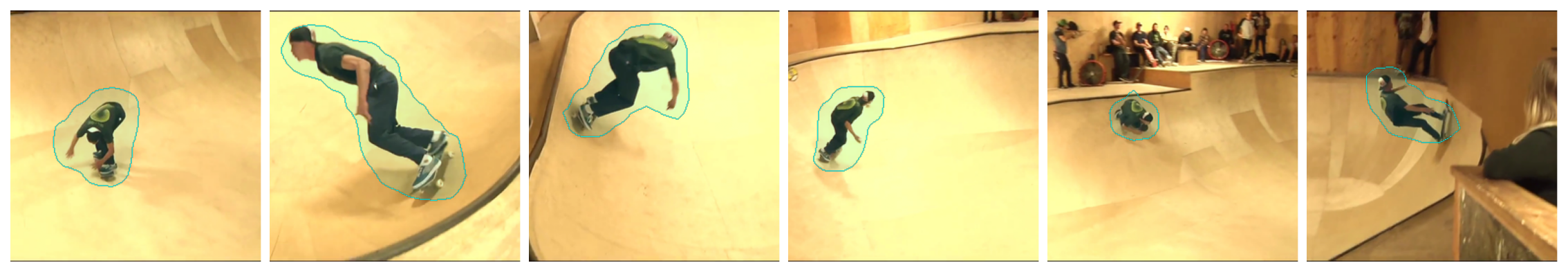}\\ \vspace{-5pt}
    \includegraphics[width=1\linewidth]{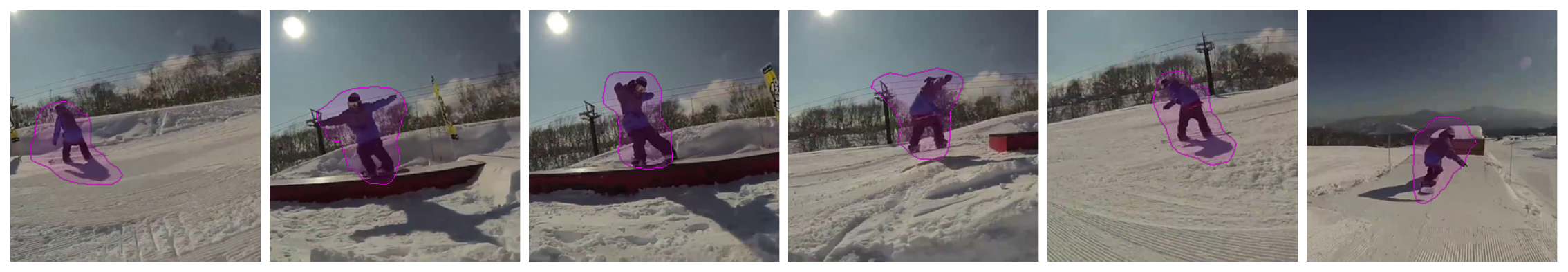}\\ \vspace{-5pt}
    \includegraphics[width=1\linewidth]{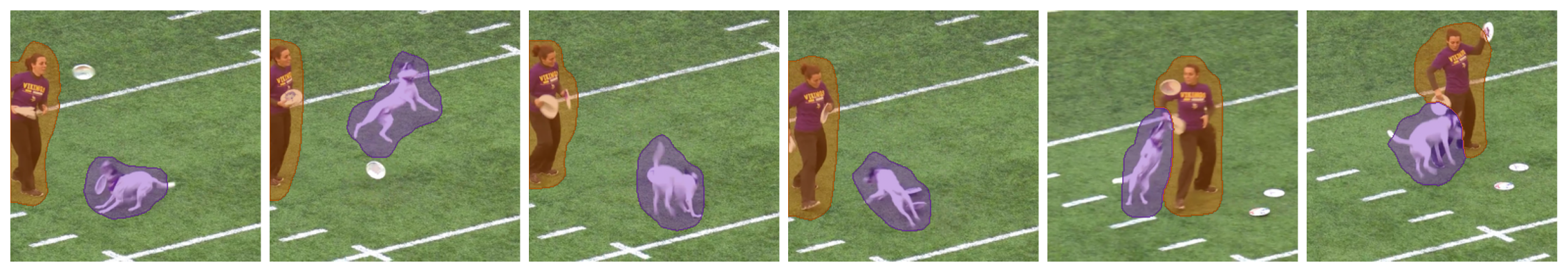}\\ \vspace{-10pt}
    \caption{UniSlot: Our method disentangles object appearance from spatial pose and \textcolor{black}{aligns appearance over time to learn robust object-centric representations for downstream tasks}.}
    \label{fig:visual_abstract}
\end{figure}

\begin{abstract}
Learning to decompose videos into persistent objects is a fundamental challenge in unsupervised object-centric representation learning. Despite recent progress, existing methods struggle to simultaneously achieve accurate object segmentation, consistent identities over time, and reliable foreground–background separation. To address these challenges, we introduce UniSlot (Unified Slots), an unsupervised framework for learning robust and disentangled object-centric representations from videos. UniSlot explicitly separates object appearance from its 3D-aware geometric pose in the scene, linking object identity to appearance while leveraging depth to better distinguish objects from their surroundings. UniSlot achieves state-of-the-art performance in unsupervised object-centric video decomposition and tracking across synthetic and real-world benchmarks, yielding substantially tighter object masks and reducing background leakage while preserving object identities. Beyond decomposition and tracking, these improved representations translate directly to downstream object dynamics prediction \textcolor{black}{and visual question answering}, enabling more accurate forecasting of future object trajectories \textcolor{black}{and reasoning about object interactions}. Code available at \linebreak\url{https://github.com/AmauryWEI/unislot}.
\end{abstract}

\section{Introduction} \label{sec:intro}

Object-centric representations provide a structured abstraction of visual scenes that has proven useful for video prediction and dynamics modeling \citep{slotgnn,ocvp,wu2023slotformer}, causal reasoning \citep{phycine,Xiao_2024_CVPR,chen2025compositional}, and planning \citep{playslot,zou2025tdgnet}. Videos offer particularly rich supervision for learning such representations without object-level annotations, as objects persist, move, and interact over time. A robust object-centric video representation should therefore accurately localize objects while preserving their identities as the scene evolves.

Despite substantial progress, learning such representations from videos remains challenging. Slot-based methods \citep{burgess2019monet,slot_attention} have established a powerful paradigm for unsupervised object discovery, with recent approaches leveraging self-supervised visual features \citep{dino,dinosaur} to extend object-centric learning to complex real-world scenes. In video, methods such as SAVi \citep{savi,savi++}, VideoSAUR \citep{videosaur}, and SlotContrast \citep{slot_contrast} further introduce temporal mechanisms to maintain object correspondence across frames. Yet existing methods still struggle to simultaneously produce tight object masks, preserve identities over time, and cleanly separate foreground objects from their surroundings. We argue that these failures originate from the \textit{entanglement of object appearance and geometry in standard slot representations}. Appearance and geometry inherently evolve differently over time: while an object’s position and scale change with motion, its appearance provides a more stable cue for correspondence. Enforcing temporal consistency on a representation that mixes these factors \citep{videosaur,slot_contrast} therefore constrains properties that should remain free to change. Furthermore, appearance cues alone can be insufficient to separate objects from complex backgrounds, particularly when foreground and background regions are visually similar. These limitations highlight the need to model appearance and geometry separately in object-centric video representations.

To this end, we propose UniSlot (UnifiedSlots), an unsupervised representation learning framework that explicitly disentangles object appearance from its 3D-aware geometric pose in the scene. UniSlot models appearance as a stable cue for temporal correspondence, while position and scale capture the object's evolving geometric state. This separation enables temporal consistency to be enforced on appearance while allowing geometry to evolve with object motion. UniSlot further leverages monocular depth as a spatial prior to capture the 3D structure of objects, better distinguish them from their surroundings, and reduce background leakage in complex scenes~(\cref{fig:visual_abstract}). Overall, UniSlot learns more stable object-centric representations, with tight localization and consistent identities.

We evaluate UniSlot on synthetic and real-world video benchmarks~\citep{kubric,Yang2019vis} where it achieves state-of-the-art performance in unsupervised object-centric video decomposition and tracking. Importantly, these improvements extend to downstream tasks such as unsupervised object dynamics prediction~\citep{wu2023slotformer} \textcolor{black}{and visual question answering}. Our main contributions are summarized as follows:\vspace{-4pt}

\begin{enumerate}[leftmargin=*]
    \item We identify the entanglement of appearance and geometry as a key limitation of existing object-centric video representations, motivating their separate modeling.\vspace{-2pt}
    \item We propose UniSlot, an unsupervised representation learning framework that disentangles appearance from 3D-aware geometry and leverages monocular depth pseudo-supervision for improved object decomposition and tracking.\vspace{-2pt}
    \item We demonstrate SOTA performance across synthetic and real-world video benchmarks, validate the contributions of individual components through ablations, \textcolor{black}{test disentanglement with interventions}, and show improved downstream object dynamics prediction \textcolor{black}{and visual question answering}.
\end{enumerate}

\section{Related work} \label{sec:related}

\paragraph{Object-centric representation learning}
Slot Attention (SA)~\citep{slot_attention} established a widely used paradigm for unsupervised object-centric learning, grouping visual features into a fixed set of latent representations (called ``slots'') that are decoded to reconstruct the input. Subsequent work has improved object-centric representations along two main directions. First, methods such as ISA~\citep{isa} and DISA~\citep{disa} explore disentangled representations to improve object decomposition, separating spatial attributes from appearance in ISA and texture from contour information in DISA. While effective, these methods primarily target individual images from synthetic scenes. The second line of work focuses on scaling object discovery to more visually complex scenes. More expressive decoders, including mixture and compositional architectures~\citep{slate,jiang2023objectcentric,slot_mixture_module}, increase the capacity of slots to reconstruct complex visual content, while methods such as DINOSAUR~\citep{dinosaur} leverage self-supervised features from pretrained vision models~\citep{dino} to enable object discovery in real-world images. However, these directions remain difficult to reconcile, as expressive decoders weaken the separation encouraged by structured representations. Learning disentangled object representations in real-world videos therefore remains an open challenge.

\paragraph{Temporal object-centric learning}
Extending object-centric learning to video requires maintaining object correspondence as objects move, interact, and become occluded. SAVi~\citep{savi} propagates slots recurrently across frames, using previous slot states to initialize subsequent inference, while SAVi++~\citep{savi++} additionally exploits sparse ground-truth depth obtained from LiDAR sensors. STEVE~\citep{steve} and SOLV~\citep{solv} extend object-centric learning to more complex and naturalistic videos, while VideoSAUR~\citep{videosaur} leverages self-supervised visual features and learns their temporal evolution to improve object discovery and tracking in real-world videos, whereas SlotContrast~\citep{slot_contrast} introduces a contrastive objective that promotes temporal consistency of slot representations. Recent methods improve video object decomposition through synergistic encoder--decoder learning (SRL)~\citep{seong2026from} and progressive slot allocation \textcolor{black}{(SlotCurri)} to reduce over-fragmentation~\citep{moon2026reconstruction}. Despite these advances, existing methods still struggle with accurate object segmentation in complex videos, exhibiting oversegmentation, background leakage, and identity switches over time.

\paragraph{Object-centric dynamics modeling}
Object-centric representations enable visual dynamics to be modeled at the level of individual entities rather than pixels. SlotGNN~\citep{slotgnn} models interactions between object slots using graph neural networks, while OCVP~\citep{ocvp} and SlotFormer~\citep{wu2023slotformer} predict future object representations from their temporal evolution. However, these methods rely on the quality and structure of the underlying object representations, motivating better separation between persistent and motion-based object properties.

\section{Method} \label{sec:method}

\begin{figure}[h]
    \centering
    \includegraphics[width=1\linewidth]{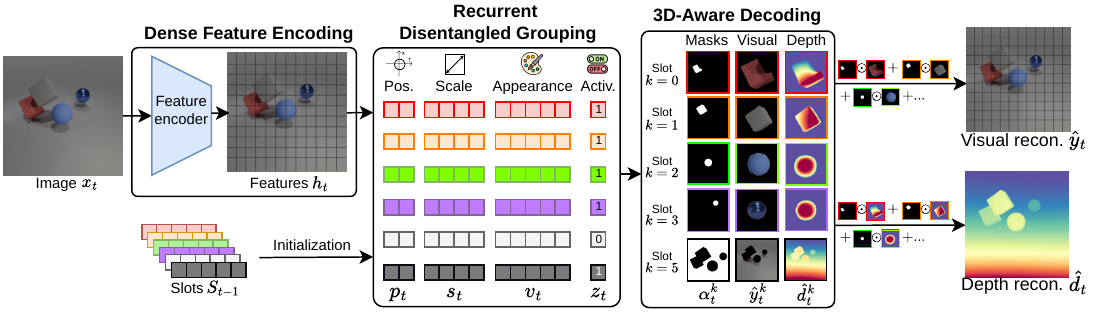}
    \vspace{-0.8cm}
    \caption{Overview of UniSlot. Given an RGB frame $x_t$, a visual encoder extracts dense features that are grouped into recurrent, disentangled object slots $S_t^k=(v_t^k,p_t^k,s_t^k)$ representing appearance, position, and scale, with $z_t^k$ controlling slot activation. Active slots are independently decoded into alpha masks, visual content, and depth, which are finally composited to reconstruct the visual and monocular depth targets.}
    \label{fig:method_overview}
\end{figure}

We propose \textbf{UniSlot}, an unsupervised object-centric representation learning framework for learning spatially precise and temporally consistent object representations from videos. As illustrated in \cref{fig:method_overview}, UniSlot represents each object through separate appearance and geometric components, propagates these representations recurrently across frames, and reconstructs both visual content and monocular depth. This structure allows temporal constraints to act on appearance while geometric attributes evolve with object motion, and uses monocular depth as an additional self-supervised training signal to improve object--background separation.

\subsection{UniSlot representation} \label{subsec:representation}

Given a video $\mathcal{V}=\{x_1,\ldots,x_T\}$, with
$x_t\in\mathbb{R}^{H\times W\times 3}$, our goal is to decompose each frame into $K$ object-centric representations
$S_t=\{S_t^1,\ldots,S_t^K\}$.
Rather than encoding appearance and spatial information within a single latent vector, UniSlot explicitly represents each slot as
\begin{equation}
    S_t^k=(v_t^k,p_t^k,s_t^k),
    \qquad
    v_t^k\in\mathbb{R}^{D},
    \quad
    p_t^k,s_t^k\in\mathbb{R}^{2},
    \label{eq:slot_representation}
\end{equation}
where $v_t^k$ describes object appearance, and $p_t^k$ and $s_t^k$ represent its 2D position and scale.
This separation allows appearance to provide a stable cue for correspondence across frames while geometry can evolve with object motion.

Each frame is first encoded into dense visual features using a frozen self-supervised DINO encoder~\citep{dino}, followed by a lightweight learned projection:
\begin{equation}
    h_t = F_{\mathrm{proj}}\!\left(F_{\mathrm{feat}}(x_t)\right).
\end{equation}
Architectural details of the encoder and feature projection are provided in Appendix~\ref{app:implementation}.

\subsection{Recurrent disentangled slot grouping} \label{subsec:method_recurrent_grouping}

UniSlot employs a recurrent, disentangled, and adaptive grouping mechanism for learning persistent object representations from videos. \textcolor{black}{Building on ISA~\citep{isa}, we leverage pose-normalized grouping to separate object appearance from position and scale. This factorization enables temporal alignment to be applied exclusively to appearance.} UniSlot further incorporates temporal propagation and adaptive slot activation~\citep{fan2024adaptive}. This allows slots to consistently track the same objects as their geometry changes and the number of active objects varies over time.

At each timestep, the previous slot states are propagated to initialize grouping in the current frame. Appearance is propagated through a learned stochastic predictor $G_{\mathrm{pred}}$, while position and scale are carried forward directly:
\begin{equation}
    \tilde S_t^k =
    \begin{cases}
        \bigl(G_{\mathrm{pred}}(v_{t-1}^k),\,p_{t-1}^k,\,s_{t-1}^k\bigr),
        & t > 1, \\[2pt]
        S_{\mathrm{init}}^k,
        & t = 1,
    \end{cases}
    \label{eq:slot_initialization}
\end{equation}
where $S_{\mathrm{init}}^k$ denotes a learned initial slot state. The predicted states $\tilde S_t=\{\tilde S_t^k\}_{k=1}^K$ are then refined by the grouping module, followed by adaptive slot activation:
\begin{align}
    \left\{(v_t^k,p_t^k,s_t^k)\right\}_{k=1}^{K}
    &=
    G_{\mathrm{slot}}(h_t,\tilde S_t), \\
    z_t^k
    &=
    G_{\mathrm{gate}}(v_t^k),
\end{align}
where $z_t^k\in\{0,1\}$ determines whether slot $k$ is active. Inactive slots are excluded from reconstruction, allowing the active set to adapt to the number of objects in the scene and reducing over-segmentation from redundant slots. Further details of the recurrent predictor $G_{\mathrm{pred}}$, pose-normalized grouping $G_{\mathrm{slot}}$, and adaptive gating $G_{\mathrm{gate}}$ are provided in Appendix~\ref{app:grouping}.

\subsection{3D-aware slot decoding} \label{subsec:method_decoding}

As UniSlot is trained without object-level annotations, we learn the object decomposition through reconstruction. Each active slot is decoded independently following the spatial broadcasting principle~\citep{watters2019spatial}, intentionally avoiding feature mixing between slots. The position and scale $(p_t^k,s_t^k)$ define a slot-specific coordinate system from which the decoder constructs positional embeddings, while $v_t^k$ provides the appearance representation.

UniSlot jointly reconstructs visual content and monocular depth. Depending on the visual complexity of the dataset, the reconstruction target $y_t$ is either the RGB frame $x_t$ or the dense DINO features $h_t$. For each active slot, the decoder predicts a visual component $\hat{y}_t^k$, a depth component $\hat{d}_t^k$, and an alpha mask $\alpha_t^k$. The final reconstructions are obtained by combining the active slots:
\begin{align}
    (\hat{y}_t^k,\hat{d}_t^k,\alpha_t^k)
    &= D_{\mathrm{spatial}}\!\left(v_t^k,p_t^k,s_t^k\right), \label{eq:decoder}\\
    \hat{y}_t
    &= \sum_{k=1}^{K} z_t^k\,\alpha_t^k\odot\hat{y}_t^k,
    \qquad
    \hat{d}_t
    = \sum_{k=1}^{K} z_t^k\,\alpha_t^k\odot\hat{d}_t^k.
    \label{eq:compositing}
\end{align}

The pseudo-depth target $d_t$ is generated from each RGB frame using a frozen monocular depth estimator, Depth Anything v3~\citep{depth_anything_v3}. Unlike sensor-based depth supervision~\citep{savi++}, it requires neither LiDAR measurements nor ground-truth 3D annotations. Visual content and depth are reconstructed using the same object-specific position and scale, allowing depth cues to inform the geometric decomposition and improve foreground--background separation. Depth is used only as a training signal; UniSlot requires only an RGB video at inference. Architectural details of $D_{\mathrm{spatial}}$ are provided in Appendix~\ref{app:decoding}.

\subsection{Learning objectives} \label{subsec:method_objectives}

UniSlot uses mean squared error for the visual and depth reconstruction objectives. We further guide the learning of disentangled slot representations with three regularizations: temporal alignment to maintain object--slot associations across frames, spatial separation to promote slot specialization, and sparsity to reduce redundant slots and over-segmentation. The main training objectives are:
\begin{equation}
    \mathcal{L}
    = \lambda_{\mathrm{recon}}\mathcal{L}_{\mathrm{recon}}
    + \lambda_{\mathrm{depth}}\mathcal{L}_{\mathrm{depth}}
    + \lambda_{\mathrm{time}}\mathcal{L}_{\mathrm{time}} \\
    + \lambda_{\mathrm{sep}}\mathcal{L}_{\mathrm{sep}}
    + \lambda_{\mathrm{spars}}\mathcal{L}_{\mathrm{spars}} .
\label{eq:full_objective}
\end{equation}

Unlike SlotContrast~\citep{slot_contrast}, disentanglement allows UniSlot to align only appearance $v_t^k$ across time, avoiding conflicts between temporal consistency and evolving object geometry:
\begin{equation}
    \mathcal{L}_{\mathrm{time}}
    =
    \frac{
        \sum_{t,k} z_{t-1}^k z_t^k
        [1-\cos(\tilde v_t^k,v_t^k)]
    }{
        \sum_{t,k} z_{t-1}^k z_t^k
    }.
    \label{eq:temporal_loss}
\end{equation}
Here, $\tilde v_t^k$ is the appearance predicted from the preceding frame. Finally, $\mathcal{L}_{\mathrm{sep}}$ penalizes positive cosine similarity between active slot appearances, while $\mathcal{L}_{\mathrm{spars}}$ penalizes unnecessary slot activation. Further details about the training objectives and their weights are provided in Appendix~\ref{app:objectives}.

\section{Experiments} \label{sec:experiments}

\subsection{Experimental setup} \label{subsec:experimental_setup}

\paragraph{Datasets}
We evaluate UniSlot across synthetic and real-world video benchmarks of increasing visual complexity. CLEVRER~\citep{clevrer} provides a controlled synthetic setting, while MOVi-C and MOVi-E~\citep{kubric} feature increasingly complex photorealistic scenes with diverse objects, backgrounds, and motion. For real-world evaluation, we use YouTube-VIS 2019 and 2021~\citep{Yang2019vis}. We additionally evaluate transfer of the learned representations on DAVIS 2017~\citep{davis2017}. More details about the datasets are given in Appendix~\ref{app:datasets}, and results on extra MOVi benchmarks are provided in Appendix~\ref{app:extra_quantitative}.\vspace{-4pt}

\paragraph{Metrics}
We report complementary metrics for object decomposition and tracking. Adjusted Rand Index (ARI)~\citep{rand1971objective,hubert1985comparing} evaluates the complete scene decomposition, including background, while FG-ARI isolates foreground object separation. Mean Best Overlap (mBO)~\citep{pont2016multiscale} measures object-mask overlap. $\mathcal{J}\&\mathcal{F}$~\citep{Perazzi_2016_CVPR} evaluates video segmentation and tracking quality through region and boundary accuracy.\vspace{-4pt}

\paragraph{Baselines}
We compare UniSlot against established and recent methods: SAVi~\citep{savi}, VideoSAUR~\citep{videosaur}, SlotContrast~\citep{slot_contrast}, SRL~\citep{seong2026from} and SlotCurri~\citep{moon2026reconstruction}, using their official implementations and matching feature encoders where applicable. For completeness, additional comparisons with single-image object-centric methods, including ISA~\citep{isa}, Adaptive Slot Attention~\citep{fan2024adaptive}, and DINOSAUR~\citep{dinosaur}, are provided in Appendix~\ref{app:extra_quantitative}.\vspace{-4pt}

\paragraph{Implementation}
UniSlot uses a frozen DINO ViT-B/16 feature encoder and a frozen Depth Anything v3~\citep{depth_anything_v3} model (\texttt{DA3MONO-LARGE}) to generate monocular pseudo-depth targets during training. Depth is not required at inference. Full architecture, training settings, and dataset-specific hyperparameters are provided in Appendix~\ref{app:implementation}.

\subsection{Object-centric video decomposition and tracking} \label{subsec:results_decomposition}

We first evaluate unsupervised object discovery and tracking using sequence-level metrics in \cref{tab:video_decomposition}. UniSlot achieves the best performance across all metrics and datasets. The improvement is particularly pronounced in terms of ARI, which evaluates the complete scene decomposition including background; on CLEVRER, UniSlot reaches $.927$ compared with $.095$ for the strongest baseline. Together with the highest FG-ARI, mBO, and $\mathcal{J}\&\mathcal{F}$, these results indicate that UniSlot improves foreground--background separation without sacrificing object localization or temporal consistency.

Importantly, these gains are not only achieved on synthetic scenes, but also on real-world videos. On YouTube-VIS 2019, UniSlot nearly doubles ARI over SlotContrast ($.304$ vs.\ $.153$), while also improving mBO from $.406$ to $.516$ and $\mathcal{J}\&\mathcal{F}$ from $.308$ to $.360$. The same trend holds on YouTube-VIS 2021, where UniSlot again outperforms all baselines across all metrics. Consistent improvements from controlled synthetic scenes to complex real-world videos show that UniSlot’s disentangled representation supports both foreground–background decomposition and temporal object consistency across substantially different scene conditions.

\begin{table}[t]
\vspace{-0.5cm}
\caption{Object-centric video decomposition and tracking. Metrics are computed
over video sequences. Best results are shown in bold and second-best are
underlined.}
\label{tab:video_decomposition}
\centering
\scriptsize
\vspace{2pt}
\setlength{\tabcolsep}{4.5pt}
\begin{tabular}{lcccccccccccc}
\toprule
\multirow{2}{*}{\textbf{Method}}
& \multicolumn{4}{c}{\textbf{CLEVRER}}
& \multicolumn{4}{c}{\textbf{MOVi-C}}
& \multicolumn{4}{c}{\textbf{MOVi-E}} \\
\cmidrule(lr){2-5}
\cmidrule(lr){6-9}
\cmidrule(lr){10-13}
& ARI$\uparrow$ & FG-ARI$\uparrow$ & mBO$\uparrow$ & J\&F$\uparrow$
& ARI$\uparrow$ & FG-ARI$\uparrow$ & mBO$\uparrow$ & J\&F$\uparrow$
& ARI$\uparrow$ & FG-ARI$\uparrow$ & mBO$\uparrow$ & J\&F$\uparrow$ \\
\midrule
SAVi
& .001 & .241 & .046 & .045
& .011 & .138 & .075 & .097
& .002 & .227 & .099 & .123 \\
VideoSAUR
& .015 & .210 & .028 & .014
& \underline{.220} & .614 & \underline{.334} & \underline{.218}
& \underline{.236} & .585 & .251 &  .173 \\
SlotContrast
& \underline{.095} & .618 & \underline{.118} & \underline{.067}
& .103 & .568 & .274 & .107
& .204 & .744 & \underline{.262} &  \underline{.214} \\
SRL
& .080 & .451 & .071 & .052
& .036 & .195 & .057 & .030
& .143 & \underline{.754} & .246 & .198 \\
SlotCurri
& .024 & \underline{.620} & .067 & .053
& .186 & \underline{.652} & .289 & .197
& .046 & .570 & .169 & .142 \\
\rowcolor{lightgray!30}
UniSlot (Ours)
& \textbf{.927} & \textbf{.781} & \textbf{.647} & \textbf{.558}
& \textbf{.505} & \textbf{.654} & \textbf{.358} & \textbf{.280}
& \textbf{.267} & \textbf{.785} & \textbf{.282} & \textbf{.230} \\
\bottomrule
\end{tabular}

\vspace{2pt}

\begin{tabular}{lcccccccc}
\toprule
\multirow{2}{*}{\textbf{Method}}
& \multicolumn{4}{c}{\textbf{YouTube-VIS 2019}}
& \multicolumn{4}{c}{\textbf{YouTube-VIS 2021}} \\
\cmidrule(lr){2-5}
\cmidrule(lr){6-9}
& ARI$\uparrow$ & FG-ARI$\uparrow$ & mBO$\uparrow$ & J\&F$\uparrow$
& ARI$\uparrow$ & FG-ARI$\uparrow$ & mBO$\uparrow$ & J\&F$\uparrow$ \\
\midrule
SAVi
& .042 & .122 & .271 & .187
& .034 & .109 & .181 & .162 \\
VideoSAUR
& .150 & .145 & .398 & .298
& .133 & .161 & .368 & .256 \\
SlotContrast
& .153 & \underline{.194} & .406 & \underline{.308}
& \underline{.134} & \underline{.206} & \underline{.371} & \underline{.275} \\
SRL
& \underline{.225} & .139 & \underline{.409} & .219
& .097 & .179 & .243 & .186 \\
SlotCurri
& .083 & .112 & .247 & .206
& .058 & .154 & .250 & .193 \\
\rowcolor{lightgray!30}
UniSlot (Ours)
& \textbf{.304} & \textbf{.225} & \textbf{.516} & \textbf{.360}
& \textbf{.228} & \textbf{.231} & \textbf{.455} & \textbf{.315} \\
\bottomrule
\end{tabular}
\vspace{-0.4cm}
\end{table}

\begin{figure}[h!]
    \centering
    \includegraphics[width=1\linewidth]{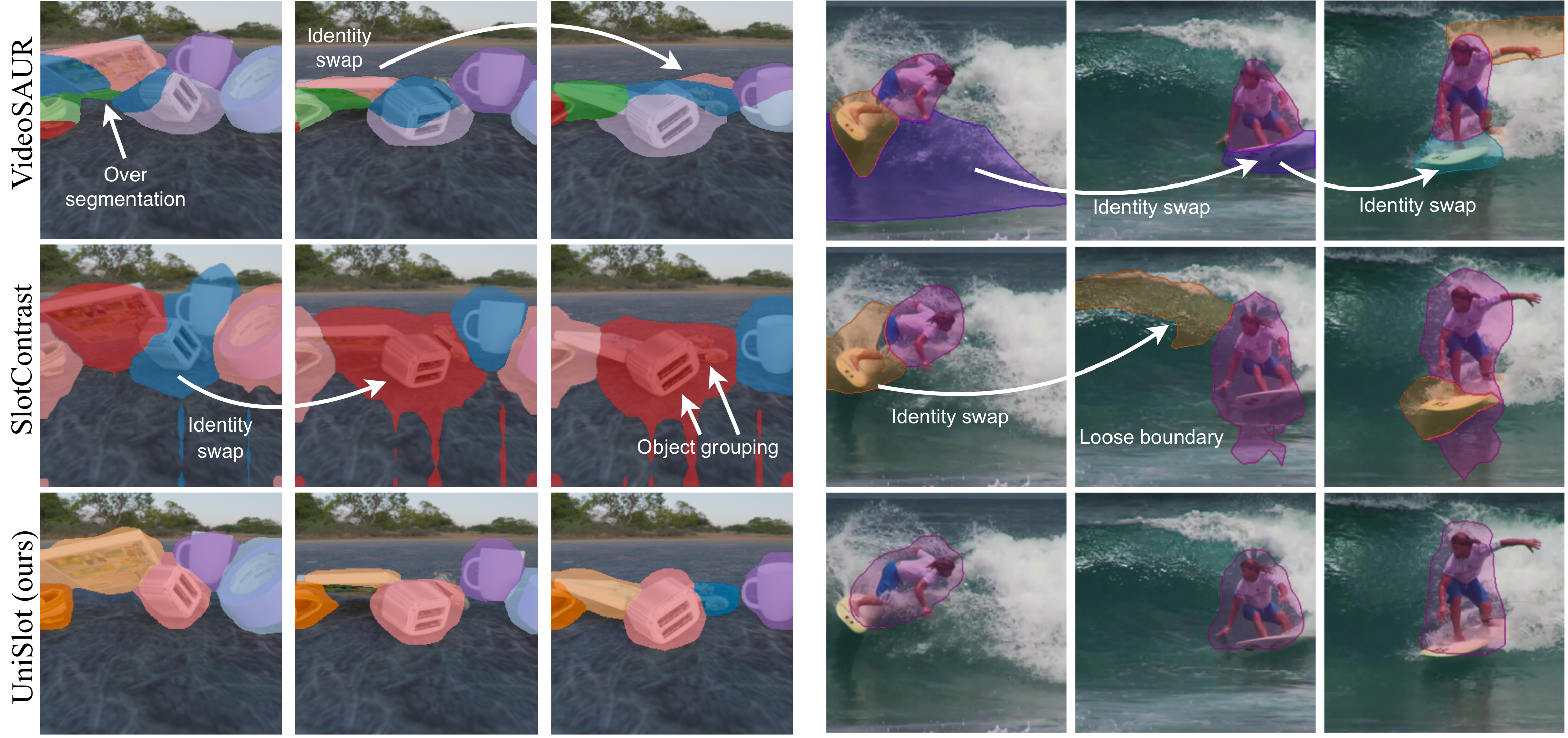}\\ \vspace{-0.3cm}
    \caption{Qualitative comparison of object decomposition and tracking. Colors denote fixed slot indices within each sequence. On MOVi-C (\emph{left}), VideoSAUR exhibits over-segmentation and identity swaps, while SlotContrast merges objects and changes their slot assignments. On YouTube-VIS (\emph{right}), both baselines exhibit identity switches and inaccurate boundaries. UniSlot maintains consistent object--slot associations and tightly localized masks across both settings.}
    \label{fig:qualitative_main}
    \vspace{-0.2cm}
\end{figure}

\vspace{-0.2cm}
\paragraph{Qualitative results}
\Cref{fig:qualitative_main} illustrates the failure modes behind the quantitative differences across methods. VideoSAUR frequently over-segments objects and changes their slot assignments over time, while SlotContrast can merge distinct objects and include large background regions in its foreground masks. These failures occur in both synthetic and real-world videos, degrading decomposition (ARI) as well as region and contour accuracy ($\mathcal{J}\&\mathcal{F}$). In contrast, UniSlot maintains more stable object assignments while producing tighter masks with substantially less background leakage.

\begin{figure}[t]
    \vspace{-0.7cm}
    \centering
    \includegraphics[width=1\linewidth]{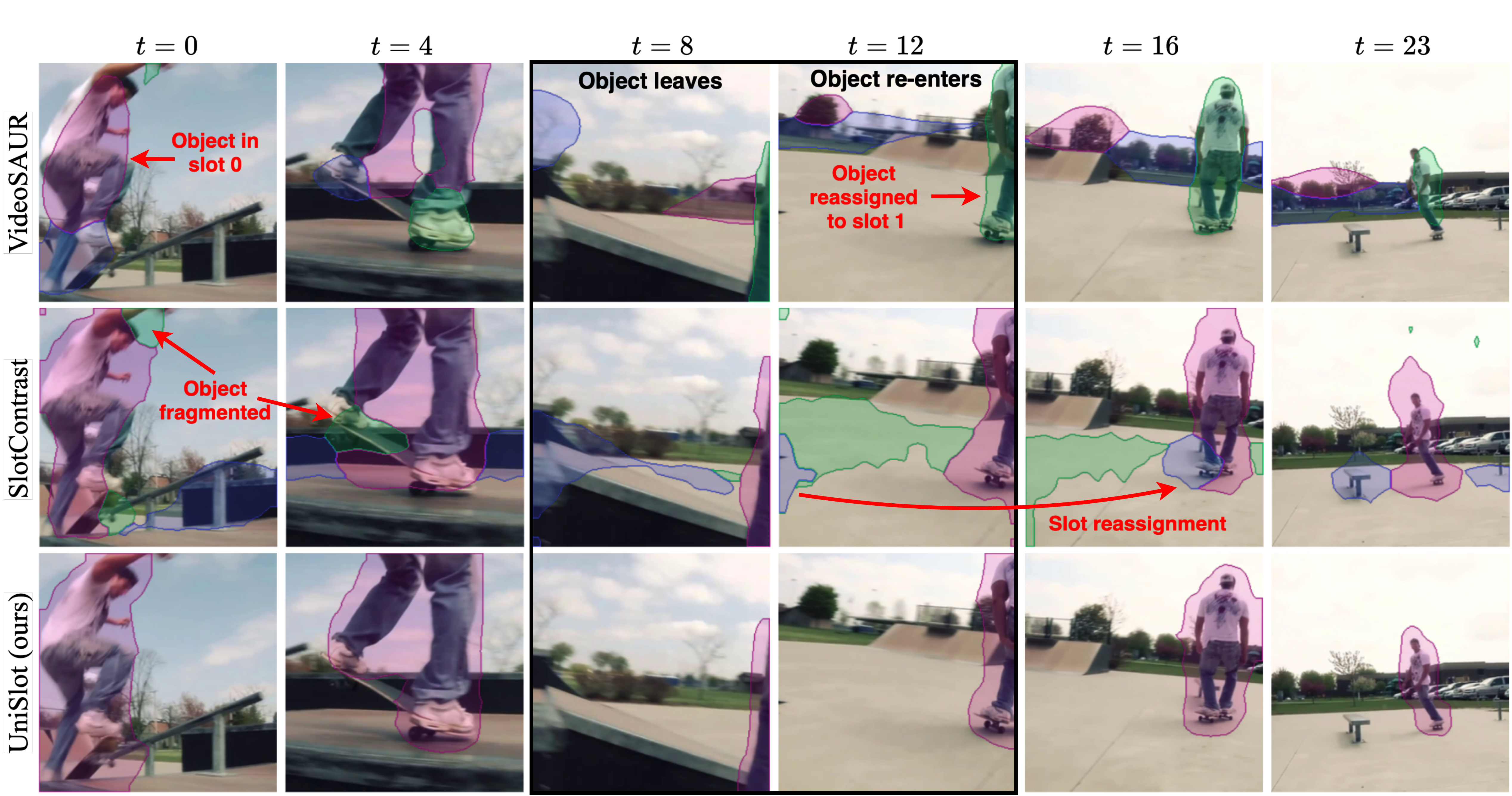}\\ \vspace{-0.3cm}
    \caption{Temporal correspondence under interrupted visibility. Colors denote fixed slot indices across the sequence. The target leaves at $t{=}8$ and re-enters at $t{=}12$. VideoSAUR assigns the returning person to a different slot, while SlotContrast exhibits fragmentation and slot reassignment. UniSlot preserves the object--slot association after re-entry while maintaining a tight localized mask.}
    \label{fig:qualitative_skater}
    \vspace{-0.4cm}
\end{figure}

\Cref{fig:qualitative_skater} examines a more challenging case involving temporary disappearance and re-entry. After the target person leaves the field of view at $t=8$ and re-enters at $t=12$, VideoSAUR assigns the person to a different slot, while SlotContrast exhibits fragmentation and reassigns slots. UniSlot instead recovers the target with the same slot after re-entry and maintains this association for the remainder of the sequence. This example shows that UniSlot's temporal consistency extends beyond short-term correspondence between adjacent frames. While \cref{fig:qualitative_main,fig:qualitative_skater} highlight specific examples, similar failure modes are observed across many sequences.

\subsection{Object dynamics prediction} \label{subsec:results_dynamics}

\textcolor{black}{To assess the utility of UniSlot's representations for downstream prediction, we evaluate object dynamics rollouts using SlotFormer~\citep{wu2023slotformer}}. We freeze each object-centric model and train a SlotFormer dynamics predictor on the resulting slot trajectories. For UniSlot, the predictor operates on its structured appearance, position, and scale representation, while for the baseline models it operates on their entangled slot embeddings. Predictor architecture and training details are provided in Appendix~\ref{app:object_dynamics_prediction}.

\begin{figure}[t]
    \vspace{-0.4cm}
    \centering
    \includegraphics[width=1\linewidth]{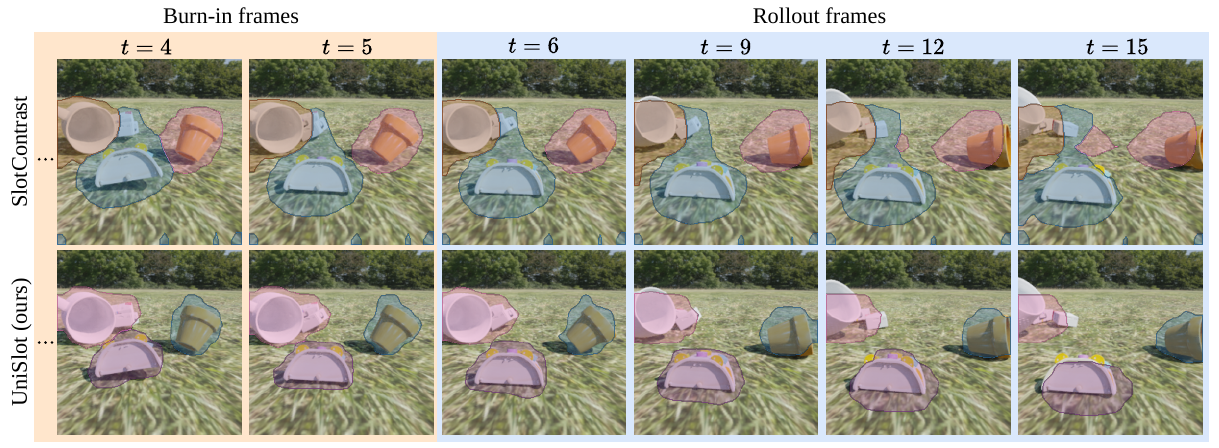}\\ \vspace{-0.4cm}
    \caption{Object dynamics prediction on MOVi-C. After six burn-in frames ($t\leq5$), future object states are predicted autoregressively. SlotContrast exhibits fragmentation and loose boundaries, while UniSlot maintains coherent object identities and localized masks throughout the rollout.}
    \label{fig:dynamics_prediction}
    \vspace{-0.2cm}
\end{figure}

As shown in \cref{tab:dynamics_results}, UniSlot improves downstream object dynamics prediction across both synthetic and real-world videos, achieving the best performance on 11 of 12 dataset--metric combinations. The improvement is particularly pronounced on MOVi-C, where ARI increases from $.123$ to $.318$ over the strongest baseline. These results show that UniSlot's \textcolor{black}{representations improve downstream} prediction of future object dynamics, \textcolor{black}{with the mask-based metrics reflecting both decomposition quality and predicted spatial evolution}. \Cref{fig:dynamics_prediction} illustrates this behavior qualitatively, with UniSlot preserving object identities and localized masks throughout the rollout, while SlotContrast exhibits fragmentation and less accurate predictions.

\begin{table}[h]
\vspace{-0.4cm}
\caption{Object dynamics prediction using frozen object-centric representations.
Best results are shown in bold and second-best are underlined.}
\label{tab:dynamics_results}
\centering
\scriptsize
\vspace{2pt}
\setlength{\tabcolsep}{4.5pt}
\begin{tabular}{lcccccccccccc}
\toprule
\multirow{2}{*}{\textbf{Method}}
& \multicolumn{4}{c}{\textbf{MOVi-C}}
& \multicolumn{4}{c}{\textbf{MOVi-E}}
& \multicolumn{4}{c}{\textbf{YouTube-VIS 2021}} \\
\cmidrule(lr){2-5}
\cmidrule(lr){6-9}
\cmidrule(lr){10-13}
& ARI$\uparrow$ & FG-ARI$\uparrow$ & mBO$\uparrow$ & J\&F$\uparrow$
& ARI$\uparrow$ & FG-ARI$\uparrow$ & mBO$\uparrow$ & J\&F$\uparrow$
& ARI$\uparrow$ & FG-ARI$\uparrow$ & mBO$\uparrow$ & J\&F$\uparrow$ \\
\midrule
SAVi
& .008 & .055 & .034 & .023
& .001 & .125 & .086 & .071
& .024 & .092 & .142 & .118 \\
VideoSAUR
& \underline{.123} & \underline{.402} & \underline{.214} & \underline{.161}
& .157 & .560 & .234 & .202
& \underline{.121} & .161 & \underline{.359} & \underline{.287} \\
SlotContrast
& .055 & .378 & .133 & .097
& .166 & \textbf{.692} & .249 & .214
& .118 & \underline{.192} & .346 & .225 \\
\rowcolor{lightgray!30}
UniSlot (Ours)
& \textbf{.318} & \textbf{.504} & \textbf{.229} & \textbf{.175}
& \textbf{.202} & \underline{.636} & \textbf{.279} & \textbf{.242}
& \textbf{.194} & \textbf{.213} & \textbf{.389} & \textbf{.297} \\
\bottomrule
\end{tabular}
\vspace{-0.2cm}
\end{table}

\subsection{Transfer to DAVIS 2017} \label{subsec:results_transfer}

\begin{wraptable}[5]{r}{0.48\textwidth}
\vspace{-1.55cm}
\caption{Transfer to DAVIS 2017. All models are trained on YouTube-VIS 2021.}
\label{tab:davis_transfer}
\centering
\scriptsize
\vspace{2pt}
\begin{tabular}{lcccc}
\toprule
\textbf{Method}
& ARI$\uparrow$
& FG-ARI$\uparrow$
& mBO$\uparrow$
& $\mathcal{J}\&\mathcal{F}\uparrow$ \\
\midrule
SAVi         & .018 & .067 & .183 & .173 \\
VideoSAUR    & .097 & \underline{.111} & .328 & .213 \\
SlotContrast & \underline{.158} & .085 & \underline{.330} & \underline{.234} \\
\rowcolor{lightgray!30}
UniSlot (Ours) & \textbf{.179} & \textbf{.148} & \textbf{.379} & \textbf{.242} \\
\bottomrule
\end{tabular}
\end{wraptable}

We evaluate transfer to unseen real-world videos by applying models trained on YouTube-VIS 2021 directly to DAVIS 2017~\citep{davis2017}, without retraining or fine-tuning. DAVIS emphasizes challenging temporal conditions, such as occlusion and fast object motion. As shown in \cref{tab:davis_transfer}, UniSlot achieves the best performance across all metrics, improving over the best baseline for each metric from $.158$ to $.179$ ARI, $.111$ to $.148$ FG-ARI, $.330$ to $.379$ mBO, and $.234$ to $.242$ $\mathcal{J}\&\mathcal{F}$. These results show that the learned object representations transfer to unseen real-world videos without target-domain adaptation.

{\color{black}
\subsection{Visual question answering}\label{sec:vqa}

Object-centric representations can also support reasoning about objects and their interactions. We evaluate this on CLEVRER visual question answering~\citep{clevrer}, covering descriptive, explanatory, predictive, and counterfactual questions about object properties, collisions, and their causes. Following SlotFormer~\citep{wu2023slotformer}, an Aloe Transformer~\citep{ding2021attention} reasons jointly over the frozen slots of each method and the question tokens, and predictive questions use future slots rolled out by each method's SlotFormer (\cref{subsec:results_dynamics}). All methods share the same VQA architecture, hyperparameters, and training budget; only the input representations differ. Details are provided in Appendix~\ref{app:vqa}.
}

\begin{wraptable}[9]{r}{0.60\textwidth}
    \vspace{-0.78cm}
    \centering
    \caption{\textcolor{black}{CLEVRER visual question answering with frozen slots. Per-question accuracy (\%) on the validation set. Predictive questions use slots rolled out by each method's SlotFormer. \emph{Question only} receives no visual input.}}
    \label{tab:vqa}
    \scriptsize
    \begin{tabular}{lcccc}
    \toprule
    Method & Descriptive & Explanatory & Predictive & Counterfactual \\
    \midrule
    Question only & 34.54 & 13.42 & 24.15 & 10.04 \\
    VideoSAUR     & 47.68 & 20.92 & 25.61 & 11.69 \\
    SlotContrast  & \underline{82.66} & \underline{77.27} & \underline{69.64} & \underline{47.01} \\
    \rowcolor{gray!15}
    UniSlot (Ours) & \textbf{89.75} & \textbf{89.76} & \textbf{83.19} & \textbf{67.85} \\
    \bottomrule
    \end{tabular}
\end{wraptable}

\textcolor{black}{As shown in \cref{tab:vqa}, UniSlot achieves the highest accuracy across all four question types, with particularly large gains over SlotContrast on counterfactual ($+20.84$ pp) and predictive ($+13.55$ pp) questions. The latter use slots rolled out by each method's dynamics model, demonstrating that UniSlot's predicted representations are more useful for reasoning about future events. The particularly large improvement on counterfactual questions further demonstrates the utility of UniSlot’s representations for reasoning about hypothetical object interactions.
}

\begin{figure}[t]
    \vspace{-0.4cm}
    \centering
    \includegraphics[width=\linewidth]{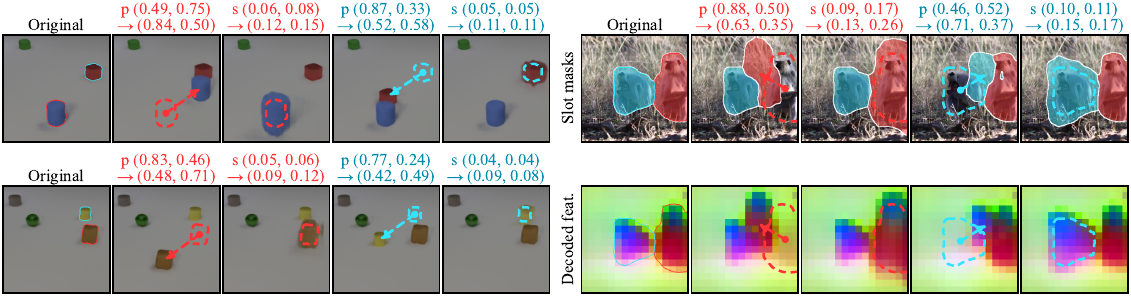}
    \vspace{-0.7cm}
    \caption{\textcolor{black}{Slot interventions on CLEVRER (left) and YT-VIS (right). We modify the position $p$ or scale $s$ of one slot while keeping all other latent components fixed. Object~1 is shown in red and object~2 in cyan. Titles indicate values before $\rightarrow$ after; dashed contours and arrows mark the original extent and displacement. For YT-VIS, we visualize decoded slot masks over the original frame and PCA-projected decoded DINO features.}}
    \label{fig:disentanglement}
    \vspace{-0.4cm}
\end{figure}

{\color{black}
\subsection{Disentanglement of appearance and geometry} \label{sec:disentanglement}

We further examine the separation of appearance and geometry through controlled interventions on the position $p_t^k$ and scale $s_t^k$. Since each slot is decoded in its own coordinate frame defined by $(p_t^k,s_t^k)$ (\cref{subsec:method_decoding}), we can intervene on these components in a trained model and decode without retraining. \Cref{fig:disentanglement} shows interventions on the position or scale of one slot while keeping its appearance and all other slots fixed. On CLEVRER (left), changing $p$ moves the object, while changing $s$ enlarges it, preserving its color, shape, and material without affecting other objects. On YT-VIS (right), the decoder reconstructs DINO features rather than pixels, so the input frame remains unchanged. Instead, the edits move or enlarge the decoded slot masks and features, while the other object's features remain in place. These interventions provide direct evidence of appearance–geometry disentanglement in the learned representations: position and scale can be manipulated independently while largely preserving object appearance and the remaining scene. Additional examples are provided in \cref{app:interventions}.}

\subsection{Ablation studies} \label{subsec:results_ablation}

We ablate the principal components of UniSlot to determine their contributions to object decomposition and temporal consistency. We evaluate the performance on two complementary benchmarks: MOVi-C provides photorealistic multi-object scenes with object motion and occlusions, while YouTube-VIS 2021 captures complex real-world videos with camera motion.

\begin{figure}[t]
    \centering
    \includegraphics[width=1\linewidth]{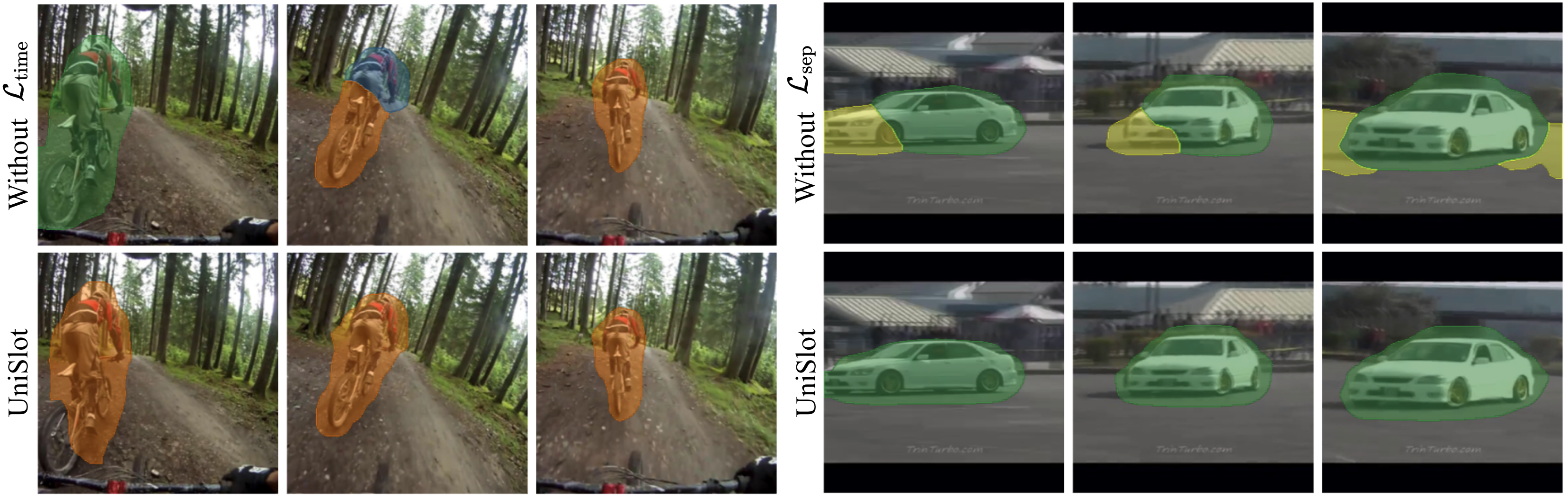}\\ \vspace{-0.2cm}
    \caption{Qualitative effect of UniSlot regularization losses. Without temporal alignment (\emph{left}), the same object switches identities across frames. Without slot separation (\emph{right}), an object can be decomposed in multiple redundant slots.}
    \label{fig:ablation_losses}
    \vspace{-0.2cm}
\end{figure}

\Cref{tab:ablation} shows that the components of UniSlot play complementary roles. Replacing \textcolor{black}{ISA-style pose-normalized grouping} with standard Slot Attention (SA) substantially degrades performance, \textcolor{black}{highlighting the importance of separating appearance and geometry}. \textcolor{black}{However, pose-normalized grouping alone does not account for UniSlot’s improvements: removing $\mathcal{L}_{\mathrm{time}}$ while retaining this grouping also substantially reduces performance (\textit{e.g.} FG-ARI on MOVi-C). Moreover, applying temporal alignment to all slot components $[v,p,s]$, rather than appearance alone, degrades performance, confirming the importance of allowing geometry to evolve with motion.} \Cref{fig:ablation_losses} illustrates the distinct roles of the two losses: without $\mathcal{L}_{\mathrm{time}}$, object identities become inconsistent across frames, while without $\mathcal{L}_{\mathrm{sep}}$, distinct objects can be assigned to the same slot. These results support treating temporal correspondence and separation between object representations as complementary constraints.

\begin{table}[h]
\vspace{-0.4cm}
\caption{Ablation of the principal UniSlot components on MOVi-C and
YouTube-VIS 2021.}
\label{tab:ablation}
\centering
\scriptsize
\vspace{2pt}
\begin{tabular}{lcccccccc}
\toprule
\multirow{2}{*}{\textbf{Variant}}
& \multicolumn{4}{c}{\textbf{MOVi-C}}
& \multicolumn{4}{c}{\textbf{YouTube-VIS 2021}} \\
\cmidrule(lr){2-5}\cmidrule(lr){6-9}
& ARI$\uparrow$ & FG-ARI$\uparrow$ & mBO$\uparrow$ & J\&F$\uparrow$
& ARI$\uparrow$ & FG-ARI$\uparrow$ & mBO$\uparrow$ & J\&F$\uparrow$ \\
\midrule
\textcolor{black}{$\mathcal{L}_{\mathrm{time}}$ on $[v,p,s]$}
    & \textcolor{black}{.205} & \textcolor{black}{.235} & \textcolor{black}{.211} & \textcolor{black}{.220}
    & \textcolor{black}{.075} & \textcolor{black}{.144} & \textcolor{black}{.253} & \textcolor{black}{.199} \\
w/o $\mathcal{L}_{\mathrm{time}}$
    & .266 & .204 & .205 & .204
    & .202 & .213 & .439 & .299 \\
w/o $\mathcal{L}_{\mathrm{sep}}$
    & .283 & .225 & .214 & .213
    & .211 & \underline{.218} & .443 & \underline{.303} \\
w/o adaptive activation
    & .476 & .237 & .258 & \underline{.251}
    & .136 & .203 & .378 & .283 \\
w/o disentanglement (SA)
    & .111 & .233 & .161 & .164
    & .165 & .181 & .411 & .296 \\
w/o depth pseudo-supervision
    & \underline{.477} & \underline{.646} & \underline{.348} & .195
    & \underline{.219} & .217 & \underline{.444} & .297 \\
\rowcolor{lightgray!30}
UniSlot
    & \textbf{.505} & \textbf{.654} & \textbf{.358} & \textbf{.280}
    & \textbf{.228} & \textbf{.231} & \textbf{.455} & \textbf{.315} \\
\bottomrule
\end{tabular}
\end{table}

\begin{figure}[t]
    \vspace{-0.2cm}
    \centering
    \includegraphics[width=1\linewidth]{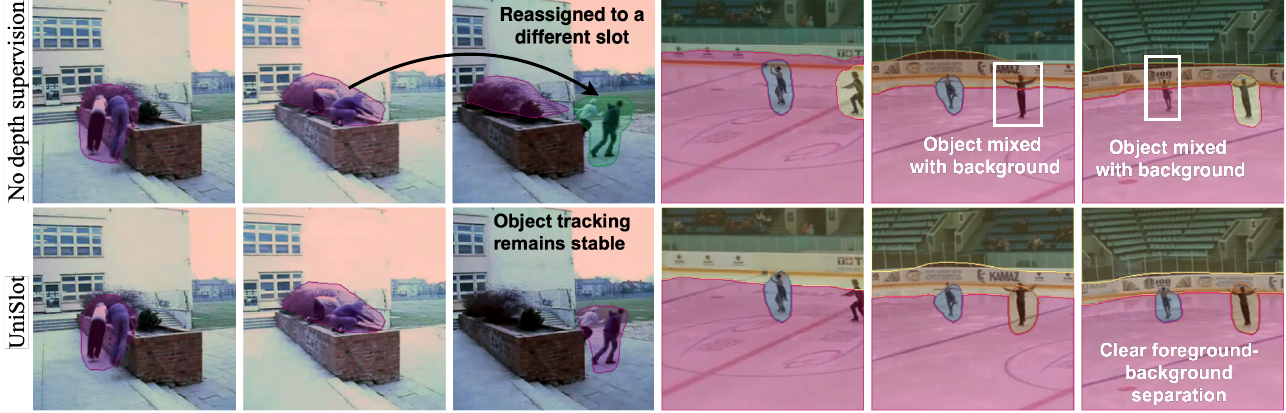}\\ \vspace{-0.3cm}
    \caption{Effect of monocular depth pseudo-supervision. Without depth, object assignments rely only on visual features, sometimes leading to unstable associations after interactions (\emph{left}) and foreground-background mixing (\emph{right}).}
    \label{fig:ablation_depth}
\end{figure}

Adaptive activation is particularly important on YouTube-VIS, where the number and composition of visible objects vary substantially; removing it reduces ARI from $.228$ to $.136$ and mBO from $.455$ to $.378$. Monocular depth pseudo-supervision provides consistent improvements across both datasets, including a substantial gain in $\mathcal{J}\&\mathcal{F}$ on MOVi-C ($.195$ to $.280$). As illustrated in \cref{fig:ablation_depth}, depth provides complementary geometric information that improves foreground--background separation and object association. Overall, the results show that disentanglement, regularization, adaptive activation, and depth play complementary roles, with the complete model performing best.

Additional robustness studies across feature encoders, slot configurations, and decoder architectures are provided in Appendix~\ref{app:additional_ablations}.

\section{Discussion} \label{sec:discussion}

Our results support the central assumption of UniSlot: appearance and geometry play different roles in object-centric video representations and benefit from being modeled separately. Applying temporal constraints to appearance while allowing position and scale to evolve enables stable object representations without suppressing motion. The downstream dynamics \textcolor{black}{and visual question answering} results further show that the learned representation is useful beyond video decomposition, \textcolor{black}{supporting both the prediction of future object states and reasoning about object interactions}. At the same time, our ablations show that disentanglement alone is not sufficient: temporal alignment, slot separation, adaptive activation, and monocular depth provide complementary mechanisms for learning robust object representations. \textcolor{black}{Our depth sensitivity analysis confirms the benefits of reliable depth supervision, while showing that inaccurate depth can be detrimental and that UniSlot remains effective without it.}

UniSlot relies on pretrained visual and monocular depth models to provide appearance and geometric cues during training. While these priors enable learning from videos without object or depth annotations, inaccuracies or domain biases in the pretrained models may propagate to the learned representations. Reducing this dependence, particularly for domains that differ substantially from the pretraining data, remains an interesting direction for future work.

\section{Conclusion} \label{sec:conclusion}

We propose UniSlot, an unsupervised framework for learning robust object-centric representations from videos by disentangling appearance from geometric states. By introducing temporal consistency to appearance while allowing position and scale to evolve, and complementing visual features with monocular depth, UniSlot improves object decomposition and temporal consistency across synthetic and real-world videos. The resulting representations provide effective object states for downstream dynamics prediction \textcolor{black}{and visual question answering, while} also generalizing to unseen videos.

Future work could explore higher-level semantic or language-based cues to maintain persistent object identities over longer videos, particularly under prolonged occlusions and substantial appearance changes.

\newpage
\section*{AI use statement}

In this work, we used generative AI tools to assist with drafting and polishing parts of the manuscript and with writing parts of the software code used to generate the figures. All AI-assisted text was reviewed and edited by the authors, and AI-assisted code was manually reviewed and tested for correctness. We did not use generative AI to generate synthetic datasets or prove mathematical claims. We take responsibility for the final content of this work, including text, code, claims, and artifacts produced with the aid of generative AI.

\section*{Ethics statement}

UniSlot studies unsupervised object-centric representation learning from video using standard synthetic and publicly available real-world benchmarks. While video representation and tracking technologies may be misused for surveillance, our work focuses on generic object representations and does not perform biometric identification or person recognition. We encourage responsible use with appropriate consideration of privacy and consent.

\section*{Reproducibility statement}

All datasets used in this work are publicly available. We provide detailed architectural configurations, training procedures, hyperparameters, and implementation details in the Appendix. The complete source code will be made publicly available upon acceptance.

\section*{Acknowledgments}

This work was supported by the Swiss National Science Foundation (SNSF) Grant Number 200021\_200461.

\bibliographystyle{iclr2027_conference}
\bibliography{iclr2027_conference}

\begin{thebibliography}{43}
\providecommand{\natexlab}[1]{#1}
\providecommand{\url}[1]{\texttt{#1}}
\expandafter\ifx\csname urlstyle\endcsname\relax
  \providecommand{\doi}[1]{doi: #1}\else
  \providecommand{\doi}{doi: \begingroup \urlstyle{rm}\Url}\fi

\bibitem[Aydemir et~al.(2023)Aydemir, Xie, and G\"uney]{solv}
G\"orkay Aydemir, Weidi Xie, and Fatma G\"uney.
\newblock Self-supervised object-centric learning for videos.
\newblock In \emph{NeurIPS}, volume~36, pp.\  32879--32899, 2023.

\bibitem[Biza et~al.(2023)Biza, Van~Steenkiste, Sajjadi, Elsayed, Mahendran, and Kipf]{isa}
Ondrej Biza, Sjoerd Van~Steenkiste, Mehdi S.~M. Sajjadi, Gamaleldin~F. Elsayed, Aravindh Mahendran, and Thomas Kipf.
\newblock Invariant slot attention: Object discovery with slot-centric reference frames.
\newblock In \emph{ICML}, pp.\  2507--2527, 2023.

\bibitem[Burgess et~al.(2019)Burgess, Matthey, Watters, Kabra, Higgins, Botvinick, and Lerchner]{burgess2019monet}
Christopher~P Burgess, Loic Matthey, Nicholas Watters, Rishabh Kabra, Irina Higgins, Matt Botvinick, and Alexander Lerchner.
\newblock {MONet}: Unsupervised scene decomposition and representation.
\newblock \emph{arXiv preprint arXiv:1901.11390}, 2019.

\bibitem[Caron et~al.(2021)Caron, Touvron, Misra, J\'egou, Mairal, Bojanowski, and Joulin]{dino}
Mathilde Caron, Hugo Touvron, Ishan Misra, Herv\'e J\'egou, Julien Mairal, Piotr Bojanowski, and Armand Joulin.
\newblock Emerging properties in self-supervised vision transformers.
\newblock In \emph{ICCV}, pp.\  9650--9660, October 2021.

\bibitem[Chen et~al.(2025)Chen, Dong, Yi, Li, Ding, Torralba, Tenenbaum, and Gan]{chen2025compositional}
Zhenfang Chen, Shilong Dong, Kexin Yi, Yunzhu Li, Mingyu Ding, Antonio Torralba, Joshua~B. Tenenbaum, and Chuang Gan.
\newblock Compositional physical reasoning of objects and events from videos.
\newblock \emph{IEEE TPAMI}, 47\penalty0 (9):\penalty0 7689--7703, 2025.

\bibitem[Darcet et~al.(2024)Darcet, Oquab, Mairal, and Bojanowski]{darcet2024vision}
Timoth{\'e}e Darcet, Maxime Oquab, Julien Mairal, and Piotr Bojanowski.
\newblock Vision transformers need registers.
\newblock In \emph{ICLR}, 2024.

\bibitem[Ding et~al.(2021)Ding, Hill, Santoro, Reynolds, and Botvinick]{ding2021attention}
David Ding, Felix Hill, Adam Santoro, Malcolm Reynolds, and Matt Botvinick.
\newblock Attention over learned object embeddings enables complex visual reasoning.
\newblock \emph{Advances in neural information processing systems}, 34:\penalty0 9112--9124, 2021.

\bibitem[Elsayed et~al.(2022)Elsayed, Mahendran, van Steenkiste, Greff, Mozer, and Kipf]{savi++}
Gamaleldin~Fathy Elsayed, Aravindh Mahendran, Sjoerd van Steenkiste, Klaus Greff, Michael~Curtis Mozer, and Thomas Kipf.
\newblock {SAV}i++: Towards end-to-end object-centric learning from real-world videos.
\newblock In \emph{NeurIPS}, volume~35, pp.\  28940--28954, 2022.

\bibitem[Fan et~al.(2024)Fan, Bai, Xiao, He, Horn, Fu, Locatello, and Zhang]{fan2024adaptive}
Ke~Fan, Zechen Bai, Tianjun Xiao, Tong He, Max Horn, Yanwei Fu, Francesco Locatello, and Zheng Zhang.
\newblock Adaptive slot attention: Object discovery with dynamic slot number.
\newblock In \emph{CVPR}, pp.\  23062--23071, 2024.

\bibitem[Greff et~al.(2022)Greff, Belletti, Beyer, Doersch, Du, Duckworth, Fleet, Gnanapragasam, Golemo, Herrmann, Kipf, Kundu, Lagun, Laradji, Liu, Meyer, Miao, Nowrouzezahrai, Oztireli, Pot, Radwan, Rebain, Sabour, Sajjadi, Sela, Sitzmann, Stone, Sun, Vora, Wang, Wu, Yi, Zhong, and Tagliasacchi]{kubric}
Klaus Greff, Francois Belletti, Lucas Beyer, Carl Doersch, Yilun Du, Daniel Duckworth, David~J. Fleet, Dan Gnanapragasam, Florian Golemo, Charles Herrmann, Thomas Kipf, Abhijit Kundu, Dmitry Lagun, Issam Laradji, Hsueh-Ti~(Derek) Liu, Henning Meyer, Yishu Miao, Derek Nowrouzezahrai, Cengiz Oztireli, Etienne Pot, Noha Radwan, Daniel Rebain, Sara Sabour, Mehdi S.~M. Sajjadi, Matan Sela, Vincent Sitzmann, Austin Stone, Deqing Sun, Suhani Vora, Ziyu Wang, Tianhao Wu, Kwang~Moo Yi, Fangcheng Zhong, and Andrea Tagliasacchi.
\newblock Kubric: A scalable dataset generator.
\newblock In \emph{CVPR}, pp.\  3749--3761, 2022.

\bibitem[Hubert \& Arabie(1985)Hubert and Arabie]{hubert1985comparing}
Lawrence Hubert and Phipps Arabie.
\newblock Comparing partitions.
\newblock \emph{Journal of classification}, 2\penalty0 (1):\penalty0 193--218, 1985.

\bibitem[Jang et~al.(2017)Jang, Gu, and Poole]{jang2016categorical}
Eric Jang, Shixiang Gu, and Ben Poole.
\newblock Categorical reparameterization with {Gumbel-Softmax}.
\newblock In \emph{ICLR}, 2017.

\bibitem[Jiang et~al.(2023)Jiang, Deng, Singh, and Ahn]{jiang2023objectcentric}
Jindong Jiang, Fei Deng, Gautam Singh, and Sungjin Ahn.
\newblock Object-centric slot diffusion.
\newblock In \emph{NeurIPS}, volume~36, pp.\  8563--8601, 2023.

\bibitem[Kingma \& Welling(2014)Kingma and Welling]{kingma2013auto}
Diederik~P Kingma and Max Welling.
\newblock Auto-encoding variational {Bayes}.
\newblock In \emph{ICLR}, 2014.

\bibitem[Kipf et~al.(2022)Kipf, Elsayed, Mahendran, Stone, Sabour, Heigold, Jonschkowski, Dosovitskiy, and Greff]{savi}
Thomas Kipf, Gamaleldin~F. Elsayed, Aravindh Mahendran, Austin Stone, Sara Sabour, Georg Heigold, Rico Jonschkowski, Alexey Dosovitskiy, and Klaus Greff.
\newblock Conditional object-centric learning from video.
\newblock In \emph{ICLR}, 2022.

\bibitem[Kirilenko et~al.(2024)Kirilenko, Vorobyov, Kovalev, and Panov]{slot_mixture_module}
Daniil Kirilenko, Vitaliy Vorobyov, Alexey Kovalev, and Aleksandr Panov.
\newblock {Object-Centric Learning with Slot Mixture Module}.
\newblock In \emph{ICLR}, 2024.

\bibitem[Kirillov et~al.(2023)Kirillov, Mintun, Ravi, Mao, Rolland, Gustafson, Xiao, Whitehead, Berg, Lo, et~al.]{kirillov2023segment}
Alexander Kirillov, Eric Mintun, Nikhila Ravi, Hanzi Mao, Chloe Rolland, Laura Gustafson, Tete Xiao, Spencer Whitehead, Alexander~C Berg, Wan-Yen Lo, et~al.
\newblock Segment anything.
\newblock In \emph{2023 IEEE/CVF international conference on computer vision (ICCV)}, pp.\  3992--4003. IEEE, 2023.

\bibitem[Lin et~al.(2026)Lin, Chen, Liew, Chen, Li, Zhao, Peng, Guo, Zhou, Shi, Feng, and Kang]{depth_anything_v3}
Haotong Lin, Sili Chen, Jun~Hao Liew, Donny~Y. Chen, Zhenyu Li, Yang Zhao, Sida Peng, Hengkai Guo, Xiaowei Zhou, Guang Shi, Jiashi Feng, and Bingyi Kang.
\newblock Depth anything 3: {R}ecovering the visual space from any views.
\newblock In \emph{ICLR}, volume 2026, pp.\  141261--141285, 2026.

\bibitem[Locatello et~al.(2020)Locatello, Weissenborn, Unterthiner, Mahendran, Heigold, Uszkoreit, Dosovitskiy, and Kipf]{slot_attention}
Francesco Locatello, Dirk Weissenborn, Thomas Unterthiner, Aravindh Mahendran, Georg Heigold, Jakob Uszkoreit, Alexey Dosovitskiy, and Thomas Kipf.
\newblock Object-centric learning with slot attention.
\newblock In H.~Larochelle, M.~Ranzato, R.~Hadsell, M.F. Balcan, and H.~Lin (eds.), \emph{NeurIPS}, volume~33, pp.\  11525--11538, 2020.

\bibitem[Majellaro et~al.(2025)Majellaro, Collu, Plaat, and Moerland]{disa}
Riccardo Majellaro, Jonathan Collu, Aske Plaat, and Thomas~M. Moerland.
\newblock Explicitly disentangled representations in object-centric learning.
\newblock \emph{TMLR}, 2025.
\newblock ISSN 2835-8856.

\bibitem[Manasyan et~al.(2025)Manasyan, Seitzer, Radovic, Martius, and Zadaianchuk]{slot_contrast}
Anna Manasyan, Maximilian Seitzer, Filip Radovic, Georg Martius, and Andrii Zadaianchuk.
\newblock Temporally consistent object-centric learning by contrasting slots.
\newblock In \emph{CVPR}, pp.\  5401--5411, 2025.

\bibitem[Moon et~al.(2026)Moon, Seong, and Heo]{moon2026reconstruction}
WonJun Moon, Hyun~Seok Seong, and Jae-Pil Heo.
\newblock Reconstruction-guided slot curriculum: Addressing object over-fragmentation in video object-centric learning.
\newblock In \emph{CVPR}, pp.\  25001--25010, 2026.

\bibitem[Perazzi et~al.(2016)Perazzi, Pont-Tuset, McWilliams, Van~Gool, Gross, and Sorkine-Hornung]{Perazzi_2016_CVPR}
Federico Perazzi, Jordi Pont-Tuset, Brian McWilliams, Luc Van~Gool, Markus Gross, and Alexander Sorkine-Hornung.
\newblock A benchmark dataset and evaluation methodology for video object segmentation.
\newblock In \emph{CVPR}, pp.\  724--732, June 2016.

\bibitem[Pont-Tuset et~al.(2017{\natexlab{a}})Pont-Tuset, Arbelaez, Barron, Marques, and Malik]{pont2016multiscale}
Jordi Pont-Tuset, Pablo Arbelaez, Jonathan~T Barron, Ferran Marques, and Jitendra Malik.
\newblock Multiscale combinatorial grouping for image segmentation and object proposal generation.
\newblock \emph{IEEE transactions on pattern analysis and machine intelligence}, 39\penalty0 (1):\penalty0 128--140, 2017{\natexlab{a}}.

\bibitem[Pont-Tuset et~al.(2017{\natexlab{b}})Pont-Tuset, Perazzi, Caelles, Arbel{\'a}ez, Sorkine-Hornung, and Van~Gool]{davis2017}
Jordi Pont-Tuset, Federico Perazzi, Sergi Caelles, Pablo Arbel{\'a}ez, Alex Sorkine-Hornung, and Luc Van~Gool.
\newblock The 2017 {DAVIS} challenge on video object segmentation.
\newblock \emph{arXiv preprint arXiv:1704.00675}, 2017{\natexlab{b}}.

\bibitem[Rand(1971)]{rand1971objective}
William~M Rand.
\newblock Objective criteria for the evaluation of clustering methods.
\newblock \emph{Journal of the American Statistical association}, 66\penalty0 (336):\penalty0 846--850, 1971.

\bibitem[Ravi et~al.(2025)Ravi, Gabeur, Hu, Hu, Ryali, Ma, Khedr, R{\"a}dle, Rolland, Gustafson, et~al.]{ravi2025sam}
Nikhila Ravi, Valentin Gabeur, Yuan-Ting Hu, Ronghang Hu, Chaitanya Ryali, Tengyu Ma, Haitham Khedr, Roman R{\"a}dle, Chloe Rolland, Laura Gustafson, et~al.
\newblock Sam 2: Segment anything in images and videos.
\newblock In \emph{International Conference on Learning Representations}, volume 2025, pp.\  28085--28128, 2025.

\bibitem[Rezazadeh et~al.(2024)Rezazadeh, Badithela, Desingh, and Choi]{slotgnn}
Alireza Rezazadeh, Athreyi Badithela, Karthik Desingh, and Changhyun Choi.
\newblock {SlotGNN}: Unsupervised discovery of multi-object representations and visual dynamics.
\newblock In \emph{ICRA}, pp.\  17508--17514, 2024.

\bibitem[Seitzer et~al.(2023)Seitzer, Horn, Zadaianchuk, Zietlow, Xiao, Simon-Gabriel, He, Zhang, Sch{\"o}lkopf, Brox, and Locatello]{dinosaur}
Maximilian Seitzer, Max Horn, Andrii Zadaianchuk, Dominik Zietlow, Tianjun Xiao, Carl-Johann Simon-Gabriel, Tong He, Zheng Zhang, Bernhard Sch{\"o}lkopf, Thomas Brox, and Francesco Locatello.
\newblock Bridging the gap to real-world object-centric learning.
\newblock In \emph{ICLR}, 2023.

\bibitem[Seong et~al.(2026)Seong, Moon, and Heo]{seong2026from}
Hyun~Seok Seong, WonJun Moon, and Jae-Pil Heo.
\newblock From vicious to virtuous cycles: Synergistic representation learning for unsupervised video object-centric learning.
\newblock In \emph{ICLR}, 2026.

\bibitem[Siméoni et~al.(2026)Siméoni, Vo, Seitzer, Baldassarre, Oquab, Jose, Khalidov, Szafraniec, Yi, Ramamonjisoa, Massa, Haziza, Wehrstedt, Wang, Darcet, Moutakanni, Sentana, Roberts, Vedaldi, Tolan, Brandt, Couprie, Mairal, Jégou, Labatut, and Bojanowski]{dinov3}
Oriane Siméoni, Huy~V. Vo, Maximilian Seitzer, Federico Baldassarre, Maxime Oquab, Cijo Jose, Vasil Khalidov, Marc Szafraniec, Seungeun Yi, Michaël Ramamonjisoa, Francisco Massa, Daniel Haziza, Luca Wehrstedt, Jianyuan Wang, Timothée Darcet, Théo Moutakanni, Leonel Sentana, Claire Roberts, Andrea Vedaldi, Jamie Tolan, John Brandt, Camille Couprie, Julien Mairal, Hervé Jégou, Patrick Labatut, and Piotr Bojanowski.
\newblock {DINOv3}.
\newblock \emph{TMLR}, 2026.

\bibitem[Singh et~al.(2022{\natexlab{a}})Singh, Deng, and Ahn]{slate}
Gautam Singh, Fei Deng, and Sungjin Ahn.
\newblock Illiterate {DALL-E} learns to compose.
\newblock In \emph{ICLR}, 2022{\natexlab{a}}.

\bibitem[Singh et~al.(2022{\natexlab{b}})Singh, Wu, and Ahn]{steve}
Gautam Singh, Yi-Fu Wu, and Sungjin Ahn.
\newblock Simple unsupervised object-centric learning for complex and naturalistic videos.
\newblock In \emph{NeurIPS}, volume~35, pp.\  18181--18196, 2022{\natexlab{b}}.

\bibitem[Tang et~al.(2023)Tang, Zhu, Lei, and Zhang]{phycine}
Qu~Tang, Xiangyu Zhu, Zhen Lei, and Zhaoxiang Zhang.
\newblock Intrinsic physical concepts discovery with object-centric predictive models.
\newblock In \emph{CVPR}, pp.\  23252--23261, 2023.

\bibitem[Villar-Corrales \& Behnke(2025)Villar-Corrales and Behnke]{playslot}
Angel Villar-Corrales and Sven Behnke.
\newblock {PlaySlot}: Learning inverse latent dynamics for controllable object-centric video prediction and planning.
\newblock In \emph{ICML}, volume 267, pp.\  61534--61563, 2025.

\bibitem[Villar-Corrales et~al.(2023)Villar-Corrales, Wahdan, and Behnke]{ocvp}
Angel Villar-Corrales, Ismail Wahdan, and Sven Behnke.
\newblock Object-centric video prediction via decoupling of object dynamics and interactions.
\newblock In \emph{ICIP}, pp.\  570--574. IEEE, 2023.

\bibitem[Watters et~al.(2019)Watters, Matthey, Burgess, and Lerchner]{watters2019spatial}
Nicholas Watters, Loic Matthey, Christopher~P Burgess, and Alexander Lerchner.
\newblock Spatial broadcast decoder: A simple architecture for learning disentangled representations in {VAE}s.
\newblock \emph{arXiv preprint arXiv:1901.07017}, 2019.

\bibitem[Wu et~al.(2023)Wu, Dvornik, Greff, Kipf, and Garg]{wu2023slotformer}
Ziyi Wu, Nikita Dvornik, Klaus Greff, Thomas Kipf, and Animesh Garg.
\newblock {SlotFormer}: Unsupervised visual dynamics simulation with object-centric models.
\newblock In \emph{ICLR}, 2023.

\bibitem[Xiao et~al.(2024)Xiao, Yao, Li, and Chua]{Xiao_2024_CVPR}
Junbin Xiao, Angela Yao, Yicong Li, and Tat-Seng Chua.
\newblock Can {I} trust your answer? {V}isually grounded video question answering.
\newblock In \emph{CVPR}, pp.\  13204--13214, June 2024.

\bibitem[Yang et~al.(2019)Yang, Fan, and Xu]{Yang2019vis}
Linjie Yang, Yuchen Fan, and Ning Xu.
\newblock Video instance segmentation.
\newblock In \emph{ICCV}, pp.\  5188--5197, 2019.

\bibitem[Yi et~al.(2020)Yi, Gan, Li, Kohli, Wu, Torralba, and Tenenbaum]{clevrer}
Kexin Yi, Chuang Gan, Yunzhu Li, Pushmeet Kohli, Jiajun Wu, Antonio Torralba, and Joshua~B Tenenbaum.
\newblock {CLEVRER}: Collision events for video representation and reasoning.
\newblock In \emph{ICLR}, 2020.

\bibitem[Zadaianchuk et~al.(2023)Zadaianchuk, Seitzer, and Martius]{videosaur}
Andrii Zadaianchuk, Maximilian Seitzer, and Georg Martius.
\newblock {Object-Centric Learning for Real-World Videos by Predicting Temporal Feature Similarities}.
\newblock In \emph{NeurIPS}, volume~36, pp.\  61514--61545, 2023.

\bibitem[Zou et~al.(2025)Zou, Zhu, Zhang, and Lei]{zou2025tdgnet}
Junhong Zou, Xiangyu Zhu, Zhaoxiang Zhang, and Zhen Lei.
\newblock Top-down guidance for learning object-centric representations.
\newblock In \emph{IJCAI}, pp.\  2548--2556, 2025.
\newblock \doi{10.24963/ijcai.2025/284}.

\end{thebibliography}

\appendix
\section{Implementation details} \label{app:implementation}

This section provides additional implementation details for UniSlot. Given an RGB frame $x_t\in\mathbb{R}^{H\times W\times3}$, a frozen self-supervised encoder $F_{\mathrm{feat}}$~\citep{dino} extracts $N$ dense patch features, which are projected to the slot-grouping space using a two-layer MLP:
\begin{align}
    g_t &= F_{\mathrm{feat}}(x_t),
    & g_t &\in\mathbb{R}^{N\times D_{\mathrm{feat}}}, \\
    h_t &= F_{\mathrm{proj}}(g_t),
    & h_t &\in\mathbb{R}^{N\times D}.
\end{align}
Each slot is represented as $S_t^k=(v_t^k,p_t^k,s_t^k)$, with appearance $v_t^k\in\mathbb{R}^{D}$ and 2D position and scale $p_t^k,s_t^k\in\mathbb{R}^{2}$. Architecture and training hyperparameters are reported in \cref{tab:unislot_hyperparams}.

\subsection{Recurrent disentangled slot grouping} \label{app:grouping}

At timestep $t$, the grouping module receives projected features $h_t\in\mathbb{R}^{N\times D}$ and predicted slot states $\tilde S_t^k=(\tilde v_t^k,\tilde p_t^k,\tilde s_t^k)$. Building on ISA~\citep{isa}, these states are refined through $T_{\mathrm{slot}}$ iterations of pose-normalized attention.

\paragraph{Pose-normalized attention}
Each slot defines a relative coordinate system from its current position and scale:
\begin{equation}
    \mathrm{rel\_grid}_t^k
    =
    \frac{\mathrm{abs\_grid}-p_t^k}{s_t^k},
    \qquad
    \mathrm{abs\_grid}\in\mathbb{R}^{N\times2}.
\end{equation}
The relative coordinates are injected into slot-specific keys and values,
\begin{align}
    \mathrm{keys}_t^k
    &= f_K(h_t)+g(\mathrm{rel\_grid}_t^k),
    &
    \mathrm{values}_t^k
    &= f_V(h_t)+g(\mathrm{rel\_grid}_t^k),
\end{align}
while queries are computed from the slot appearance embeddings. Attention is first normalized across slots and then, after adding $\epsilon$, normalized across spatial locations:
\begin{align}
    \bar a_{t,n}^k
    &=
    \operatorname{softmax}_{k}(a_{t,n}^k), \\
    \mathrm{attn}_{t,n}^k
    &=
    \frac{\bar a_{t,n}^k+\epsilon}
    {\sum_m(\bar a_{t,m}^k+\epsilon)}.
\end{align}
Position and scale are updated from the center and spatial spread of the resulting attention maps:
\begin{align}
    p_t^k
    &=
    \sum_n
    \mathrm{attn}_{t,n}^k\,\mathrm{abs\_grid}_n,
    \\
    s_t^k
    &=
    \sqrt{
        \sum_n
        \mathrm{attn}_{t,n}^k
        (\mathrm{abs\_grid}_n-p_t^k)^2
    }.
\end{align}
Appearance $v_t^k$ is updated using the Slot Attention GRU and MLP. Spatial information is therefore represented explicitly by $(p_t^k,s_t^k)$ rather than being absorbed entirely into the appearance embedding.

\paragraph{Recurrent initialization}
Previous slot states initialize grouping in the next frame. Appearance is propagated through a stochastic predictor $G_{\mathrm{pred}}$:
\begin{equation}
    (\mu_t^k,\log\sigma_t^k)
    =
    G_{\mathrm{pred}}(v_{t-1}^k),
    \qquad
    \tilde v_t^k
    \sim
    \mathcal{N}(\mu_t^k,\sigma_t^k),
\end{equation}
using the reparameterization trick~\citep{kingma2013auto}. Geometry is propagated directly:
\begin{equation}
    (\tilde p_t^k,\tilde s_t^k)
    =
    (p_{t-1}^k,s_{t-1}^k).
\end{equation}
The resulting state is refined as
\begin{equation}
    \{(v_t^k,p_t^k,s_t^k)\}_{k=1}^{K}
    =
    G_{\mathrm{slot}}(h_t,\tilde S_t).
\end{equation}
For the first frame, $\tilde S_1^k=S_{\mathrm{init}}^k$, where $S_{\mathrm{init}}^k$ is a learned initial slot state.

\paragraph{Adaptive gating}
Following Adaptive Slot Attention~\citep{fan2024adaptive}, a lightweight gate predicts whether each refined slot is active:
\begin{equation}
    z_t^k=G_{\mathrm{gate}}(v_t^k),
    \qquad
    z_t^k\in\{0,1\}.
\end{equation}
Only active slots contribute to reconstruction. We use the hard Gumbel--Softmax relaxation~\citep{jang2016categorical} for end-to-end optimization.

\begin{algorithm}[t]
\caption{Recurrent Disentangled Slot Grouping}
\label{alg:gslot}
\small
\begin{algorithmic}[1]
\REQUIRE Patch features $h_t$, previous slots
$\{S_{t-1}^k\}_{k=1}^K$
\IF{$t=1$}
    \STATE $\tilde S_t^k \leftarrow S_{\mathrm{init}}^k$
\ELSE
    \STATE $(\mu_t^k,\log\sigma_t^k)
    \leftarrow G_{\mathrm{pred}}(v_{t-1}^k)$
    \STATE $\tilde v_t^k
    \sim \mathcal{N}(\mu_t^k,\sigma_t^k)$
    \STATE $(\tilde p_t^k,\tilde s_t^k)
    \leftarrow (p_{t-1}^k,s_{t-1}^k)$
\ENDIF
\STATE $(v_t^k,p_t^k,s_t^k)
\leftarrow(\tilde v_t^k,\tilde p_t^k,\tilde s_t^k)$
\FOR{$i=1,\ldots,T_{\mathrm{slot}}$}
    \FOR{each slot $k$}
        \STATE $\mathrm{rel\_grid}_t^k
        \leftarrow(\mathrm{abs\_grid}-p_t^k)/s_t^k$
        \STATE $\mathrm{keys}_t^k
        \leftarrow f_K(h_t)+g(\mathrm{rel\_grid}_t^k)$
        \STATE $\mathrm{values}_t^k
        \leftarrow f_V(h_t)+g(\mathrm{rel\_grid}_t^k)$
        \STATE $\mathrm{queries}_t^k
        \leftarrow f_Q(v_t^k)$
    \ENDFOR
    \STATE Normalize attention across slots and spatial locations
    \FOR{each slot $k$}
        \STATE $p_t^k \leftarrow$ attention-weighted center
        \STATE $s_t^k \leftarrow$ attention-weighted spatial spread
        \STATE $v_t^k \leftarrow$ Slot Attention GRU + MLP update
    \ENDFOR
\ENDFOR
\FOR{each slot $k$}
    \STATE $z_t^k \leftarrow G_{\mathrm{gate}}(v_t^k)$
\ENDFOR
\RETURN $\{(v_t^k,p_t^k,s_t^k,z_t^k)\}_{k=1}^K$
\end{algorithmic}
\end{algorithm}

\subsection{3D-Aware spatial broadcast decoding} \label{app:decoding}

Each active slot is decoded independently following the spatial broadcasting principle~\citep{watters2019spatial}. At the initial decoder resolution $r\times r$, position and scale define a slot-specific coordinate grid:
\begin{equation}
    \mathrm{rel\_grid}_t^k
    =
    \frac{\mathrm{abs\_grid}-p_t^k}{s_t^k},
    \qquad
    \mathrm{rel\_grid}_t^k\in\mathbb{R}^{r\times r\times2}.
\end{equation}
The appearance embedding is spatially broadcast and combined with a learned projection of this grid:
\begin{equation}
    U_t^k
    =
    \mathrm{SB}(v_t^k)
    +
    D_{\mathrm{pos}}(\mathrm{rel\_grid}_t^k).
\end{equation}
A shared CNN decoder independently predicts a visual component $\hat y_t^k$, depth component $\hat d_t^k$, and alpha mask $\alpha_t^k$:
\begin{equation}
    (\hat y_t^k,\hat d_t^k,\alpha_t^k)
    =
    D_{\mathrm{spatial}}(U_t^k).
\end{equation}
The visual and depth predictions are composited over active slots:
\begin{align}
    \hat y_t
    &=
    \sum_{k=1}^{K}
    z_t^k\,\alpha_t^k\odot\hat y_t^k,
    &
    \hat d_t
    &=
    \sum_{k=1}^{K}
    z_t^k\,\alpha_t^k\odot\hat d_t^k.
\end{align}
The alpha masks are predicted independently and are not normalized across slots.

\paragraph{Visual reconstruction targets}
For CLEVRER, MOVi-A, and MOVi-B, we reconstruct RGB pixels directly, \textit{i.e.}, $y_t=x_t$. For MOVi-C, MOVi-E, and YouTube-VIS, we reconstruct the projected dense features, \textit{i.e.}, $y_t=h_t$. Decoder output dimensions are adjusted accordingly.

\paragraph{Monocular depth targets}
For each RGB frame, we generate pseudo-depth using the frozen \texttt{DA3MONO-LARGE} model from Depth Anything v3~\citep{depth_anything_v3}. Depth predictions are normalized independently per video by clipping to the 2nd and 98th percentiles, linearly rescaling to $[0,1]$, and then mapping to $[-1,1]$ via $d\leftarrow(d-0.5)/0.5$. They require neither depth sensors nor ground-truth 3D annotations and are used only during training.

Visual content and depth are decoded using the same slot-specific position, scale, and alpha mask. The depth objective therefore supervises the same spatial decomposition as the visual reconstruction, introducing depth information into the learned geometric representation.

\subsection{Training Objectives} \label{app:objectives}

UniSlot combines visual and depth reconstruction with temporal alignment, slot separation, adaptive-slot sparsity, and stochastic predictor regularization:
\begin{equation}
    \mathcal{L}
    =
    \lambda_{\mathrm{recon}}\mathcal{L}_{\mathrm{recon}}
    +\lambda_{\mathrm{depth}}\mathcal{L}_{\mathrm{depth}}
    +\lambda_{\mathrm{time}}\mathcal{L}_{\mathrm{time}}
    +\lambda_{\mathrm{sep}}\mathcal{L}_{\mathrm{sep}}
    +\lambda_{\mathrm{spars}}\mathcal{L}_{\mathrm{spars}}
    +\lambda_{\mathrm{KL}}\mathcal{L}_{\mathrm{KL}}.
\label{eq:app_full_loss}
\end{equation}

\paragraph{Reconstruction}
Visual and pseudo-depth targets are reconstructed using mean squared error:
\begin{align}
    \mathcal{L}_{\mathrm{recon}}
    &=
    \mathbb{E}_{t}
    \left[
        \lVert y_t-\hat y_t\rVert_2^2
    \right],
    &
    \mathcal{L}_{\mathrm{depth}}
    &=
    \mathbb{E}_{t}
    \left[
        \lVert d_t-\hat d_t\rVert_2^2
    \right].
\end{align}

\paragraph{Temporal alignment}
Temporal consistency is imposed only on the appearance embeddings of slots active in consecutive frames:
\begin{equation}
    \mathcal{L}_{\mathrm{time}}
    =
    \frac{
        \sum_{t,k}
        z_{t-1}^kz_t^k
        \left[1-\cos(\tilde v_t^k,v_t^k)\right]
    }{
        \sum_{t,k}z_{t-1}^kz_t^k
    }.
\end{equation}
The geometric components $(p_t^k,s_t^k)$ remain free to evolve with object motion. \textcolor{black}{For the ablation aligning all slot components, we apply the same temporal consistency objective separately to appearance, position, and scale, with equal weights, rather than computing similarity on their concatenation.}

\paragraph{Slot separation and sparsity}
We penalize positive cosine similarity between different active appearance embeddings:
\begin{equation}
    \mathcal{L}_{\mathrm{sep}}
    =
    \frac{
        \sum_t\sum_{i\neq j}
        z_t^iz_t^j
        \max\!\left(0,\cos(v_t^i,v_t^j)\right)
    }{
        \sum_t\sum_{i\neq j}z_t^iz_t^j
    }.
\end{equation}
Adaptive-slot sparsity encourages unnecessary slots to deactivate:
\begin{equation}
    \mathcal{L}_{\mathrm{spars}}
    =
    \mathbb{E}_{t,k}[z_t^k].
\end{equation}

\paragraph{Stochastic predictor regularization}
The recurrent predictor parameterizes
\begin{equation}
    q(\tilde v_t^k\mid v_{t-1}^k)
    =
    \mathcal{N}
    \left(
        \mu_t^k,
        \operatorname{diag}\!\left((\sigma_t^k)^2\right)
    \right).
\end{equation}
We regularize its variance toward $0.01I$ while leaving the predicted mean
unconstrained:
\begin{equation}
    \mathcal{L}_{\mathrm{KL}}
    =
    \mathbb{E}_{t,k}
    \left[
        D_{\mathrm{KL}}
        \left(
            \mathcal{N}\!\left(
                \mu_t^k,
                \operatorname{diag}\!\left((\sigma_t^k)^2\right)
            \right)
            \,\middle\|\,
            \mathcal{N}\!\left(\mu_t^k,0.01I\right)
        \right)
    \right].
\end{equation}
This preserves the predicted appearance mean used for recurrent initialization while controlling the sampling variance.

\subsection{Optimization and training} \label{app:optimization}

\paragraph{Optimization}
We use Adam with $(\beta_1,\beta_2)=(0.9,0.999)$ and an exponentially decayed learning rate following linear warmup, with the decay schedule spanning the full training duration. Dataset-specific training configurations are reported in \cref{tab:unislot_hyperparams}.

\paragraph{Progressive regularization}
Regularization is introduced progressively after an initial decomposition has formed. $\mathcal{L}_{\mathrm{spars}}$ is enabled after $50$k steps on CLEVRER and MOVi-A/B and after $100$k steps on MOVi-C/E and YouTube-VIS. $\mathcal{L}_{\mathrm{time}}$ and $\mathcal{L}_{\mathrm{sep}}$ are active from the beginning on CLEVRER and MOVi-A/B and enabled after $50$k steps on MOVi-C/E and YouTube-VIS.

\paragraph{Loss balancing}
Reconstruction provides the primary learning signal, while the remaining objectives shape the slot representation. Exact loss weights are reported in \cref{tab:unislot_hyperparams}.

\section{Experimental details} \label{app:experimental_details}

\subsection{Datasets} \label{app:datasets}

Unless stated otherwise, we use a spatial resolution of $256\times256$. This is the native resolution of CLEVRER and MOVi used in our experiments, and we adopt the same resolution for YouTube-VIS for consistency. CLEVRER and MOVi sequences contain 24 frames, while YouTube-VIS sequences contain up to 84 frames. During training, UniSlot processes 6-frame clips and is applied recurrently to the full sequence at inference.

\paragraph{CLEVRER}
CLEVRER~\citep{clevrer} contains synthetic videos of simple geometric objects undergoing collisions and interactions. Its controlled appearance and dynamics, together with dense instance annotations, provide a clean setting for evaluating object decomposition and temporal identity consistency.

\paragraph{MOVi}
The MOVi benchmarks~\citep{kubric} consist of procedurally generated multi-object videos with increasing visual and geometric complexity. MOVi-A contains simple objects and backgrounds, while MOVi-B introduces more diverse object shapes and appearances. MOVi-C replaces these objects with scanned real-world assets, and MOVi-E further increases scene complexity through more objects, diverse backgrounds, and camera motion. All variants provide dense instance masks and consistent object identities across frames.

\paragraph{YouTube-VIS}
YouTube-VIS~\citep{Yang2019vis} contains real-world videos with temporally consistent instance annotations across diverse object categories. Compared with the synthetic benchmarks, its unconstrained appearance, camera motion, occlusions, and background clutter provide a substantially more challenging setting for unsupervised object decomposition and tracking. We independently train and evaluate on YouTube-VIS 2019 and YouTube-VIS 2021, which contain different video collections.

\paragraph{DAVIS 2017}
DAVIS 2017~\citep{davis2017} is a densely annotated video object segmentation benchmark containing multiple foreground objects under challenging appearance changes, occlusions, and motion. We use DAVIS exclusively for transfer evaluation: models trained on YouTube-VIS 2021 are evaluated on DAVIS without training or fine-tuning on DAVIS.

\subsection{Evaluation Metrics} \label{app:evaluation_metrics}

We evaluate object decomposition and tracking using Adjusted Rand Index (ARI),
Foreground ARI (FG-ARI), mean Best Overlap (mBO), and
$\mathcal{J}\&\mathcal{F}$, reporting each metric where applicable.

\paragraph{Adjusted Rand Index (ARI)}
ARI~\citep{rand1971objective,hubert1985comparing} measures agreement between predicted slot assignments and ground-truth instance segmentation, corrected for chance. Let $n_{ij}$ denote the number of pixels assigned to ground-truth cluster $i$ and predicted cluster $j$, with $a_i=\sum_j n_{ij}$, $b_j=\sum_i n_{ij}$, and $n=\sum_{ij}n_{ij}$. We compute:
\begin{equation}
    \mathrm{ARI}
    =
    \frac{
        \sum_{ij}\binom{n_{ij}}{2}
        -
        \frac{\sum_i\binom{a_i}{2}\sum_j\binom{b_j}{2}}
             {\binom{n}{2}}
    }{
        \frac{1}{2}
        \left[
            \sum_i\binom{a_i}{2}
            +
            \sum_j\binom{b_j}{2}
        \right]
        -
        \frac{\sum_i\binom{a_i}{2}\sum_j\binom{b_j}{2}}
             {\binom{n}{2}}
    }.
    \label{eq:ari}
\end{equation}
ARI is computed over all pixels, including background, whereas FG-ARI uses the
same definition after excluding background pixels. Thus, ARI additionally
reflects foreground--background separation, while FG-ARI isolates foreground
object grouping.

\paragraph{Mean Best Overlap (mBO)}
Given ground-truth masks $\mathcal{M}$ and predicted masks $\hat{\mathcal{M}}$, mBO~\citep{pont2016multiscale} matches each ground-truth
object to its best-overlapping predicted mask:
\begin{equation}
    \mathrm{mBO}
    =
    \frac{1}{|\mathcal{M}|}
    \sum_{M\in\mathcal{M}}
    \max_{\hat M\in\hat{\mathcal{M}}}
    \frac{|M\cap\hat M|}{|M\cup\hat M|}.
    \label{eq:mbo}
\end{equation}
Higher mBO indicates better object-mask overlap.

\paragraph{$\boldsymbol{\mathcal{J}\&\mathcal{F}}$}
For video object segmentation and tracking, we additionally report the standard DAVIS $\mathcal{J}\&\mathcal{F}$ metric~\citep{davis2017}. Region similarity $\mathcal{J}$ measures mask intersection over union, while contour accuracy $\mathcal{F}$ is the harmonic mean of boundary precision $P_c$ and recall $R_c$:
\begin{equation}
    \mathcal{J}=\frac{|M\cap\hat M|}{|M\cup\hat M|},
    \qquad
    \mathcal{F}=\frac{2P_cR_c}{P_c+R_c},
    \qquad
    \mathcal{J}\&\mathcal{F}=\frac{\mathcal{J}+\mathcal{F}}{2}.
    \label{eq:j_and_f}
\end{equation}
Boundary precision and recall are computed by matching predicted and ground-truth contours within the standard boundary tolerance. We report the sequence-averaged score.

\subsection{Baseline implementation} \label{app:baseline_implementation}

We evaluate all baselines using their official implementations and preserve their original architectures and training objectives wherever possible. For comparisons based on pretrained visual features, we use the feature encoder specified by each method and otherwise match preprocessing and evaluation protocols to UniSlot. The number of slots is chosen consistently with the maximum scene complexity of each dataset.

SAVi~\citep{savi}, VideoSAUR~\citep{videosaur}, SlotContrast~\citep{slot_contrast}, SRL~\citep{seong2026from}, and SlotCurri~\citep{moon2026reconstruction} form the video-based baselines in the main paper. We additionally evaluate the image-based ISA~\citep{isa}, Adaptive Slot Attention~\citep{fan2024adaptive}, and DINOSAUR~\citep{dinosaur} in the extended results in Appendix~\ref{app:extra_quantitative}.

SlotContrast, SRL, and SlotCurri rely on DINOv2 features extracted with ViT backbones of patch size $14$. We therefore evaluate these methods at $252\times252$, the closest resolution to our default $256\times256$ that is divisible by $14$. For SRL and SlotCurri, we retrain the models using their official implementations rather than their released checkpoints, which use $336\times336$ inputs on MOVi and $518\times518$ inputs on YouTube-VIS. This ensures that all methods are evaluated at comparable input resolutions. Complete baseline configurations are reported in \cref{app:baseline_hyperparams}.

\subsection{Object dynamics prediction} \label{app:object_dynamics_prediction}

To evaluate the learned representations for downstream dynamics modeling, we train SlotFormer~\citep{wu2023slotformer} on frozen slot trajectories extracted by each object-centric model. We condition on $H$ observed frames and autoregressively predict future states using a $4$-layer Transformer with latent dimension $128$, $8$ attention heads, feed-forward dimension $512$, and sinusoidal temporal positional embeddings. No slot-index positional encoding is used.

For conventional representations, SlotFormer operates directly on the slot embeddings. For UniSlot, we use the structured state
\begin{equation}
    q_t^k=[\,v_t^k;\,p_t^k;\,s_t^k;\,z_t^k\,]
    \in\mathbb{R}^{D+5}.
\end{equation}
The model predicts appearance and scale together with a position displacement $\Delta p_t^k$, from which $\hat p_{t+1}^k=p_t^k+\Delta p_t^k$. Predicted scales are constrained to be positive using Softplus, while slot activation is propagated from the latest state rather than predicted.

We train SlotFormer using a weighted MSE over appearance, position, and scale,
\begin{equation}
    \mathcal{L}_{\mathrm{dyn}}
    =
    \lambda_v\mathcal{L}_v
    +\lambda_p\mathcal{L}_p
    +\lambda_s\mathcal{L}_s,
\end{equation}
Loss weights and training configurations are reported in \cref{tab:dynamics_hyperparameters}. The object-centric models remain frozen throughout dynamics training.

\subsection{DAVIS 2017 transfer} \label{app:davis2017_transfer}

To evaluate transfer to unseen real-world videos, we apply models trained on
YouTube-VIS 2021 directly to DAVIS 2017, without training or fine-tuning on
DAVIS. We use the same inference procedure and slot configuration as on
YouTube-VIS and evaluate the resulting object masks over complete video
sequences using ARI, FG-ARI, mBO, and $\mathcal{J}\&\mathcal{F}$.

\section{Model and training configurations} \label{app:configurations}

\subsection{UniSlot hyperparameters} \label{app:unislot_hyperparams}

Unless stated otherwise, training uses $6$-frame clips at $H\times W=256\times256$ resolution, while inference is performed recurrently over the full video sequence. The frozen feature encoder $F_{\mathrm{feat}}$ is a DINO ViT-B/16~\citep{dino} with $D_{\mathrm{feat}}=768$, followed by the projection $F_{\mathrm{proj}}$ to the slot dimension $D=128$. Recurrent disentangled grouping uses $T_{\mathrm{slot}}=2$ refinement iterations. The stochastic appearance predictor $G_{\mathrm{pred}}$ is a two-layer MLP with hidden dimension $256$, and the adaptive gate $G_{\mathrm{gate}}$ uses the same hidden dimension.

We train with Adam using a peak learning rate of $10^{-4}$, $5$k warmup steps, batch size $64$, and gradient clipping at norm $0.05$. Pseudo-depth targets $d_t$ are generated offline using the frozen \texttt{DA3MONO-LARGE} model~\citep{depth_anything_v3} and are used only during training. \Cref{tab:unislot_hyperparams} reports the dataset-specific configurations and loss weights used in \cref{eq:app_full_loss}.

\begin{table}[h]
\centering
\caption{Dataset-specific UniSlot hyperparameters.}
\label{tab:unislot_hyperparams}
\vspace{2pt}
\small
\setlength{\tabcolsep}{4pt}
\begin{tabular}{lccccc}
\toprule
\textbf{Hyperparameter}
& \textbf{CLEVRER}
& \textbf{MOVi-A/B}
& \textbf{MOVi-C}
& \textbf{MOVi-E}
& \textbf{YTVIS 2019/21} \\
\midrule
Training steps
    & 100k & 100k & 200k & 200k & 200k \\
Slots $K$
    & 7 & 11 & 11 & 15 & 7 \\
Crop
    & Center & Full & Full & Full & Center \\
Augmentation
    & -- & -- & -- & -- & Horiz.\ flip \\
Visual target $y_t$
    & RGB & RGB & $h_t$ & $h_t$ & $h_t$ \\
Decoder hidden dim.
    & 64 & 64 & 64 & 64 & 256 \\
Upsampling stages
    & 4 & 4 & 0 & 0 & 0 \\
\midrule
$\lambda_{\mathrm{recon}}$
    & 1.0 & 1.0 & 1.0 & 0.3 & 0.3 \\
$\lambda_{\mathrm{depth}}$
    & $10^{-2}$ & $10^{-2}$ & $10^{-2}$ & $10^{-2}$ & $10^{-2}$ \\
$\lambda_{\mathrm{time}}$
    & $10^{-3}$ & $10^{-3}$ & $10^{-3}$ & $10^{-3}$ & $10^{-3}$ \\
$\lambda_{\mathrm{sep}}$
    & $10^{-3}$ & $10^{-3}$ & $10^{-3}$ & $10^{-3}$ & $10^{-3}$ \\
$\lambda_{\mathrm{spars}}$
    & $10^{-2}$ & $10^{-2}$ & $10^{-2}$ & $10^{-2}$ & $10^{-2}$ \\
$\lambda_{\mathrm{KL}}$
    & $10^{-4}$ & $10^{-4}$ & $10^{-4}$ & $10^{-4}$ & $10^{-4}$ \\
\bottomrule
\end{tabular}
\end{table}

\subsection{Object dynamics prediction hyperparameters} \label{app:dynamics_hyperparameters}

We use dataset-specific burn-in and rollout lengths and balance the prediction of appearance, position, and scale through the weights $\lambda_v$, $\lambda_p$, and $\lambda_s$, respectively. The complete training configuration is reported in \cref{tab:dynamics_hyperparameters}.

\begin{table}[h]
    \centering
    \caption{Hyperparameters for downstream object dynamics prediction. Dataset-specific settings for the SlotFormer predictor trained on UniSlot representations.}
    \vspace{2pt}
    \label{tab:dynamics_hyperparameters}
    \small
    \begin{tabular}{lccc}
        \toprule
        \textbf{Hyperparameter}
        & \textbf{MOVi-C}
        & \textbf{MOVi-E}
        & \textbf{YTVIS21} \\
        \midrule
        Burn-in frames                   & 6  & 6  & 10 \\
        Rollout length                   & 10 & 10 & 5 \\
        Appearance weight $\lambda_v$    & 0.1 & 0.25 & 0.25 \\
        Position weight $\lambda_p$      & 5.0 & 0.25 & 0.25 \\
        Scale weight $\lambda_s$         & 2.5 & 0.25 & 0.25 \\
        Batch size                       & 64 & 64 & 64 \\
        Learning rate                    & $10^{-4}$ & $10^{-4}$ & $10^{-4}$ \\
        Training steps                   & 100'000 & 100'000 & 100'000 \\
        \bottomrule
    \end{tabular}
\end{table}

\subsection{Baseline hyperparameters} \label{app:baseline_hyperparams}

We use the official implementations and configurations of SlotContrast,
VideoSAUR, and SAVi, with the dataset-specific adaptations described below.
Across datasets, SlotContrast and VideoSAUR use $4$-frame training segments,
$3/2$ Slot Attention iterations for the first/subsequent frames, and a
single-layer, $4$-head Transformer predictor. SAVi uses $6$-frame segments,
$2$ Slot Attention iterations, $128$-dimensional slots, and the same
single-layer, $4$-head Transformer predictor. All models are trained with Adam
and gradient clipping at $0.05$.

\paragraph{SlotContrast}
For SlotContrast~\citep{slot_contrast}, all experiments use $4$-frame training
segments, batch size $64$, $2.5$k learning-rate warmup steps, and gradient
clipping at $0.05$. We use $3$ Slot Attention iterations for the first frame
and $2$ thereafter, followed by a single-layer, $4$-head Transformer predictor.
Inputs are resized to $252\times252$ to match the DINOv2 ViT patch size of $14$,
and we use $\lambda_{\mathrm{contrast}}=0.5$ throughout.
Dataset-specific settings are reported in \cref{tab:slotcontrast_hyperparams}.

\begin{table}[h]
\centering
\caption{Dataset-specific SlotContrast hyperparameters. We use the official
configurations of \citet{slot_contrast} for MOVi-C, MOVi-E, and YouTube-VIS,
and the MOVi-C configuration for CLEVRER and MOVi-A/B.}
\vspace{2pt}
\small
\label{tab:slotcontrast_hyperparams}
\begin{tabular}{lccccc}
\toprule
\textbf{Hyperparameter}
& \textbf{CLEVRER}
& \textbf{MOVi-A/B}
& \textbf{MOVi-C}
& \textbf{MOVi-E}
& \textbf{YouTube-VIS} \\
\midrule
Training steps       & 100k & 100k & 100k & 300k & 100k \\
Peak LR              & $4{\cdot}10^{-4}$ & $4{\cdot}10^{-4}$ & $4{\cdot}10^{-4}$ & $8{\cdot}10^{-4}$ & $8{\cdot}10^{-4}$ \\
DINOv2 backbone      & ViT-S/14 & ViT-S/14 & ViT-S/14 & ViT-S/14 & ViT-B/14 \\
Crop                 & Center & Full & Full & Full & Center \\
Augmentation         & -- & -- & -- & -- & H. flip \\
Slots $K$            & 7 & 11 & 11 & 15 & 7 \\
Slot dimension       & 128 & 128 & 128 & 128 & 64 \\
\bottomrule
\end{tabular}
\end{table}

\paragraph{VideoSAUR}
For VideoSAUR~\citep{videosaur}, all experiments use $4$-frame training
segments, batch size $128$, $2.5$k learning-rate warmup steps, a peak learning
rate of $4\cdot10^{-4}$, and gradient clipping at $0.05$. We use DINO
ViT-B/8 features at $256\times256$ resolution, $3/2$ Slot Attention iterations
for the first/subsequent frames, and a single-layer, $4$-head Transformer
predictor. Dataset-specific settings are reported in
\cref{tab:videosaur_hyperparams}.

\begin{table}[h]
\centering
\caption{Dataset-specific VideoSAUR hyperparameters. We use the official
configurations of \citet{videosaur} for MOVi-C, MOVi-E, and YouTube-VIS,
and the MOVi-C configuration for CLEVRER and MOVi-A/B.}
\vspace{2pt}
\small
\label{tab:videosaur_hyperparams}
\begin{tabular}{lccccc}
\toprule
\textbf{Hyperparameter}
& \textbf{CLEVRER}
& \textbf{MOVi-A/B}
& \textbf{MOVi-C}
& \textbf{MOVi-E}
& \textbf{YouTube-VIS} \\
\midrule
Crop                  & Center & Full & Full & Full & Center \\
Augmentation          & -- & -- & -- & -- & H. flip \\
Slots $K$             & 7 & 11 & 11 & 15 & 7 \\
Slot dimension        & 128 & 128 & 128 & 128 & 64 \\
Allocator layers      & 2 & 2 & 2 & 2 & 3 \\
Renderer layers       & 4 & 4 & 4 & 4 & 3 \\
$\lambda_{\mathrm{recon}}$
                      & 0 & 0 & 0 & 0 & 0.1 \\
Temperature $\tau$    & 0.075 & 0.075 & 0.075 & 0.075 & 0.25 \\
\bottomrule
\end{tabular}
\end{table}

\paragraph{SAVi}
For SAVi~\citep{savi}, all experiments use $6$-frame training segments, batch
size $64$, $2.5$k learning-rate warmup steps, a peak learning rate of
$2\cdot10^{-4}$, and gradient clipping at $0.05$. We use $128$-dimensional
slots with $2$ Slot Attention iterations and a single-layer, $4$-head
Transformer predictor. For MOVi-C, MOVi-E, and YouTube-VIS, we replace the CNN
encoder with frozen DINO ViT-B/16 features. Dataset-specific settings are
reported in \cref{tab:savi_hyperparams}.

\begin{table}[h]
\centering
\caption{Dataset-specific SAVi hyperparameters. We follow the official
configuration of \citet{savi} for MOVi-A/B and CLEVRER, and use DINO features
for the more visually complex datasets.}
\vspace{2pt}
\small
\label{tab:savi_hyperparams}
\begin{tabular}{lccccc}
\toprule
\textbf{Hyperparameter}
& \textbf{CLEVRER}
& \textbf{MOVi-A/B}
& \textbf{MOVi-C}
& \textbf{MOVi-E}
& \textbf{YouTube-VIS} \\
\midrule
Encoder          & CNN & CNN & DINO ViT-B/16 & DINO ViT-B/16 & DINO ViT-B/16 \\
Crop             & Center & Full & Full & Full & Center \\
Augmentation     & -- & -- & -- & -- & H. flip \\
Slots $K$        & 7 & 11 & 11 & 15 & 7 \\
\bottomrule
\end{tabular}
\end{table}

\paragraph{SRL}
For SRL~\citep{seong2026from}, we use the official dataset-specific configurations with $4$-frame training segments, $2.5$k learning-rate warmup steps, Adam optimization, gradient clipping at $0.05$, and $2$ Slot Attention iterations. We use the prescribed DINOv2 ViT-S/14 features for MOVi-C and DINOv2 ViT-B/14 features for MOVi-E, while retaining the official dataset-specific architecture and loss settings. All experiments are run at $252\times252$ resolution. For CLEVRER, we use the MOVi-C configuration with $K=7$ slots; for YouTube-VIS 2019, we use the YouTube-VIS 2021 configuration. Dataset-specific settings are reported in \cref{tab:srl_hyperparams}.

\begin{table}[h]
\centering
\caption{Dataset-specific SRL hyperparameters. We follow the official
configurations for MOVi-C, MOVi-E, and YouTube-VIS 2021. CLEVRER uses the
MOVi-C configuration with $K=7$, and YouTube-VIS 2019 uses the YouTube-VIS
2021 configuration.}
\vspace{2pt}
\small
\label{tab:srl_hyperparams}
\begin{tabular}{lccccc}
\toprule
\textbf{Hyperparameter}
& \textbf{CLEVRER}
& \textbf{MOVi-C}
& \textbf{MOVi-E}
& \textbf{YTVIS 2019}
& \textbf{YTVIS 2021} \\
\midrule
Slots $K$
& 7 & 11 & 15 & 7 & 7 \\
Slot dimension
& 64 & 64 & 128 & 64 & 64 \\
DINOv2 backbone
& ViT-S/14 & ViT-S/14 & ViT-B/14 & ViT-B/14 & ViT-B/14 \\
Training steps
& 100k & 100k & 300k & 100k & 100k \\
\bottomrule
\end{tabular}
\end{table}

\paragraph{SlotCurri}
For SlotCurri~\citep{moon2026reconstruction}, we follow the official dataset-specific configurations with $4$-frame training segments, $2.5$k learning-rate warmup steps, Adam optimization, gradient clipping at $0.05$, and $2$ Slot Attention iterations. The official curriculum, loss functions, and remaining architecture settings are retained, while the input resolution is changed to $252\times252$. For CLEVRER, we use the MOVi-C configuration with $K=7$ slots; for YouTube-VIS 2019, we use the YouTube-VIS 2021 configuration. Dataset-specific settings are reported in \cref{tab:slotcurri_hyperparams}.

\begin{table}[h]
\centering
\caption{Dataset-specific SlotCurri hyperparameters. We follow the official
configurations for MOVi-C, MOVi-E, and YouTube-VIS 2021. CLEVRER uses the
MOVi-C configuration with $K=7$, and YouTube-VIS 2019 uses the YouTube-VIS
2021 configuration.}
\vspace{2pt}
\small
\label{tab:slotcurri_hyperparams}
\begin{tabular}{lccccc}
\toprule
\textbf{Hyperparameter}
& \textbf{CLEVRER}
& \textbf{MOVi-C}
& \textbf{MOVi-E}
& \textbf{YTVIS 2019}
& \textbf{YTVIS 2021} \\
\midrule
Slots $K$
& 7 & 11 & 15 & 7 & 7 \\
Slot dimension
& 64 & 64 & 128 & 64 & 64 \\
DINOv2 backbone
& ViT-S/14 & ViT-S/14 & ViT-B/14 & ViT-B/14 & ViT-B/14 \\
Training steps
& 100k & 100k & 300k & 100k & 100k \\
\bottomrule
\end{tabular}
\end{table}

\subsection{Compute requirements and inference efficiency} \label{app:compute_requirements}

\textbf{Training compute.}\quad All training experiments were conducted on NVIDIA RTX3090 (24GB) and A100 (80GB) GPUs. We use gradient accumulation in PyTorch Lightning to simulate larger batch sizes without multi-GPU training. For UniSlot, the effective batch size is 64 clips of 6 frames. A single training run of UniSlot with DINO ViT-B/16 for 100k steps takes approximately 18 hours on one A100 GPU. All results are averaged over 3 random seeds.

\textbf{Inference speed.}\quad We measure inference efficiency on the YouTube-VIS 2021 dataset using an NVIDIA A100-40GB GPU, PyTorch 2.8.0, and CUDA 12.9. All models are evaluated with batch size 8. We report frames per second (FPS), floating-point operations (GFLOPs), and peak GPU memory usage (VRAM). FPS is measured at the frame level, while GFLOPs and VRAM are reported for the same inference configuration.

As shown in ~\Cref{tab:inference_speed}, UniSlot achieves comparable throughput to SlotContrast while using fewer FLOPs and less memory. Compared with VideoSAUR, UniSlot is faster and more efficient in both computation and memory.

\begin{table}[ht!]
    \centering
    \footnotesize
    \caption{Inference efficiency on YouTube-VIS 2021 using an A100-40GB GPU, PyTorch 2.8.0, and batch size 8.}
    \begin{tabular}{@{}lccc@{}}
        \toprule
        \textbf{Method} & \textbf{FPS}$\uparrow$ & \textbf{GFLOPs / batch}$\downarrow$ & \textbf{VRAM (GB)}$\downarrow$ \\
        \midrule
        VideoSAUR    & 220 & 35,900 & 12.52 \\
        SlotContrast & 729 & 13,400 & 8.44 \\
        UniSlot      & 709 & 8,900  & 6.61 \\
        \bottomrule
    \end{tabular}
    \label{tab:inference_speed}
\end{table}

{\color{black}
\section{Visual question answering on CLEVRER} \label{app:vqa}

We detail the setup of \cref{sec:vqa}. The VQA code is adapted from SlotFormer~\citep{wu2023slotformer}, all methods use identical settings except for the input slots.

\paragraph{Data.}
CLEVRER~\citep{clevrer} provides 10{,}000 training and 5{,}000 validation videos with question--answer annotations (\cref{tab:vqa_data}). Descriptive questions are answered by classification over 22 answers; explanatory, predictive, and counterfactual questions are multiple-choice, and each choice is classified as correct or wrong. We train on the training set and report results on the validation set.

\begin{table}[h]
\centering
\caption{\textcolor{black}{Number of CLEVRER questions per type.}}
\label{tab:vqa_data}
\small
\begin{tabular}{lcccc}
\toprule
\textbf{Split} & \textbf{Descriptive} & \textbf{Explanatory} & \textbf{Predictive} & \textbf{Counterfactual} \\
\midrule
Train      & 109{,}952 & 16{,}799 & 7{,}179 & 18{,}642 \\
Validation & 54{,}990  & 8{,}488  & 3{,}557 & 9{,}333 \\
\bottomrule
\end{tabular}
\end{table}

\paragraph{Slot extraction.}
Videos are preprocessed as in our main experiments: the 128 frames of each video are subsampled with stride 2 (64 frames) and center-cropped to $256 \times 256$ ($252 \times 252$ for SlotContrast). Each frozen model is applied recurrently to the full 64-frame videos. Inference is deterministic: the recurrent predictor uses its mean instead of a sample, and UniSlot's gates use the most likely decision. All methods use $K=7$ slots. Each slot is represented by its full concatenated state: $[v_t^k; p_t^k; s_t^k; z_t^k]$ for UniSlot, and the slot vector for SlotContrast and VideoSAUR. We compare each method's complete learned representation, without normalization or additional processing; UniSlot's activation $z_t^k$ is provided as a feature rather than used as a mask.

\paragraph{Future slots.}
Following \citet{wu2023slotformer}, predictive questions are answered from slots predicted beyond the end of the video. For each method, we use the SlotFormer of \cref{subsec:results_dynamics} trained on CLEVRER, condition it on the last 6 observed frames, and autoregressively predict 16 future frames.

\paragraph{Question inputs.}
For each question, we sample $32$ frames at stride $2$. Descriptive, explanatory, and counterfactual questions use the $64$ observed frames, while predictive questions use the final $64$ frames of the extended sequence ($48$ observed and $16$ predicted). The sampling offset is drawn from $\{0,1\}$ during training and fixed to $0$ at evaluation, following~\citet{wu2023slotformer}.

\paragraph{Model and training.}
We use the Aloe Transformer~\citep{ding2021attention} as configured in \citet{wu2023slotformer} (\cref{tab:vqa_hparams}). Each slot at each sampled frame forms one token: its features, concatenated with a vision indicator, are linearly projected to the model dimension. Question and choice words are embedded with type indicators, and a learnable [CLS] token is prepended. The [CLS] output is classified into one of 22 answers for descriptive questions, or as correct or wrong for each choice of a multiple-choice question. The model is trained with cross-entropy on descriptive answers and binary cross-entropy on choices, with equal weights.

\begin{table}[h]
    \centering
    \caption{\textcolor{black}{VQA model and training hyperparameters (identical for all methods).}}
    \label{tab:vqa_hparams}
    \small
    \begin{tabular}{ll}
    \toprule
    Transformer & 12 layers, 8 heads, $d_\text{model}=144$, FFN 512, pre-norm \\
    Positional encoding & Learnable, 257 tokens (224 visual + 32 language + 1 [CLS]) \\
    Parameters & 2.9M \\
    Optimizer & Adam, no weight decay, no gradient clipping \\
    Learning rate & $10^{-3}$, 10\% linear warmup, cosine decay to $10^{-5}$ \\
    Batch size & 256 questions \\
    Training & 238k steps \\
    Compute & 1x A100-40GB GPU \\
    \bottomrule
    \end{tabular}
\end{table}

\paragraph{Metrics.}
We report per-question accuracy for each question type. A multiple-choice question is counted as correct only if all its choices are classified correctly.

\paragraph{Additional results.}
\Cref{tab:vqa_full} extends \cref{tab:vqa} with per-choice accuracy.

\begin{table}[h]
    \centering
    \caption{\textcolor{black}{CLEVRER visual question answering, extended results. Accuracy (\%) on the validation set. Option: per-choice accuracy. Question: per-question accuracy (a multiple-choice question is correct only if all its choices are). Descriptive questions have a single answer.}}
    \label{tab:vqa_full}
    \footnotesize
    \setlength{\tabcolsep}{4pt}
    \begin{tabular}{lccccccc}
    \toprule
    & Descriptive & \multicolumn{2}{c}{Explanatory} & \multicolumn{2}{c}{Predictive} & \multicolumn{2}{c}{Counterfactual} \\
    \cmidrule(lr){3-4} \cmidrule(lr){5-6} \cmidrule(lr){7-8}
    Method & Question & Option & Question & Option & Question & Option & Question \\
    \midrule
    Question only      & 34.54 & 57.22 & 13.42 & 49.58 & 24.15 & 50.11 & 10.04 \\
    VideoSAUR          & 47.68 & 60.42 & 20.92 & 52.29 & 25.61 & 54.09 & 11.69 \\
    SlotContrast       & \underline{82.66} & \underline{91.17} & \underline{77.27} & \underline{82.84} & \underline{69.64} & \underline{79.03} & \underline{47.01} \\
    \rowcolor{gray!15}
    UniSlot (Ours)  & \textbf{89.75} & \textbf{96.13} & \textbf{89.76} & \textbf{90.92} & \textbf{83.19} & \textbf{88.37} & \textbf{67.85} \\
    \bottomrule
    \end{tabular}
\end{table}
}

\begin{figure}[h]
    \centering
    \includegraphics[width=.8\linewidth]{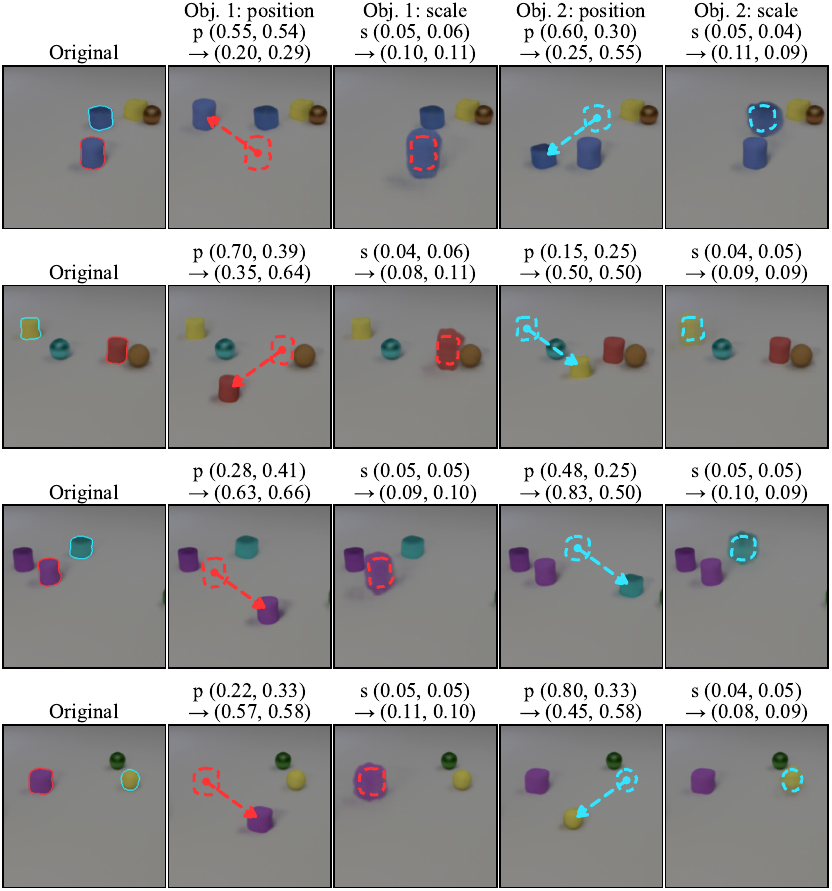}
    \vspace{-0.3cm}
    \caption{\textcolor{black}{Slot interventions on CLEVRER validation videos. Each row shows an unedited reconstruction and independent position or scale edits of two objects (red and cyan). Titles report values before $\rightarrow$ after. Dashed contours indicate the original object extent, and arrows show the displacement of its center of mass.}}
    \label{fig:interventions_clevrer}
\end{figure}

\begin{figure}[h]
    \centering
    \includegraphics[width=.8\linewidth]{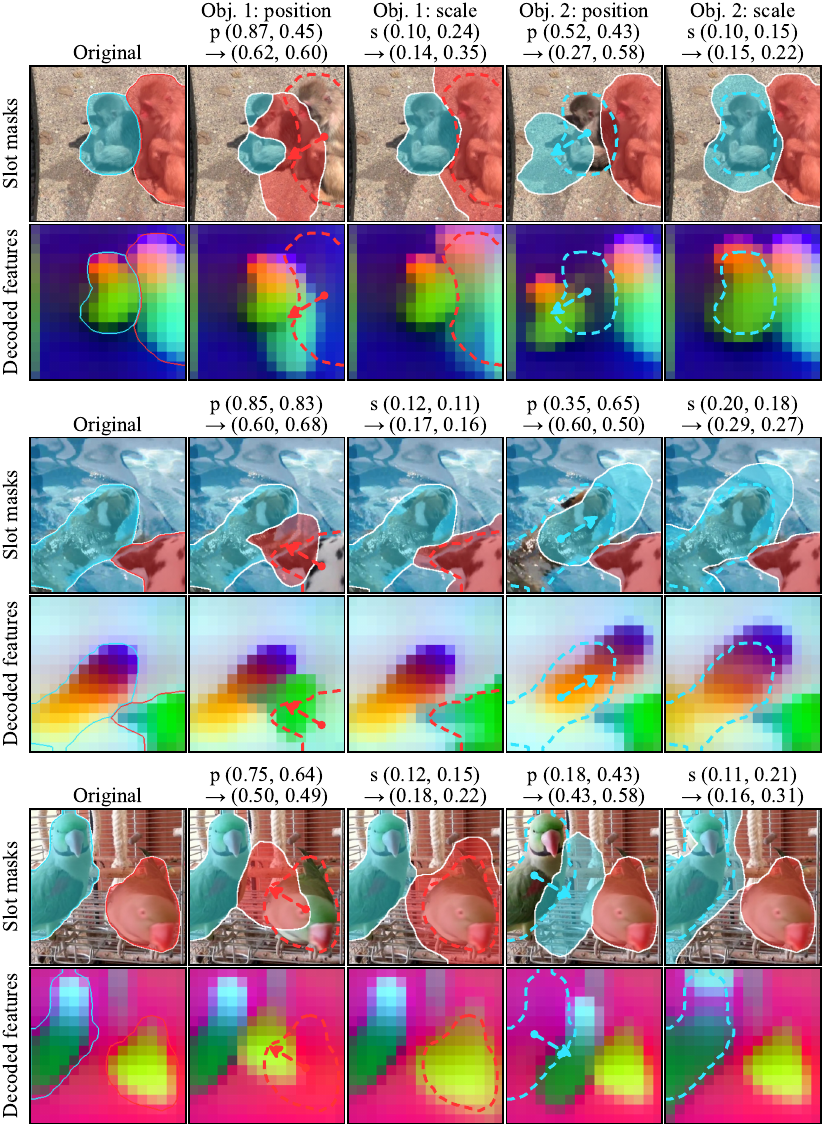}
    \vspace{-0.3cm}
    \caption{\textcolor{black}{Slot interventions on YT-VIS validation videos. Each video shows decoded masks over the original frame and PCA-projected decoded DINO features. Colors and intervention conventions follow \cref{fig:interventions_clevrer}; arrows indicate changes in slot position $p$.}}
    \label{fig:interventions_ytvis}
\end{figure}

{\color{black}
\section{Slot interventions} \label{app:interventions}

To test whether position and scale are disentangled from appearance, we intervene on the slots of trained UniSlot models (CLEVRER and YT-VIS 2021) and decode without retraining. We run the model on a validation video up to a selected frame and extract its slots $(v_t^k, p_t^k, s_t^k, z_t^k)$. We then modify either the position $p$ or scale $s$ of one slot, keeping its remaining components and all other slots fixed. We independently intervene on two objects per video to assess both geometric control and preservation of other objects. Position edits shift both coordinates by specified offsets, while scale edits multiply both components by a constant. Values are reported as $(x, y)$ in image coordinates.

\paragraph{CLEVRER.}
We randomly select two active foreground slots in validation videos. Positions are shifted by $0.35$ horizontally and $0.25$ vertically, and scales are multiplied by $2$. The decoder reconstructs RGB images directly (\cref{fig:interventions_clevrer}). The edited objects move or grow while preserving their color, shape, material, and shadow, whereas the other objects remain visually unchanged.

\paragraph{YT-VIS.}
The YT-VIS decoder reconstructs DINO features on a $16\times16$ patch grid rather than RGB pixels, so the input frame itself remains unchanged. We visualize two decoder outputs. First, the slot masks are upsampled to the frame resolution and assigned by argmax, with only the two selected slots colored. Second, the decoded features are projected to RGB using PCA fitted on the patches of the two selected slots in the unedited reconstruction. The same projection is used across all edits of a video, allowing direct comparison of feature colors. We select slots corresponding to foreground objects rather than background regions. Positions are shifted by $0.25$ horizontally and $0.15$ vertically, and scales are multiplied by $1.5$, since YT-VIS objects are generally larger. The edited slot's mask and decoded features move or grow consistently, while the other selected object's position and features remain unchanged. Regions vacated by the edited slot are taken over by other slots, and minor mask changes can occur due to competition between slots (\cref{fig:interventions_ytvis}).
}

{\color{black}
\section{Comparison with Segment Anything Model 2 (SAM~2)} \label{app:sam2}

SAM~2~\citep{ravi2025sam} produces high-quality video segmentation masks using extensive mask supervision ($1.1$ billion manually verified masks~\citep{kirillov2023segment}), whereas UniSlot learns object-centric representations without mask annotations. We investigate whether SAM~2 can provide a comparable object decomposition and whether its masks can improve UniSlot's representations.

\paragraph{Unprompted video segmentation.}
Since SAM~2 has no prompt-free video mode, we combine its automatic mask generator with its video predictor. Every $8$ frames, the mask generator proposes segments using a $32\times32$ point grid with default settings. Proposals overlapping existing tracks by less than $50\%$ initialize new tracks, while the video predictor propagates existing masks between detection frames. This allows new objects to be introduced throughout the video while propagating existing tracks to encourage temporal identity consistency. We use the largest SAM~2.1 checkpoint (Hiera-L), without manual prompts, category labels, or ground-truth annotations. Both methods are evaluated on the same YouTube-VIS 2021 validation videos at $256\times256$ resolution using identical metrics.

\begin{table}[h]
    \centering
    \caption{\textcolor{black}{Unprompted SAM~2 and UniSlot on YouTube-VIS 2021. Both methods use the same evaluation videos, resolution, and metrics. \#Segments denotes the average number of predicted foreground segments per frame (SAM~2 masks or active UniSlot slots, excluding background); the ground truth contains $1.9$ annotated objects per frame.}}
    \label{tab:sam2}
    \small
    \setlength{\tabcolsep}{4pt}
    \begin{tabular}{lccccc}
    \toprule
    \textbf{Method} & \textbf{ARI} & \textbf{FG-ARI} & \textbf{mBO} & \textbf{J\&F} & \textbf{\#Segments} \\
    \midrule
    SAM 2 (unprompted) & \underline{.227} & \underline{.218} & \textbf{.570} & \textbf{.511} & \underline{19.1} \\
    \rowcolor{gray!15}
    UniSlot (Ours)     & \textbf{.228} & \textbf{.231} & \underline{.455} & \underline{.315} & \textbf{2.9} \\
    \bottomrule
    \end{tabular}
\end{table}

\paragraph{Accurate masks do not imply object-centric decomposition.}
As shown in \cref{tab:sam2}, UniSlot achieves slightly better object grouping than SAM~2 (ARI: $0.228$ vs.\ $0.227$; FG-ARI: $0.231$ vs.\ $0.218$) while producing $6.6\times$ fewer segments. SAM~2 predicts $19.1$ segments per frame for only $1.9$ annotated objects, frequently splitting objects into parts and fragmenting textured backgrounds (\cref{fig:sam2_examples}). This over-segmentation presents a fundamental challenge for constructing object-centric representations: identifying meaningful objects, merging fragmented regions, and maintaining consistent object-level identities across time require additional processing beyond mask propagation. Simply pooling features within the predicted masks would therefore yield numerous representations that do not necessarily correspond to meaningful objects.

\begin{figure}[h!]
    \centering
    \includegraphics[width=\linewidth]{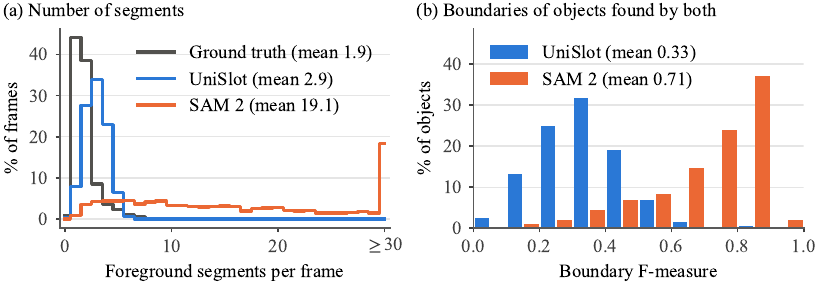}\\
    \vspace{-0.3cm}
    \caption{\textcolor{black}{Segmentation statistics on YouTube-VIS 2021 (849 annotated objects, 13{,}075 frames). (a)~Distribution of foreground segments per frame for SAM~2, UniSlot, and ground truth. (b)~Boundary F-measure for the objects detected by both methods (best-matching segment with IoU~$\geq0.5$).}}
    \label{fig:sam2_stats}
\end{figure}

UniSlot instead learns a compact decomposition into object-level slots with temporally consistent identities, without mask supervision. SAM~2 achieves higher mBO ($0.570$ vs.\ $0.455$) and J\&F ($0.511$ vs.\ $0.315$), largely reflecting its sharper segmentation boundaries. Indeed, among the $205$ objects detected by both methods, SAM~2 achieves a boundary F-measure of $0.71$, compared with $0.33$ for UniSlot (\cref{fig:sam2_stats,fig:sam2_boundaries}). While SAM 2 excels at delineating and tracking image regions, UniSlot learns to discover and track a compact set of object-level representations, providing the abstraction needed for object-centric dynamics modeling.

\paragraph{Can SAM~2 improve UniSlot?}
We additionally investigate whether SAM~2 masks can serve as spatial priors for slot initialization. At the first frame of each clip, we initialize UniSlot's slot attention using the $K-1$ largest unprompted SAM~2 masks covering $0.5$--$40\%$ of the image, assigning the remaining pixels to the background slot. The masks are pooled to the feature grid, and slot positions and scales are initialized accordingly. Subsequent frames are processed by UniSlot without additional SAM~2 masks. We apply this initialization during both training and evaluation, keeping the remaining architecture and training procedure unchanged, except for using center crops to align the masks with the input frames.

\begin{table}[h]
    \centering
    \caption{\textcolor{black}{Effect of initializing UniSlot's first-frame slots with unprompted SAM~2 masks on YouTube-VIS 2021. SAM~2 masks are used during training and inference; subsequent frames are processed by UniSlot without additional masks.}}
    \label{tab:sam2_init}
    \small
    \setlength{\tabcolsep}{4pt}
    \begin{tabular}{lccccc}
    \toprule
    \textbf{Method} & \textbf{ARI} & \textbf{FG-ARI} & \textbf{mBO} & \textbf{J\&F} & \textbf{\#Segments} \\
    \midrule
    SAM 2 (unprompted)        & .227 & \underline{.218} & \textbf{.570} & \textbf{.511} & 19.1\\
    \rowcolor{gray!15}
    UniSlot (Ours)            & \underline{.228} & \textbf{.231} & .455 & .315 & \textbf{2.9}\\
    UniSlot + SAM~2 init.     & \textbf{.232} & .215 & \underline{.459} & \underline{.321} & \underline{3.4}\\
    \bottomrule
    \end{tabular}
\end{table}

SAM~2 initialization yields a modest improvement over UniSlot, suggesting that high-quality segmentation priors can complement learned object-centric representations. However, this variant requires running a mask-supervised segmentation model during both training and inference, whereas standard UniSlot discovers its slots without segmentation annotations or mask-based initialization.

\paragraph{Why learn object-centric representations?}
Pooling pretrained features within SAM~2 masks is a plausible alternative, but does not resolve the need for object selection, grouping, and temporally consistent representations. UniSlot instead jointly learns object decomposition and structured latent states (appearance, position, scale, and presence) that support object-level dynamics prediction (\cref{subsec:results_dynamics}). While segmentation foundation models can provide useful spatial priors, they do not replace the need for learning object-centric representations.

\begin{figure}[h!]
    \centering
    \includegraphics[width=\linewidth]{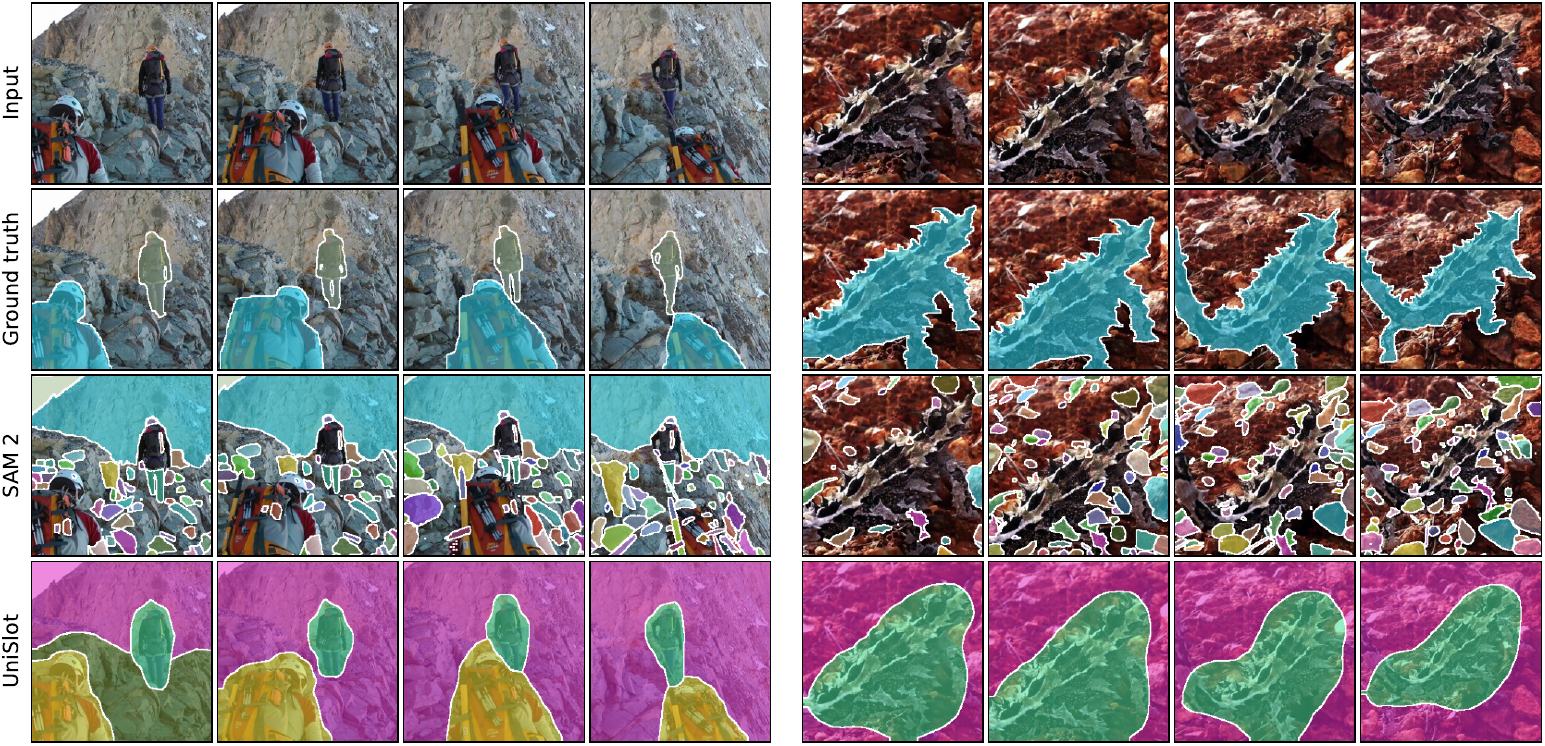}\\
    \vspace{0.4cm}
    \includegraphics[width=\linewidth]{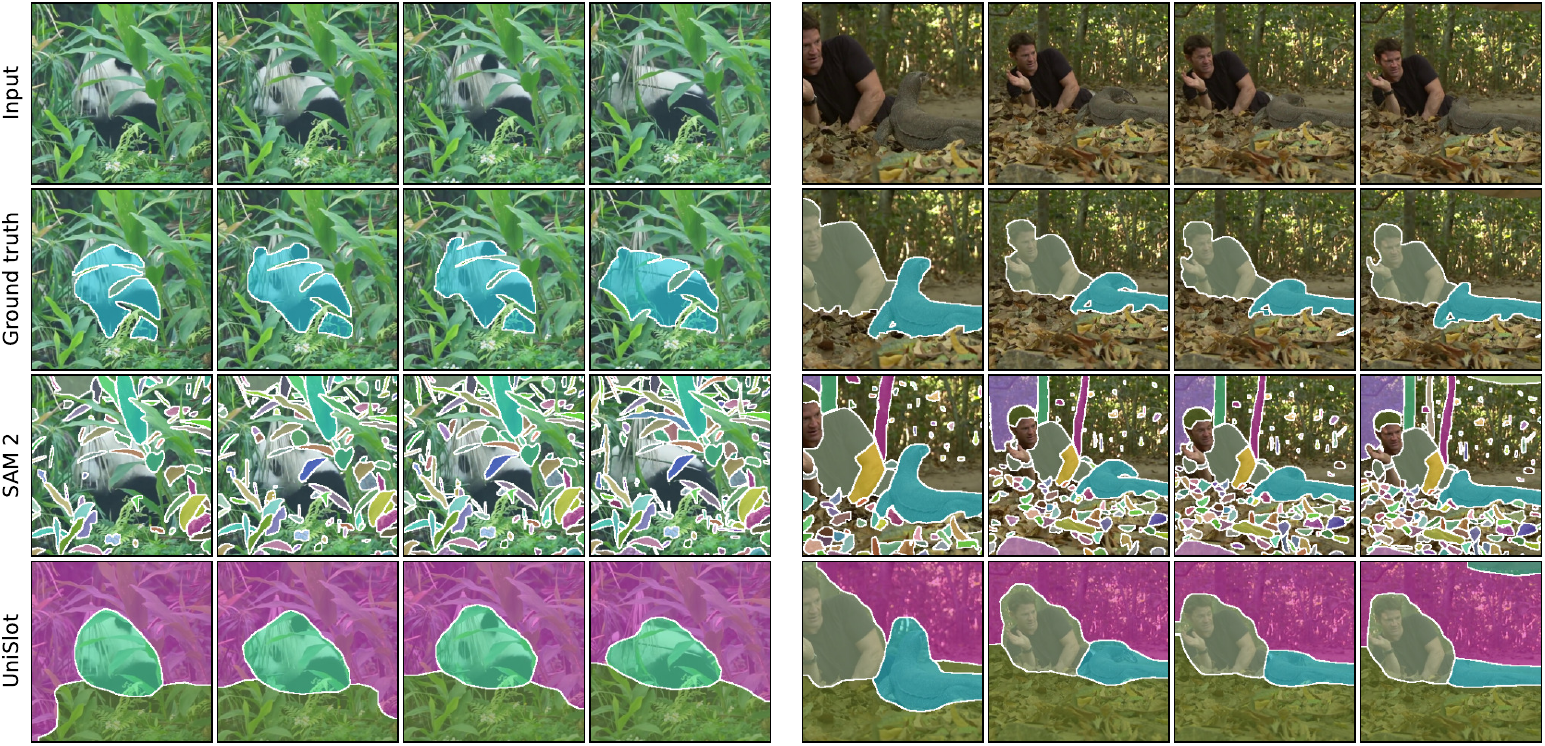}\\
    \vspace{-0.3cm}
    \caption{\textcolor{black}{Unprompted SAM~2 and UniSlot on two YouTube-VIS 2021 videos (four frames each). Left: two climbers against a textured rock face (115 SAM~2 segments over the video). Right: a thorny devil among rocks (137 SAM~2 segments over the video).}}
    \label{fig:sam2_examples}
\end{figure}

\begin{figure}[h!]
    \centering
    \includegraphics[width=\linewidth]{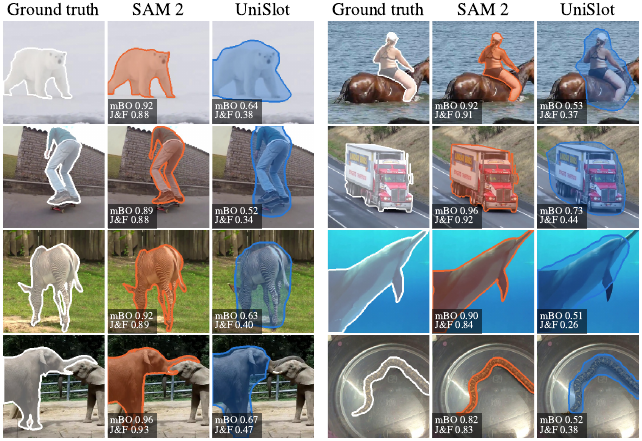}\\
    \vspace{0.3cm}
    \includegraphics[width=\linewidth]{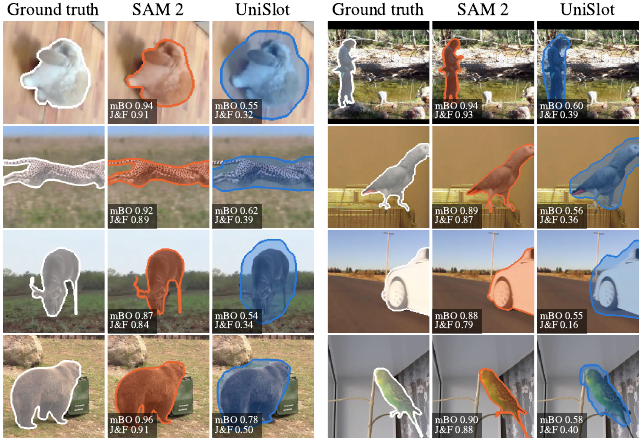}\\
    \vspace{-0.3cm}
    \caption{\textcolor{black}{Boundary comparison on objects detected by both methods. Each annotated object (white) is shown alongside its best-overlapping SAM~2 mask (orange) and UniSlot slot (blue), cropped around the object in its largest frame. Numbers report object-level mBO and J\&F over the video; J\&F uses one-to-one Hungarian matching.}}
    \label{fig:sam2_boundaries}
\end{figure}
}

\section{Additional ablations and analysis} \label{app:additional_ablations}

\subsection{Feature encoder} \label{app:feature_encoder}

\paragraph{Learnable CNN encoder}
The main experiments use frozen DINO features as the visual input to UniSlot. We additionally evaluate whether the proposed representation can be learned without a pretrained visual encoder by replacing DINO with a lightweight four-layer CNN trained jointly with UniSlot. We evaluate this setting on CLEVRER and MOVi-A, where appearance is sufficiently simple to learn visual features from scratch. The CNN produces dense features with the same dimensionality as the projected DINO features used by the grouping module.

\begin{table}[ht]
\vspace{-0.4cm}
\caption{Robustness to a learnable visual encoder. Object decomposition and tracking performance when replacing the frozen DINOv1 encoder with a four-layer CNN trained jointly with UniSlot.}
\vspace{2pt}
\footnotesize
\label{tab:cnn_encoder}
\centering
\begin{tabular}{lcccccc}
\toprule
\multirow{2}{*}{\textbf{Encoder}} & \multicolumn{3}{c}{\textbf{CLEVRER}} & \multicolumn{3}{c}{\textbf{MOVi-A}} \\
\cmidrule(lr){2-4} \cmidrule(lr){5-7}
 & \textbf{ARI}$\uparrow$ & \textbf{FG-ARI}$\uparrow$ & \textbf{mBO}$\uparrow$
 & \textbf{ARI}$\uparrow$ & \textbf{FG-ARI}$\uparrow$ & \textbf{mBO}$\uparrow$ \\
\midrule
DINOv1        & \textbf{.927} & \textbf{.781} & \textbf{.647}
              & \textbf{.882} & .740 & \textbf{.599} \\
Learnable CNN & .863 & .741 & .620
              & .818 & \textbf{.747} & .562 \\
\bottomrule
\end{tabular}
\end{table}

\begin{figure}[H]
    \vspace{-0.4cm}
    \centering
    \caption{Effect of the visual encoder. UniSlot produces similar object decompositions with a lightweight
    learnable CNN and a frozen DINOv1 encoder on CLEVRER.}
    \includegraphics[width=1\linewidth]{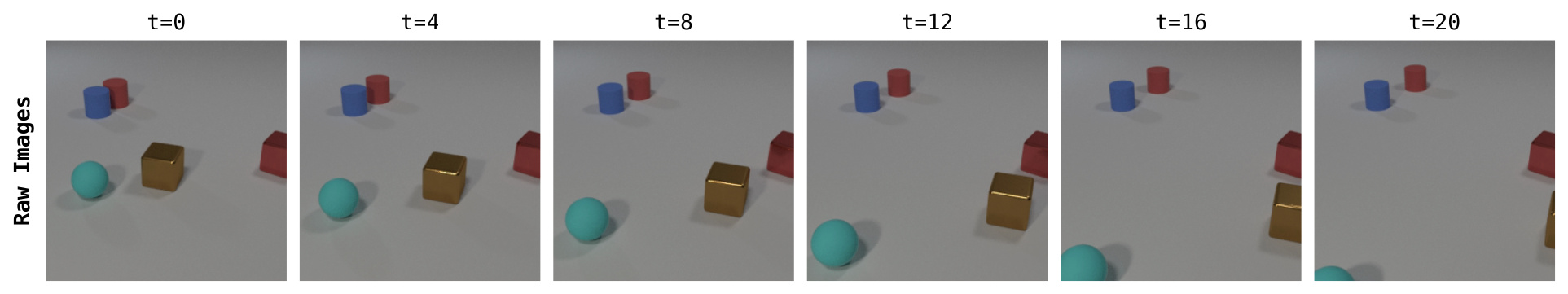}\\ \vspace{-0.2cm}
    \includegraphics[width=1\linewidth]{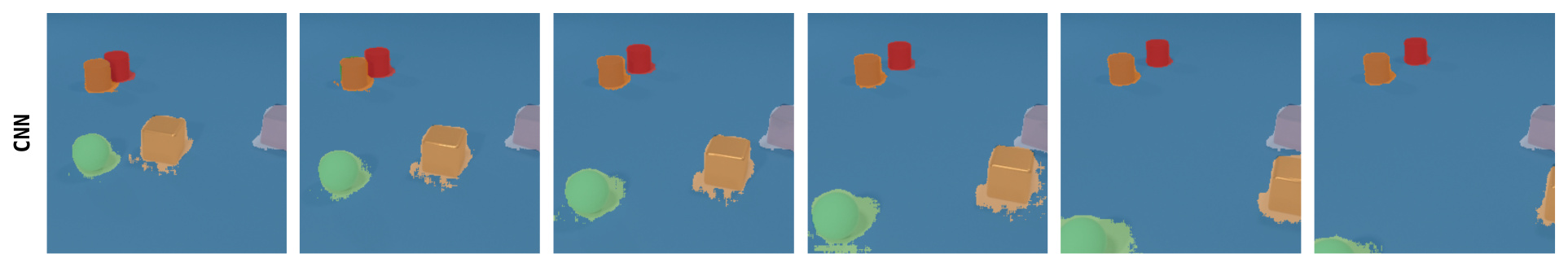}\\ \vspace{-0.2cm}
    \includegraphics[width=1\linewidth]{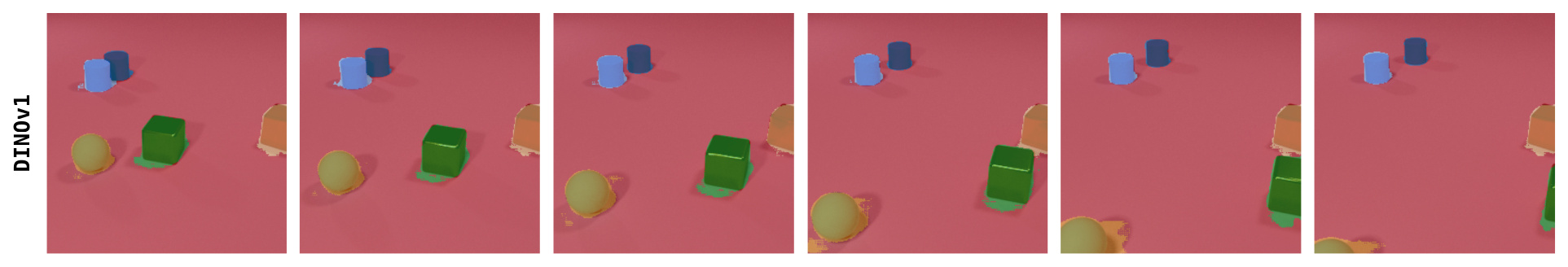}\\ \vspace{-0.3cm}
    \label{fig:cnn_encoder}
\end{figure}

As shown in \cref{tab:cnn_encoder,fig:cnn_encoder}, replacing DINOv1 with a learnable CNN leads to only a small decrease on CLEVRER and a larger but still moderate decrease on MOVi-A. This shows that UniSlot can learn object-centric representations from scratch in visually simple settings, while pretrained features provide a stronger visual signal.

\begin{figure}[h!]
    \centering
    \includegraphics[width=1\linewidth]{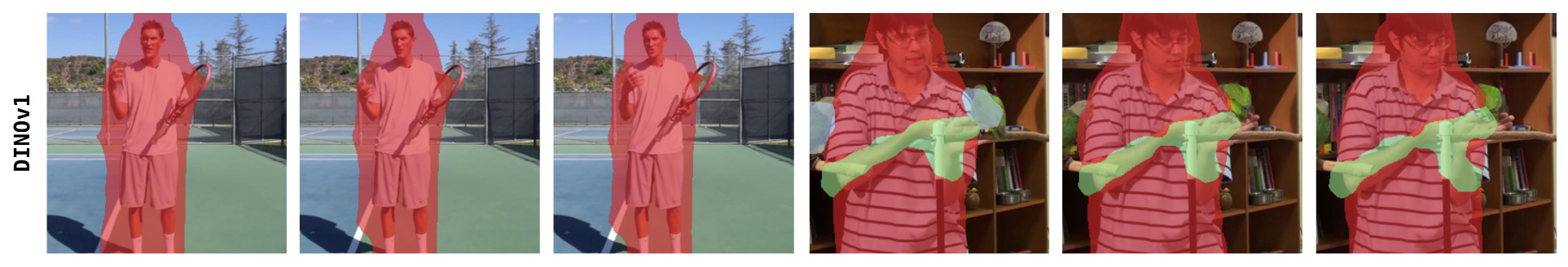}\\ \vspace{-0.2cm}
    \includegraphics[width=1\linewidth]{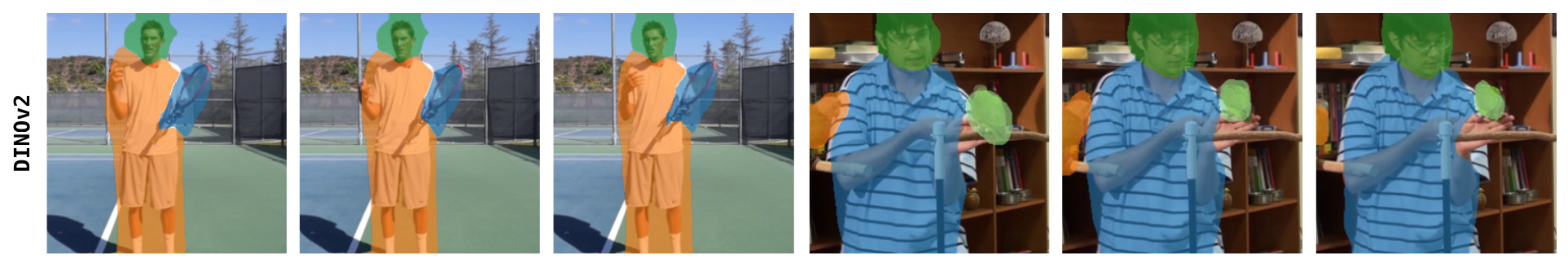}\\ \vspace{-0.2cm}
    \includegraphics[width=1\linewidth]{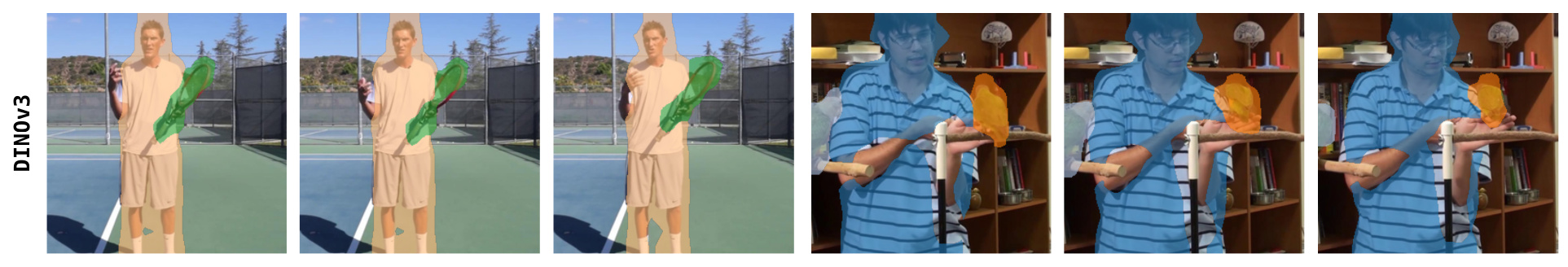}\\ \vspace{-0.4cm}
    \caption{Object tracking on the same YouTube-VIS frames using different pretrained encoders. Despite changes in the distribution of input features, UniSlot produces consistent object masks.}
    \label{fig:robustness_encoder_ytvis}
\end{figure}

\begin{figure}[h!]
    \centering
    \includegraphics[width=1\linewidth]{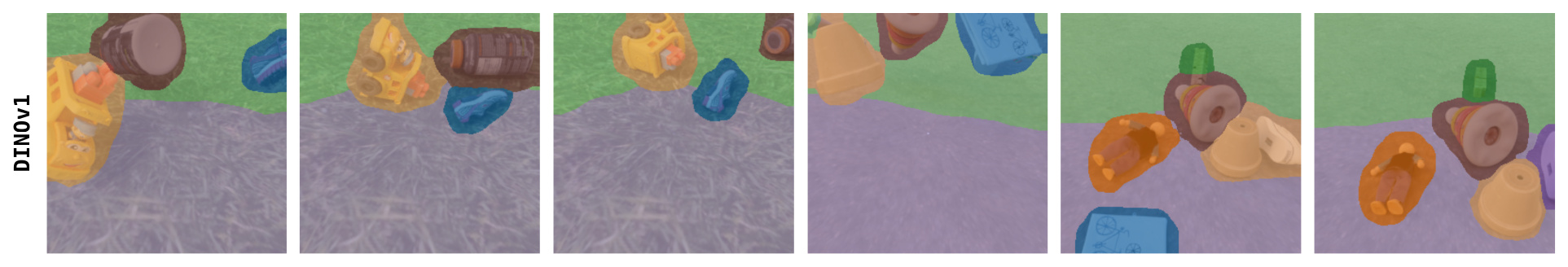}\\ \vspace{-0.2cm}
    \includegraphics[width=1\linewidth]{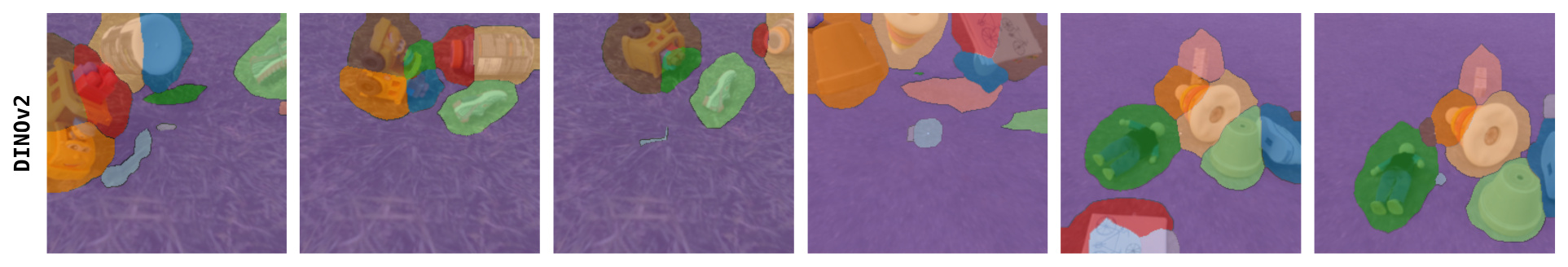}\\ \vspace{-0.2cm}
    \includegraphics[width=1\linewidth]{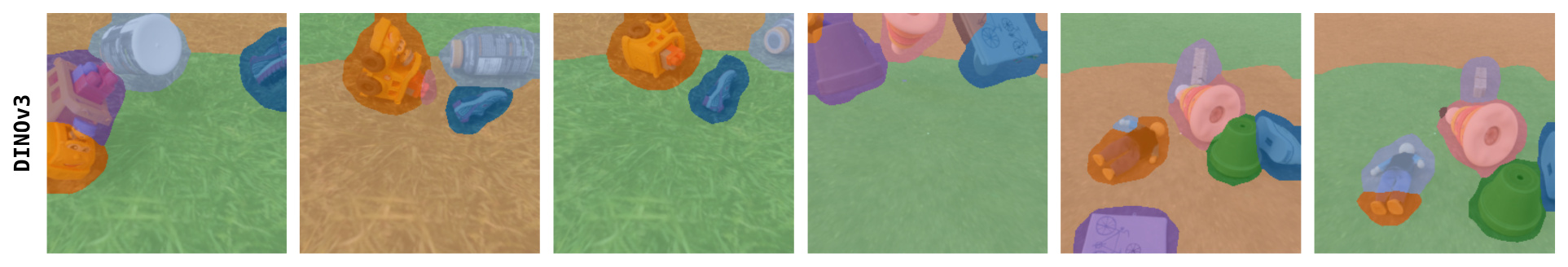}\\ \vspace{-0.4cm}
    \caption{Object tracking on the same MOVi-C frames using different pretrained encoders. Despite changes in the distribution of input features, UniSlot produces consistent object masks.}
    \label{fig:robustness_encoder_movic}
\end{figure}

\paragraph{Robustness to pretrained visual features}
We further evaluate whether UniSlot is sensitive to the particular pretrained visual representation by replacing DINOv1 with DINOv2 and DINOv3 while keeping the remaining model unchanged. Quantitative results are reported in \cref{tab:ablation_encoder}, with corresponding qualitative comparisons in \cref{fig:robustness_encoder_ytvis,fig:robustness_encoder_movic}.

\begin{table}[ht]
\vspace{-0.4cm}
\caption{Robustness to pretrained visual features. Object decomposition and tracking performance of UniSlot with different DINO encoders, evaluated on full video sequences. The DINOv1 configuration used in the main experiments is highlighted in gray.}
\label{tab:ablation_encoder}
\centering
\scriptsize
\vspace{2pt}
\begin{tabular}{lccccccccc}
\toprule
\multirow{2}{*}{\textbf{Encoder}} & \multicolumn{3}{c}{\textbf{CLEVRER}} & \multicolumn{3}{c}{\textbf{MOVi-C}} & \multicolumn{3}{c}{\textbf{YouTube-VIS 2021}} \\
\cmidrule(lr){2-4} \cmidrule(lr){5-7} \cmidrule(lr){8-10}
 & \textbf{ARI}$\uparrow$ & \textbf{FG-ARI}$\uparrow$ & \textbf{mBO}$\uparrow$
 & \textbf{ARI}$\uparrow$ & \textbf{FG-ARI}$\uparrow$ & \textbf{mBO}$\uparrow$
 & \textbf{ARI}$\uparrow$ & \textbf{FG-ARI}$\uparrow$ & \textbf{mBO}$\uparrow$ \\
\midrule
\rowcolor{lightgray!30}
DINOv1~\citep{dino}
    & \textbf{.927} & \textbf{.781} & \underline{.647}
    & \textbf{.505} & \underline{.654} & \underline{.358}
    & \textbf{.228} & \underline{.231} & \textbf{.455} \\
DINOv2~\citep{darcet2024vision}
    & .852 & .743 & .639
    & .420 & .532 & .313
    & \underline{.203} & .212 & .345 \\
DINOv3~\citep{dinov3}
    & \underline{.904} & \underline{.762} & \textbf{.676}
    & \underline{.451} & \textbf{.679} & \textbf{.365}
    & .200 & \textbf{.281} & \underline{.432} \\
\bottomrule
\end{tabular}
\vspace{-0.5cm}
\end{table}

UniSlot produces consistent object decompositions across all three feature encoders, with relatively small variations across most datasets and metrics. No encoder dominates across all settings: DINOv3 improves FG-ARI and mBO on several benchmarks, while DINOv1 remains strongest in ARI on MOVi-C and YouTube-VIS. The qualitative results show a similar pattern: all three encoders recover comparable object assignments and maintain slot identities over time, while differing in fine-grained segmentation behavior. On YouTube-VIS, DINOv2 occasionally separates object parts into different slots, whereas DINOv3 can produce more semantically coherent groupings, such as separating a tennis racket from the player or a bird from background clutter. On MOVi-C, DINOv1 produces more stable object decompositions, consistent with its quantitative performance.

Overall, these results show that UniSlot remains effective across substantially different pretrained visual features, although the choice of encoder influences the resulting object boundaries and granularity. We use DINOv1 in the main experiments to maintain comparability with prior video object-centric methods.

\newpage
\subsection{Alternative decoder architecture} \label{app:decoder_architecture}

We additionally examine whether the disentangled slot representation can be used with an alternative decoder architecture. In this experiment, conducted without pseudo-depth supervision, we replace the Spatial Broadcast Decoder (SBD)~\citep{watters2019spatial} with the per-slot MLP decoder commonly used in feature-based object-centric models~\citep{dinosaur,slot_contrast}.

The MLP independently decodes each slot into the visual target $\hat y_t^k$ and alpha mask $\alpha_t^k$. Instead of shared positional encodings, we use slot-relative coordinates derived from $(p_t^k,s_t^k)$, preserving the separation between appearance and geometry. We use a four-layer MLP with hidden dimensions $768$ and $1024$ for MOVi-E and YouTube-VIS, respectively.

\begin{table}[ht]
\vspace{-0.4cm}
\caption{Robustness to decoder architecture. Object tracking performance of UniSlot with Spatial Broadcast and per-slot MLP decoders on MOVi-E and YouTube-VIS 2021.}
\label{tab:mlp_decoder}
\centering
\small
\vspace{2pt}
\begin{tabular}{lcccccc}
\toprule
\multirow{2}{*}{\textbf{Decoder}}
& \multicolumn{3}{c}{\textbf{MOVi-E}}
& \multicolumn{3}{c}{\textbf{YouTube-VIS 2021}} \\
\cmidrule(lr){2-4} \cmidrule(lr){5-7}
& \textbf{ARI}$\uparrow$
& \textbf{FG-ARI}$\uparrow$
& \textbf{mBO}$\uparrow$
& \textbf{ARI}$\uparrow$
& \textbf{FG-ARI}$\uparrow$
& \textbf{mBO}$\uparrow$ \\
\midrule
Spatial Broadcast
& \textbf{.252} & .784 & .279
& \textbf{.219} & .217 & \textbf{.444} \\
MLP
& .242 & \textbf{.795} & \textbf{.301}
& .202 & \textbf{.230} & .418 \\
\bottomrule
\end{tabular}
\end{table}

Despite the substantial architectural change, the two decoders obtain similar performance: the MLP slightly improves FG-ARI and mBO on MOVi-E, while the SBD performs better on YouTube-VIS. This experiment shows that the disentangled appearance and geometric representation can be used with different per-slot decoders. We use the more efficient SBD in the main model.

\subsection{Robustness to the number of slots} \label{app:slot_count}

We evaluate sensitivity to the maximum number of available slots $K$. With adaptive activation, $K$ defines an upper bound on the number of represented objects, while unnecessary slots can be deactivated. The binary gate uses hard Gumbel--Softmax with a fixed decision threshold of $0.5$.

\Cref{tab:slot_sensitivity_supp} compares UniSlot with VideoSAUR and SlotContrast as $K$ increases beyond the default setting. UniSlot is significantly less sensitive to excess slot capacity, particularly in FG-ARI and mBO. On MOVi-C, these metrics remain stable across $K=11$--$20$, whereas SlotContrast collapses at $K=20$. A similar trend is observed on YouTube-VIS 2021, where the baselines degrade more substantially as $K$ increases.

\begin{table}[h]
    \vspace{-0.4cm}
    \centering
    \caption{Robustness to the maximum number of slots $K$. Sequence-level ARI/FG-ARI/mBO are reported for increasing slot capacity. The smallest $K$ corresponds to the default configuration used in the main experiments.}
    \small
    \vspace{2pt}
    \begin{tabular}{@{}lcccccc@{}}
        \toprule
        \multirow{2}{*}{\textbf{Method}} &
        \multicolumn{3}{c}{\textbf{MOVi-C}} &
        \multicolumn{3}{c}{\textbf{YTVIS 2021}} \\
        \cmidrule(lr){2-4}\cmidrule(lr){5-7}
        & $K=11$ & $K=15$ & $K=20$ & $K=7$ & $K=10$ & $K=15$ \\
        \midrule
        VideoSAUR
        & .22/.61/.33
        & .23/.55/.35
        & .18/.50/.35
        & .14/.19/.38
        & .10/.15/.31
        & .07/.12/.27 \\
        SlotContrast
        & .10/.57/.27
        & .17/.62/.27
        & .00/.01/.04
        & .14/.23/.36
        & .13/.20/.33
        & .08/.15/.27 \\
        \rowcolor{lightgray!30}UniSlot
        & \textbf{.51}/\textbf{.65}/\textbf{.36}
        & \textbf{.37}/\textbf{.64}/\textbf{.35}
        & \textbf{.28}/\textbf{.71}/\textbf{.37}
        & \textbf{.23}/\textbf{.23}/\textbf{.46}
        & \textbf{.18}/\textbf{.25}/\textbf{.43}
        & \textbf{.16}/\textbf{.21}/\textbf{.38} \\
        \bottomrule
    \end{tabular}
    \label{tab:slot_sensitivity_supp}
\end{table}

{\color{black}
\subsection{Sensitivity to depth map quality} \label{app:depth_robustness}

UniSlot uses monocular depth estimated by Depth Anything~3 as an auxiliary reconstruction target, and such estimates can be unreliable, particularly for thin, transparent, or reflective objects. Since depth is used only during training, errors in the depth estimates of test videos do not directly affect inference. Instead, inaccurate supervision influences the representations learned during training. To assess this sensitivity, we retrain UniSlot on YT-VIS 2021 with deliberately degraded depth targets, keeping all other training settings unchanged.

We consider two controlled perturbations, illustrated in \cref{fig:depth_perturbations}. First, Gaussian blur ($\sigma=4$ pixels at $256\times256$ resolution) removes sharp depth discontinuities at object boundaries. Second, a smooth spatial distortion introduces slowly varying depth offsets, altering relative depth relationships between regions while approximately preserving local boundaries. The maximum offset is half the video's depth range, with a new perturbation sampled for each training clip.

\begin{figure}[t]
    \centering
    \includegraphics[width=0.8\linewidth]{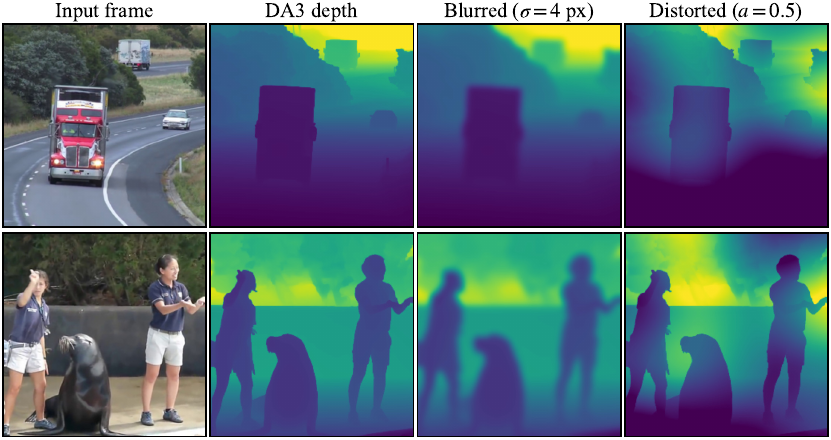}\\
    \vspace{-0.3cm}
    \caption{\textcolor{black}{Depth targets used for training on YT-VIS 2021: the Depth Anything~3 estimate (per-video normalized) and its two perturbations.}}
    \label{fig:depth_perturbations}
\end{figure}

As shown in \cref{tab:depth_robustness}, the original depth targets yield the best performance, while removing depth supervision leads to only a modest decrease (ARI $.228$ to $.219$). In contrast, substantially degraded depth targets cause considerably larger performance drops, with ARI decreasing to $.094$ under blurring and $.068$ under distortion. These results indicate that depth supervision is beneficial when its estimates are informative, but severely degraded targets can be more harmful than omitting depth entirely. Importantly, UniSlot without depth retains competitive performance, demonstrating that depth supervision is beneficial but not essential to the proposed representation.

\begin{table}[h]
    \centering
    \caption{\textcolor{black}{Sensitivity to depth target quality on YT-VIS 2021. All models share the same architecture and training schedule; only the depth targets used during training differ.}}
    \label{tab:depth_robustness}
    \small
    \begin{tabular}{lcccc}
    \toprule
    \textbf{Depth targets} & \textbf{ARI} & \textbf{FG-ARI} & \textbf{mBO} & \textbf{J\&F} \\
    \midrule
    None (UniSlot w/o depth) & \underline{.219} & \underline{.217} & \underline{.444} & \underline{.297} \\
    Blurred                  & .094 & .152 & .253 & .201 \\
    Distorted                & .068 & .137 & .231 & .183 \\
    \rowcolor{gray!15}
    Depth Anything~3 (UniSlot) & \textbf{.228} & \textbf{.231} & \textbf{.455} & \textbf{.315} \\
    \bottomrule
    \end{tabular}
\end{table}

Depth supervision can also be detrimental in individual scenes. \Cref{fig:depth_failures} shows YT-VIS 2021 validation videos where UniSlot without depth produces better object decompositions than the depth-supervised model. In the first three examples, distinct objects occupy similar or overlapping depth ranges, providing limited geometric cues for their separation. In these cases, the depth-supervised model struggles to distinguish the kayaker from the kayak, overlapping sharks, and the rider from her horse, whereas the depth-free model separates them more effectively. In the aquarium example, the predicted depth map poorly distinguishes the two turtles, potentially due to transparent surfaces and reflections. Here, the depth-supervised model merges the turtles, while the depth-free model separates them.

Since both models use only RGB inputs at inference, these differences reflect the influence of depth supervision during training. These examples suggest that depth can introduce an undesirable grouping bias when distinct objects occupy similar depths or when estimated depth boundaries are unreliable.

\begin{figure}[h]
    \centering
    \includegraphics[width=1\linewidth]{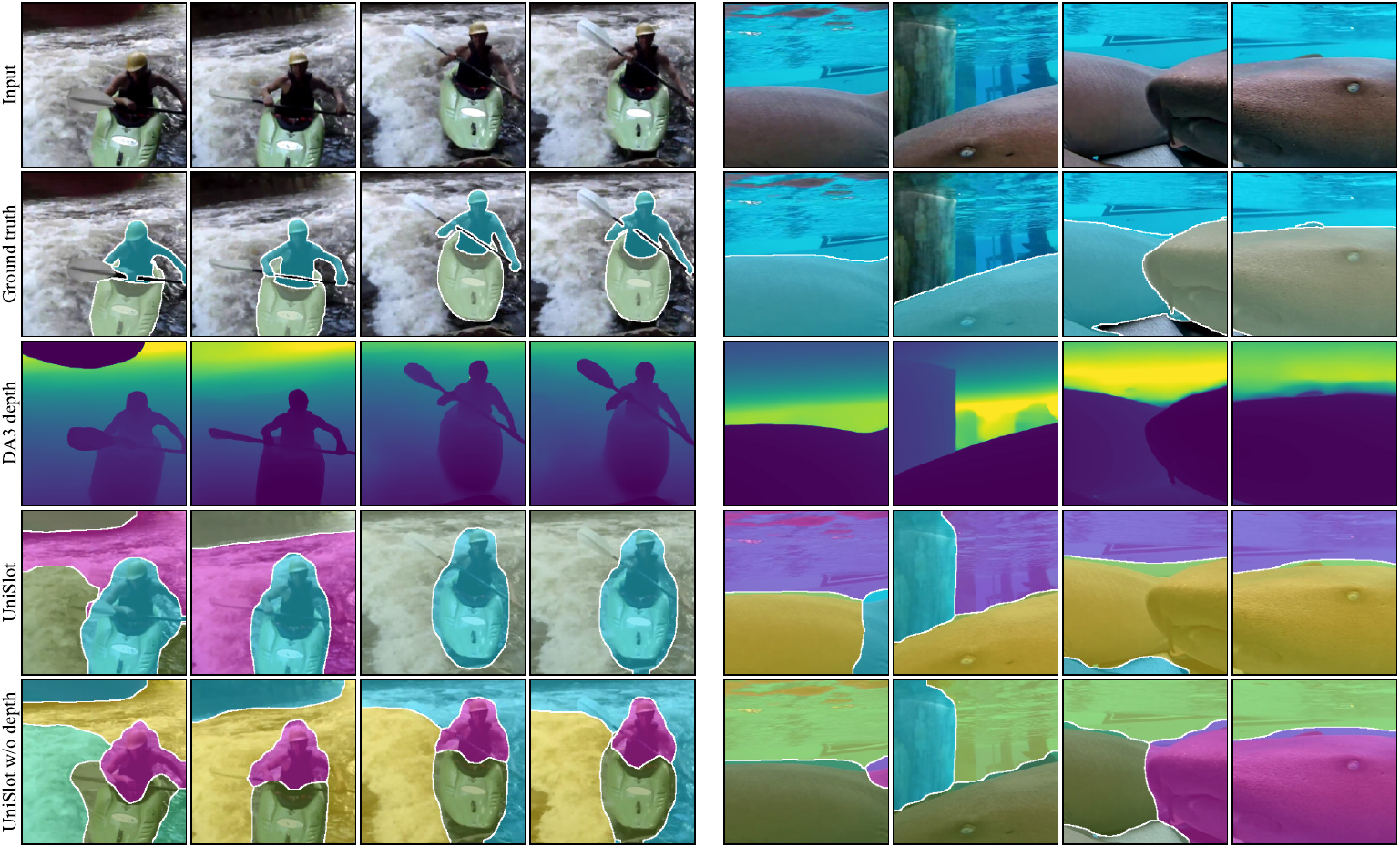}\\ \vspace{0.3cm}
    \includegraphics[width=1\linewidth]{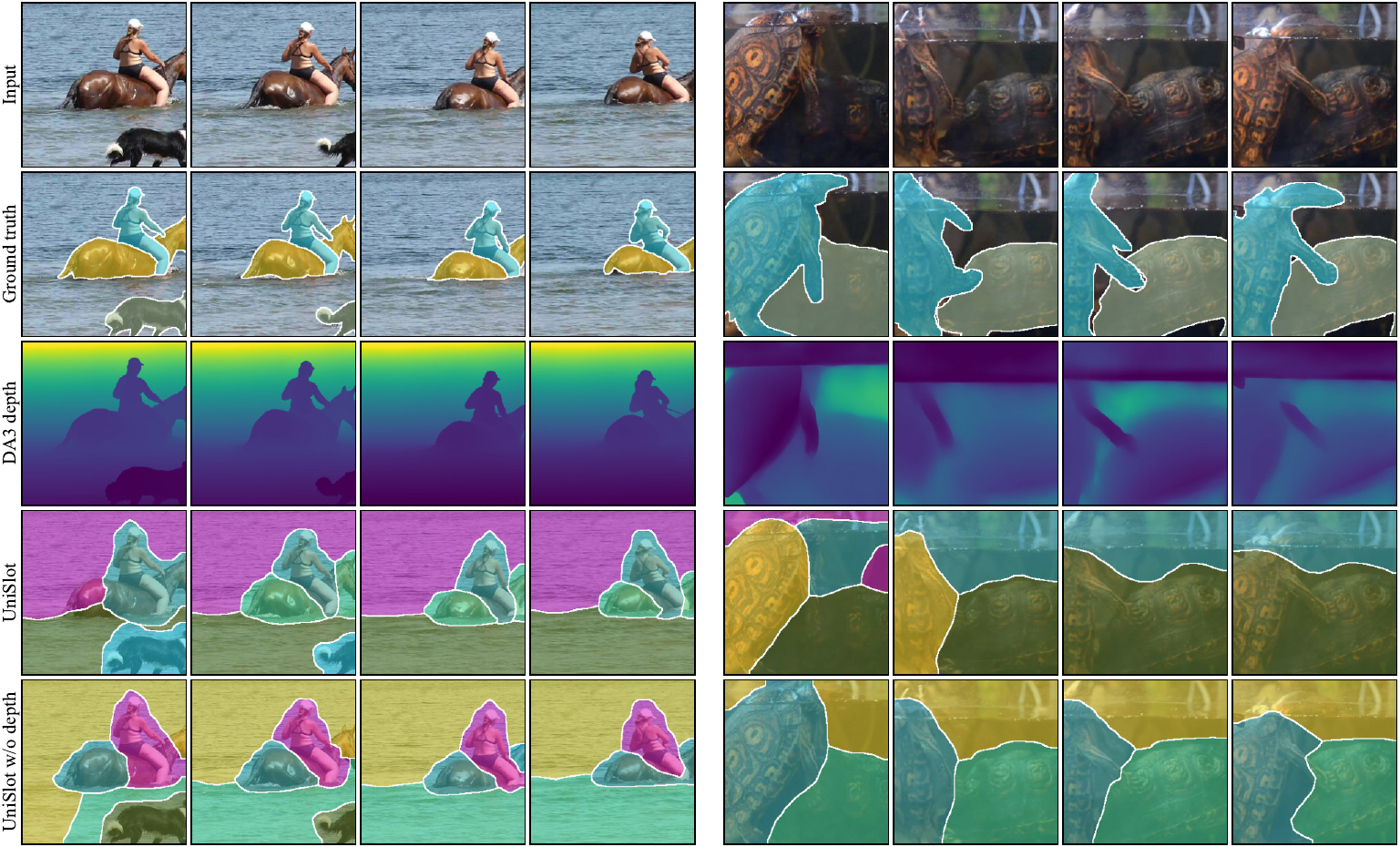}\\
    \vspace{-0.3cm}
    \caption{\textcolor{black}{YT-VIS 2021 videos where UniSlot without depth outperforms UniSlot with depth. Depth is not used at inference; the difference stems from training with depth supervision.}}
    \label{fig:depth_failures}
\end{figure}
}

\section{Additional quantitative results} \label{app:extra_quantitative}

\subsection{Per-image object discovery} \label{app:per_image_results}

The main experiments evaluate object decomposition and tracking at the sequence level, requiring consistent slot assignments across frames. We additionally report per-image results, where ARI, FG-ARI, and mBO are computed independently for each frame and then averaged over the dataset. This evaluation isolates frame-level object decomposition from temporal slot consistency and enables comparison with image-based object-centric methods---DINOSAUR~\citep{dinosaur}, ISA~\citep{isa}, and
Adaptive Slot Attention~\citep{fan2024adaptive}---that do not maintain object identities across frames.

\begin{table}[h]
\caption{Per-image object discovery. ARI, FG-ARI, and mBO are computed independently for each frame and then averaged over the dataset. We compare both image-based and video-based object-centric methods. Best and second-best results are shown in bold and underlined, respectively.}
\label{tab:image_discovery}
\vspace{2pt}
\centering
\scriptsize
\setlength{\tabcolsep}{3.5pt}

\begin{tabular}{lccccccccc}
\toprule
\multirow{2}{*}{\textbf{Method}}
& \multicolumn{3}{c}{\textbf{CLEVRER}}
& \multicolumn{3}{c}{\textbf{MOVi-C}}
& \multicolumn{3}{c}{\textbf{MOVi-E}} \\
\cmidrule(lr){2-4}
\cmidrule(lr){5-7}
\cmidrule(lr){8-10}
& ARI$\uparrow$ & FG-ARI$\uparrow$ & mBO$\uparrow$
& ARI$\uparrow$ & FG-ARI$\uparrow$ & mBO$\uparrow$
& ARI$\uparrow$ & FG-ARI$\uparrow$ & mBO$\uparrow$ \\
\midrule

\multicolumn{10}{l}{\textit{Image-based methods}} \\
DINOSAUR~\citep{dinosaur}
& .046 & \textbf{.971} & .153
& .152 & .650 & .402
& .128 & .607 & \textbf{.381} \\

ISA~\citep{isa}
& .032 & .851 & .164
& .145 & .612 & .385
& .152 & .701 & .314 \\

AdaSlot~\citep{fan2024adaptive}
& .020 & .344 & .064
& .018 & .209 & .044
& .073 & .202 & .094 \\

\midrule
\multicolumn{10}{l}{\textit{Video-based methods}} \\
SAVi~\citep{savi}
& .012 & .411 & .069
& .063 & .466 & .217
& .011 & .612 & .288 \\

VideoSAUR~\citep{videosaur}
& \underline{.372} & .281 & \underline{.184}
& \underline{.241} & \underline{.733} & \textbf{.434}
& \textbf{.253} & .667 & \underline{.331} \\

SlotContrast~\citep{slot_contrast}
& .050 & \underline{.967} & .154
& .109 & .708 & .231
& .203 & .774 & .303 \\

SRL~\citep{seong2026from}
& .084 & .891 & .112
& .057 & .310 & .078
& .148 & \underline{.803} & .284 \\

SlotCurri~\citep{moon2026reconstruction}
& .042 & .935 & .094
& .195 & .725 & .326
& .052 & .645 & .207 \\

\rowcolor{lightgray!30}
UniSlot (Ours)
& \textbf{.927} & .952 & \textbf{.741}
& \textbf{.544} & \textbf{.765} & \underline{.403}
& \underline{.243} & \textbf{.831} & .283 \\

\bottomrule
\end{tabular}

\vspace{4pt}

\begin{tabular}{lcccccc}
\toprule
\multirow{2}{*}{\textbf{Method}}
& \multicolumn{3}{c}{\textbf{YTVIS 2019}}
& \multicolumn{3}{c}{\textbf{YTVIS 2021}} \\
\cmidrule(lr){2-4}
\cmidrule(lr){5-7}
& ARI$\uparrow$ & FG-ARI$\uparrow$ & mBO$\uparrow$
& ARI$\uparrow$ & FG-ARI$\uparrow$ & mBO$\uparrow$ \\
\midrule

\multicolumn{7}{l}{\textit{Image-based methods}} \\
DINOSAUR~\citep{dinosaur}
& .201 & \underline{.264} & .524
& .175 & .215 & \underline{.502} \\

ISA~\citep{isa}
& .194 & .245 & \underline{.525}
& \underline{.183} & .231 & \textbf{.528} \\

AdaSlot~\citep{fan2024adaptive}
& .086 & .164 & .273
& .075 & .142 & .255 \\

\midrule
\multicolumn{7}{l}{\textit{Video-based methods}} \\
SAVi~\citep{savi}
& .058 & .197 & .268
& .052 & .181 & .233 \\

VideoSAUR~\citep{videosaur}
& .173 & .170 & .445
& .160 & .227 & .443 \\

SlotContrast~\citep{slot_contrast}
& .191 & .221 & .446
& .162 & \underline{.276} & .398 \\

SRL~\citep{seong2026from}
& \underline{.244} & .206 & .442
& .112 & \underline{.276} & .265 \\

SlotCurri~\citep{moon2026reconstruction}
& .102 & .164 & .290
& .073 & .252 & .293 \\

\rowcolor{lightgray!30}
UniSlot (Ours)
& \textbf{.329} & \textbf{.303} & \textbf{.537}
& \textbf{.244} & \textbf{.310} & .470 \\

\bottomrule
\end{tabular}

\end{table}

\newpage
\subsection{Additional results on MOVi-A and MOVi-B} \label{app:extra_movi_datasets}

For completeness, we additionally evaluate UniSlot on MOVi-A and MOVi-B, which provide simpler variants of the MOVi benchmark than MOVi-C and MOVi-E used in the main evaluation. \Cref{tab:extra_movi_datasets} reports sequence-level ARI, FG-ARI, and mBO, using the same evaluation protocol as in the main paper.

\begin{table}[h]
\caption{Additional results on MOVi-A and MOVi-B. Sequence-level object decomposition and tracking performance, evaluated using the same protocol as in the main experiments. Best and second-best results are shown in bold and underlined, respectively.}
\label{tab:extra_movi_datasets}
\vspace{2pt}
\centering
\scriptsize
\setlength{\tabcolsep}{3.5pt}
\begin{tabular}{lcccccc}
\toprule
\multirow{2}{*}{\textbf{Method}}
& \multicolumn{3}{c}{\textbf{MOVi-A}}
& \multicolumn{3}{c}{\textbf{MOVi-B}} \\
\cmidrule(lr){2-4}
\cmidrule(lr){5-7}
& ARI$\uparrow$ & FG-ARI$\uparrow$ & mBO$\uparrow$
& ARI$\uparrow$ & FG-ARI$\uparrow$ & mBO$\uparrow$ \\
\midrule
SAVi~\citep{savi}
& .004 & .083 & .031
& .139 & .265 & .152 \\

VideoSAUR~\citep{videosaur}
& \underline{.192} & .304 & .145
& \underline{.181} & .269 & .143 \\

SlotContrast~\citep{slot_contrast}
& .177 & \textbf{.905} & \underline{.194}
& .148 & \textbf{.826} & \underline{.268} \\

\rowcolor{lightgray!30}
UniSlot (Ours)
& \textbf{.882} & \underline{.740} & \textbf{.599}
& \textbf{.732} & \underline{.544} & \textbf{.502} \\

\bottomrule
\end{tabular}
\end{table}

\section{Additional qualitative results} \label{app:extra_qualitative}

We provide additional qualitative comparisons across synthetic and real-world benchmarks to complement the results in the main paper. The examples illustrate object decomposition and tracking over time and compare UniSlot with the main video-based baselines.

\begin{figure}[h]
    \centering
    \caption{Additional qualitative results on YouTube-VIS 2021 (part 1). Colors indicate slot assignments and are kept consistent across frames to visualize temporal object tracking.}
    \includegraphics[width=1\linewidth]{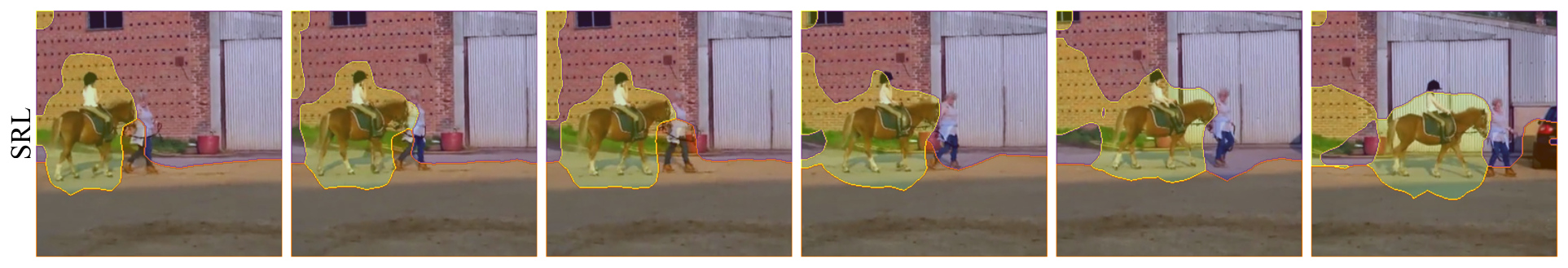}\\ \vspace{-0.2cm}
    \includegraphics[width=1\linewidth]{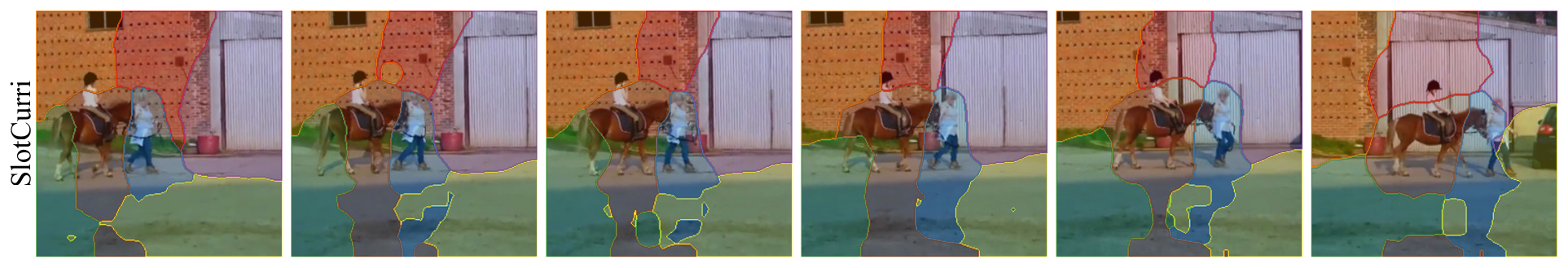}\\ \vspace{-0.2cm}
    \includegraphics[width=1\linewidth]{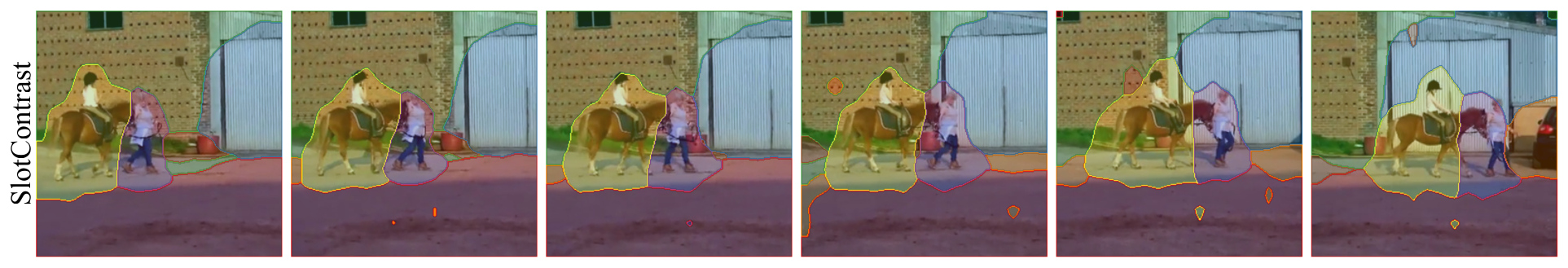}\\ \vspace{-0.2cm}
    \includegraphics[width=1\linewidth]{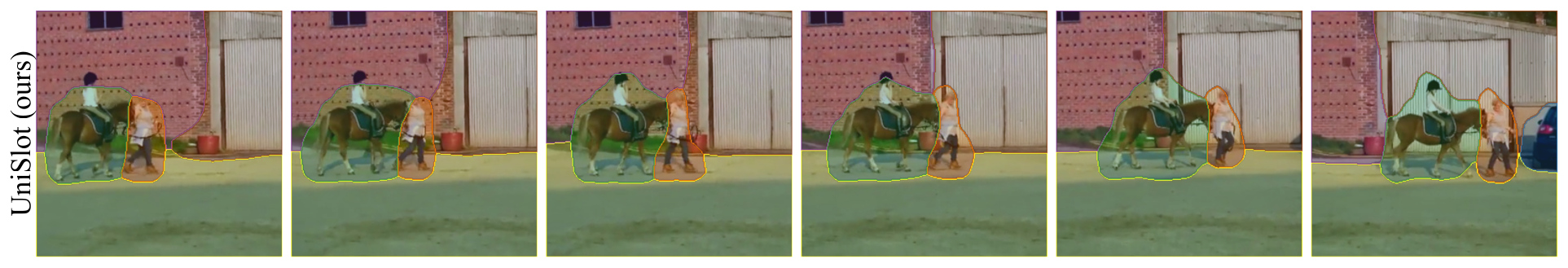}\\ \vspace{0.4cm}
    \includegraphics[width=1\linewidth]{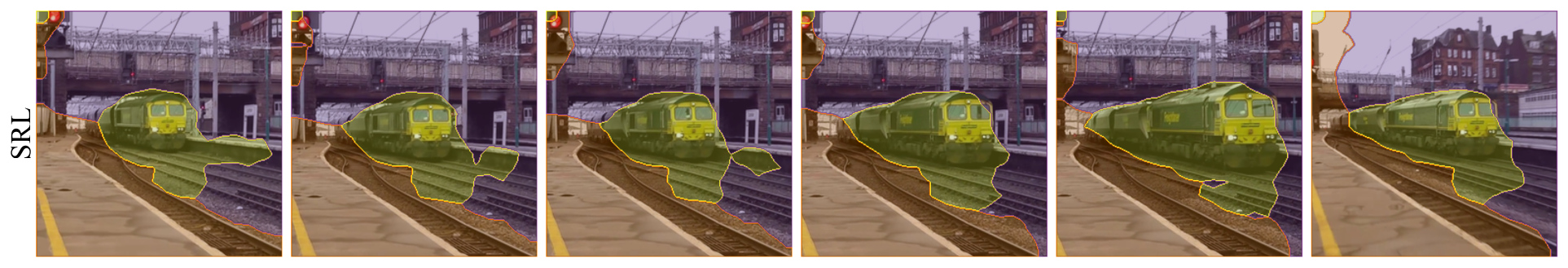}\\ \vspace{-0.2cm}
    \includegraphics[width=1\linewidth]{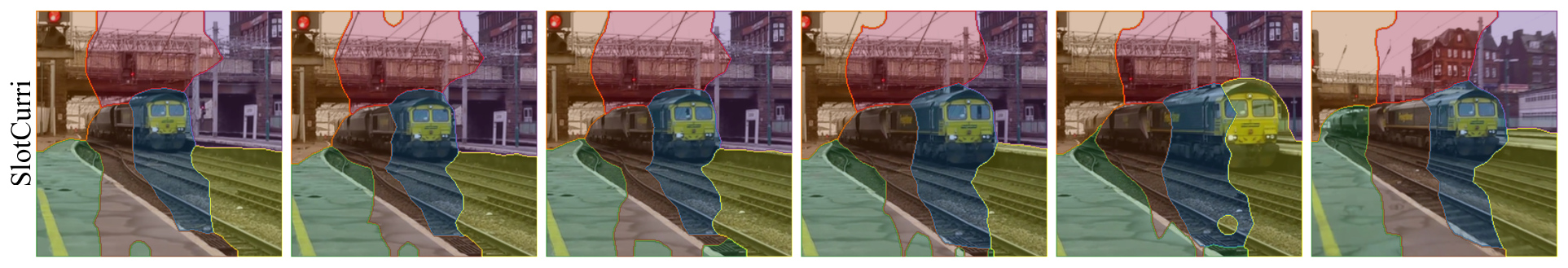}\\ \vspace{-0.2cm}
    \includegraphics[width=1\linewidth]{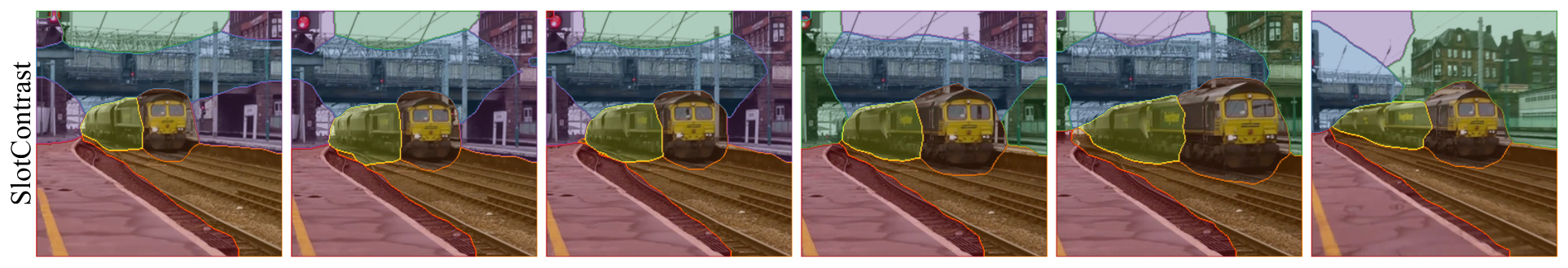}\\ \vspace{-0.2cm}
    \includegraphics[width=1\linewidth]{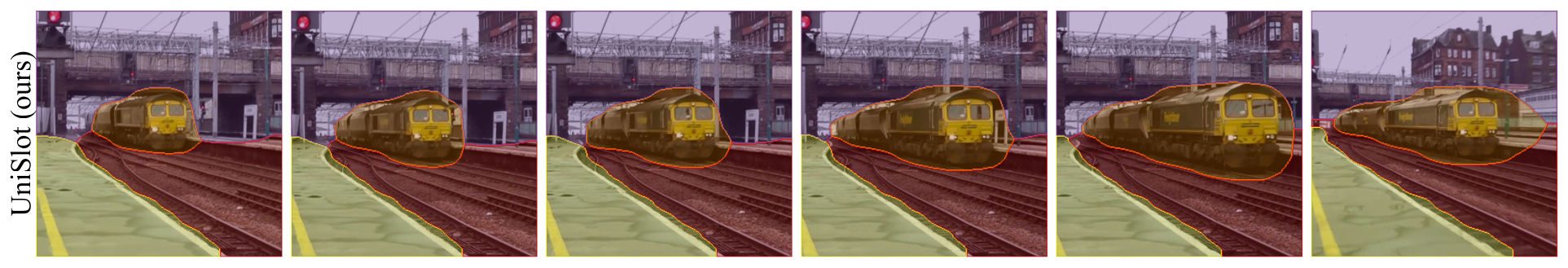}\\
    \label{fig:extra_qualitative_ytvis2021_part1}
\end{figure}

\begin{figure}[h]
    \centering
    \caption{Additional qualitative results on YouTube-VIS 2021 (part 2). Colors indicate slot assignments and are kept consistent across frames to visualize temporal object tracking.}
    \includegraphics[width=1\linewidth]{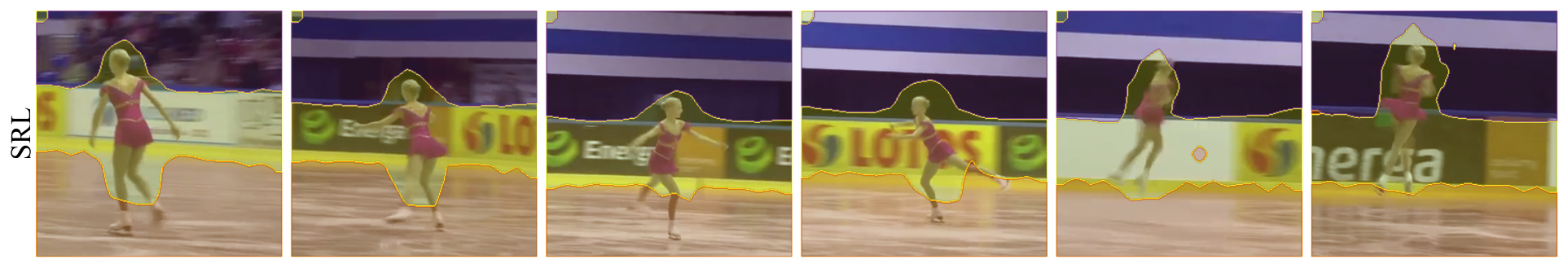}\\ \vspace{-0.2cm}
    \includegraphics[width=1\linewidth]{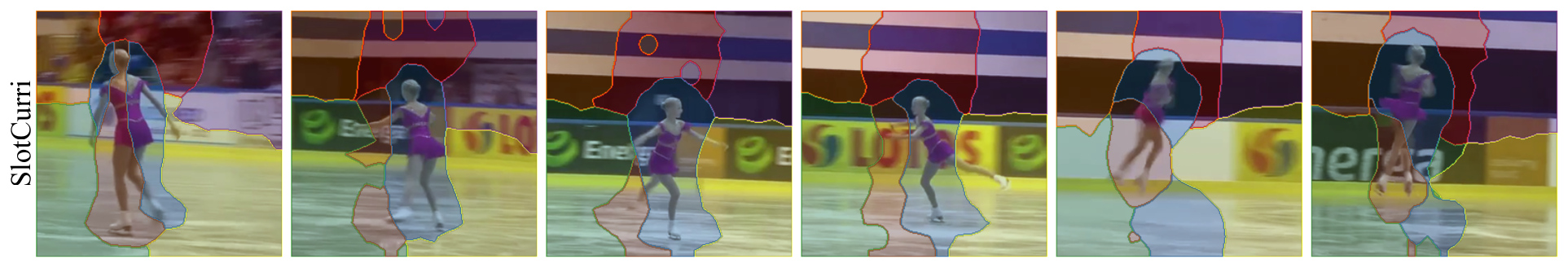}\\ \vspace{-0.2cm}
    \includegraphics[width=1\linewidth]{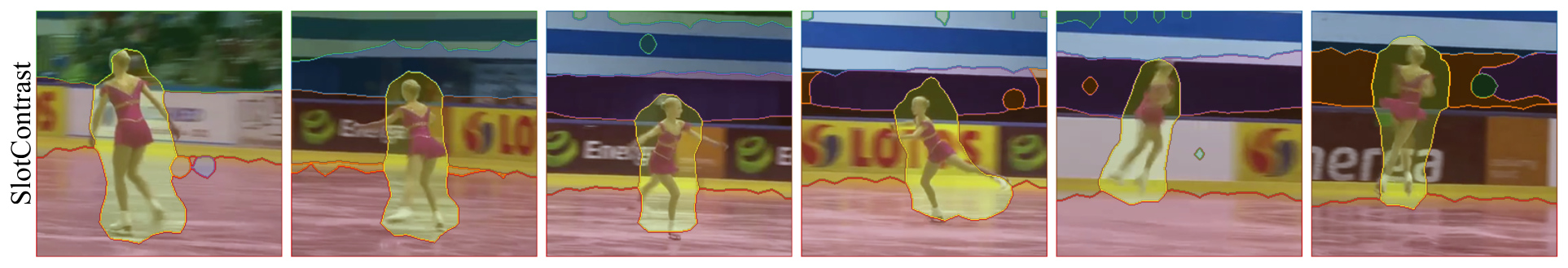}\\ \vspace{-0.2cm}
    \includegraphics[width=1\linewidth]{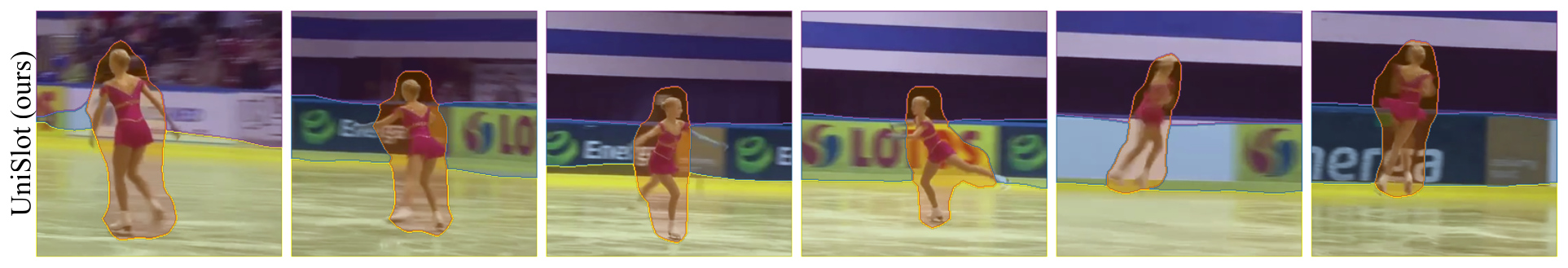}\\ \vspace{0.4cm}
    \includegraphics[width=1\linewidth]{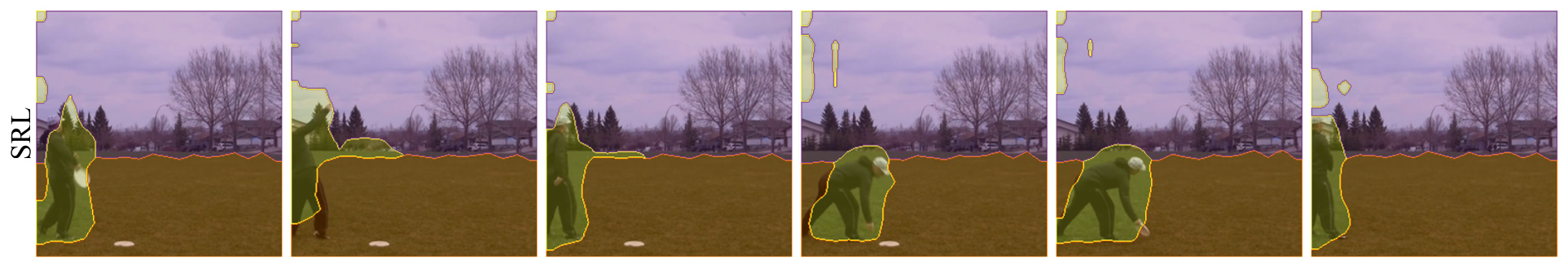}\\ \vspace{-0.2cm}
    \includegraphics[width=1\linewidth]{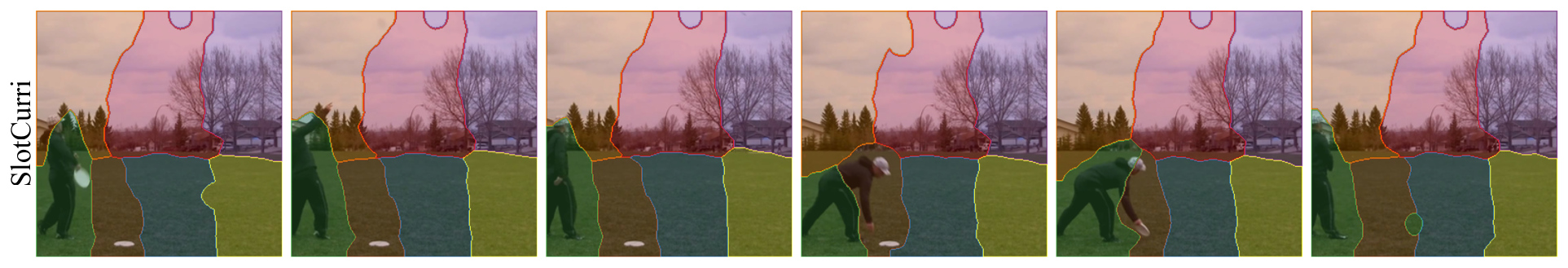}\\ \vspace{-0.2cm}
    \includegraphics[width=1\linewidth]{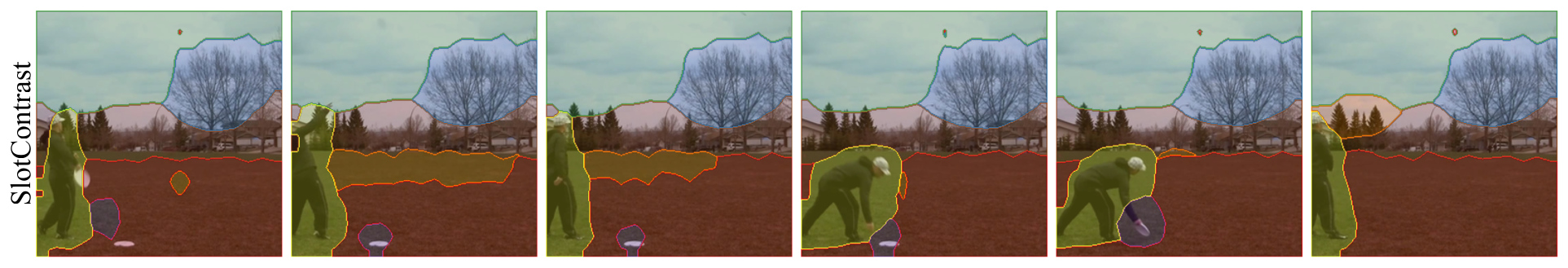}\\ \vspace{-0.2cm}
    \includegraphics[width=1\linewidth]{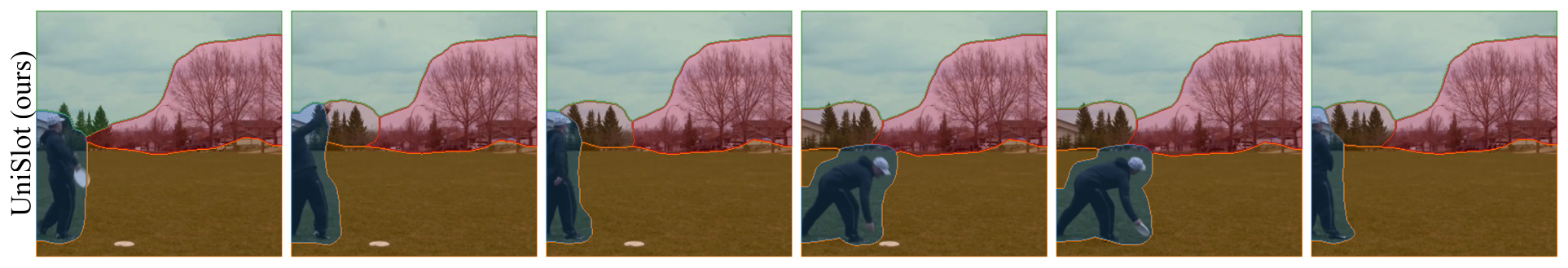}\\
    \label{fig:extra_qualitative_ytvis2021_part2}
\end{figure}

\begin{figure}[h]
    \centering
    \caption{\textcolor{black}{Additional qualitative results on MOVi-E. Colors indicate slot assignments and are kept consistent across frames to visualize temporal object tracking.}}
    \includegraphics[width=1\linewidth]{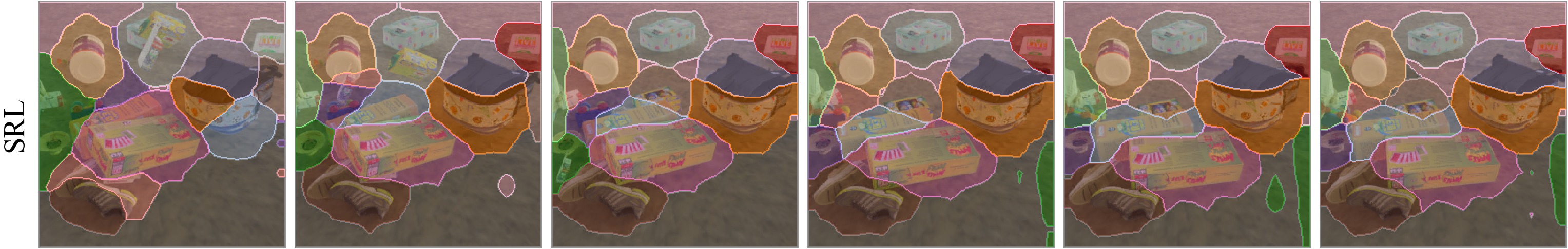}\\ \vspace{-0.2cm}
    \includegraphics[width=1\linewidth]{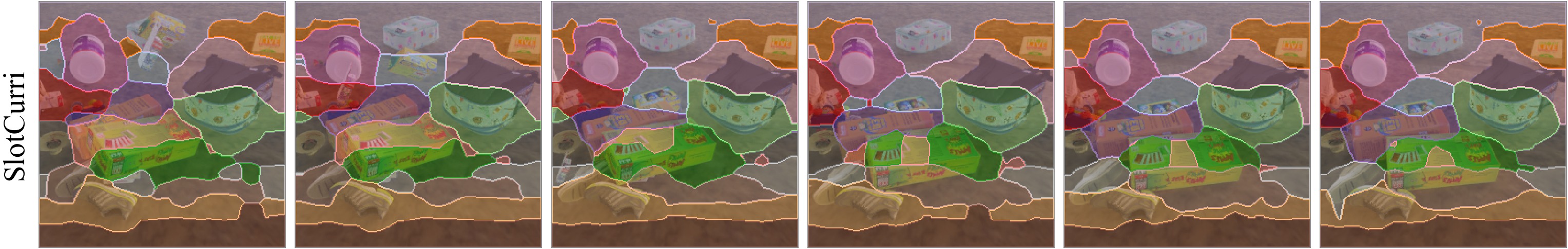}\\ \vspace{-0.2cm}
    \includegraphics[width=1\linewidth]{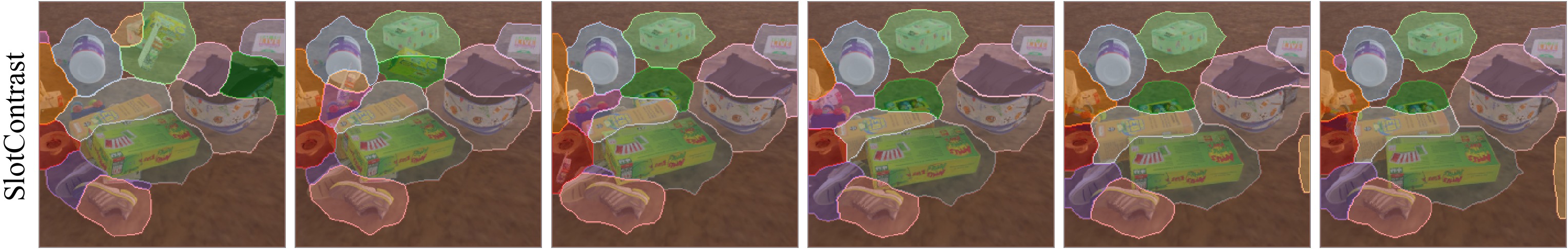}\\ \vspace{-0.2cm}
    \includegraphics[width=1\linewidth]{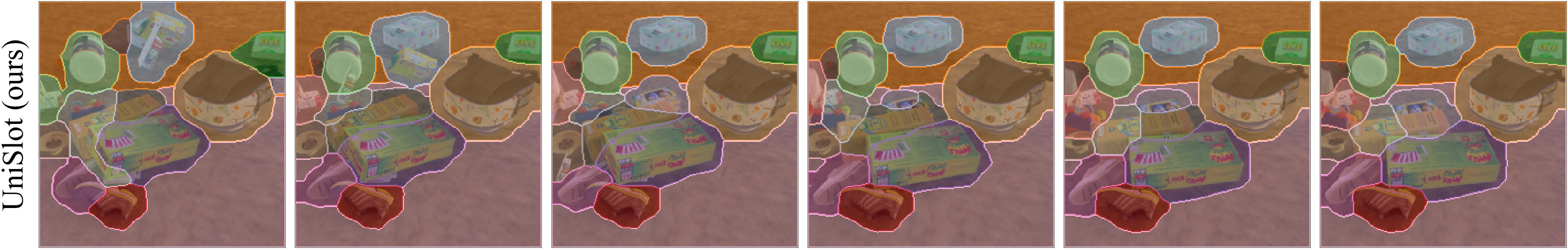}\\ \vspace{0.4cm}
    \includegraphics[width=1\linewidth]{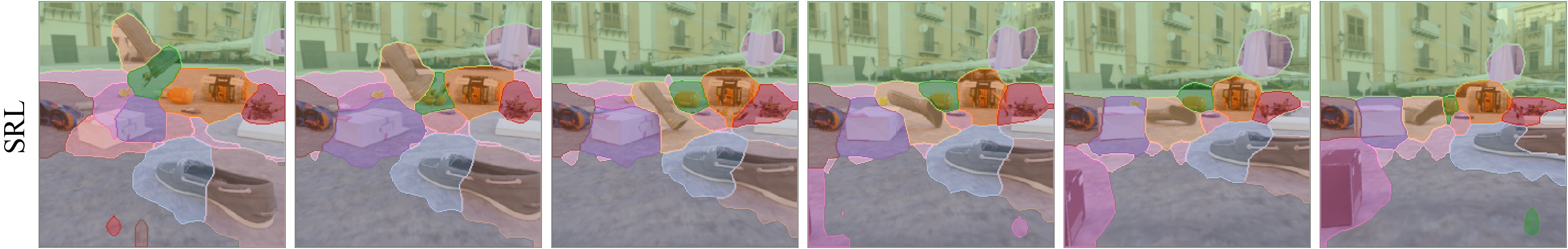}\\ \vspace{-0.2cm}
    \includegraphics[width=1\linewidth]{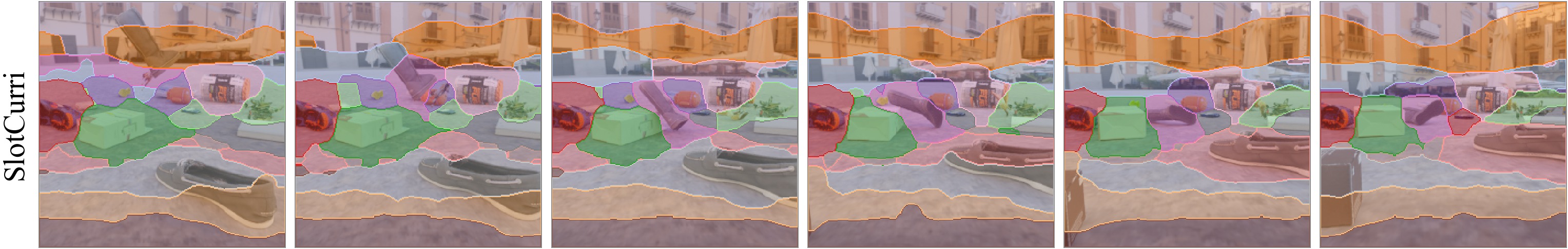}\\ \vspace{-0.2cm}
    \includegraphics[width=1\linewidth]{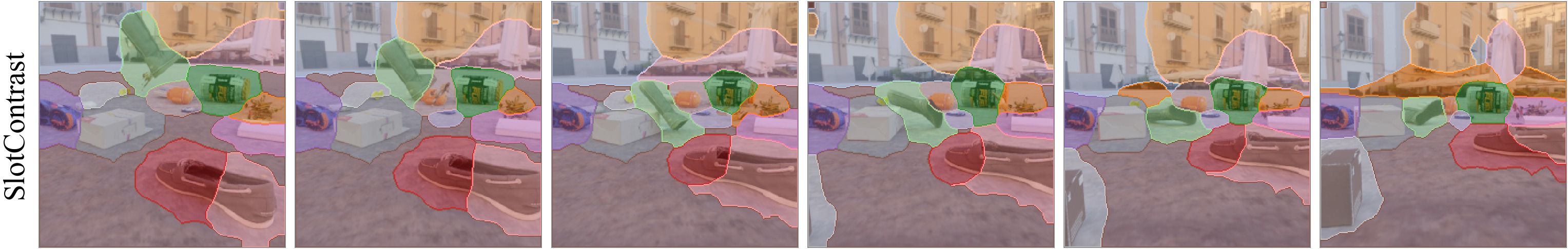}\\ \vspace{-0.2cm}
    \includegraphics[width=1\linewidth]{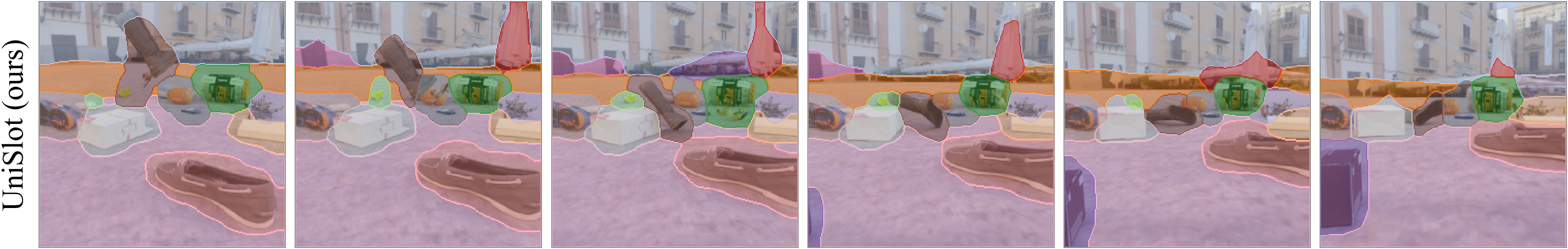}\\
    \label{fig:extra_qualitative_movie}
\end{figure}

\begin{figure}[h]
    \centering
    \caption{Additional qualitative results on MOVi-C. Colors indicate slot assignments and are kept consistent across frames to visualize temporal object tracking.}
    \includegraphics[width=1\linewidth]{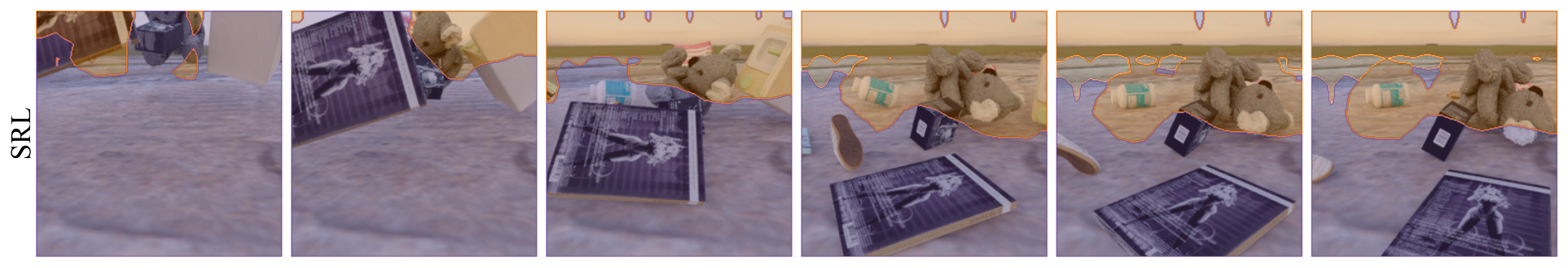}\\ \vspace{-0.2cm}
    \includegraphics[width=1\linewidth]{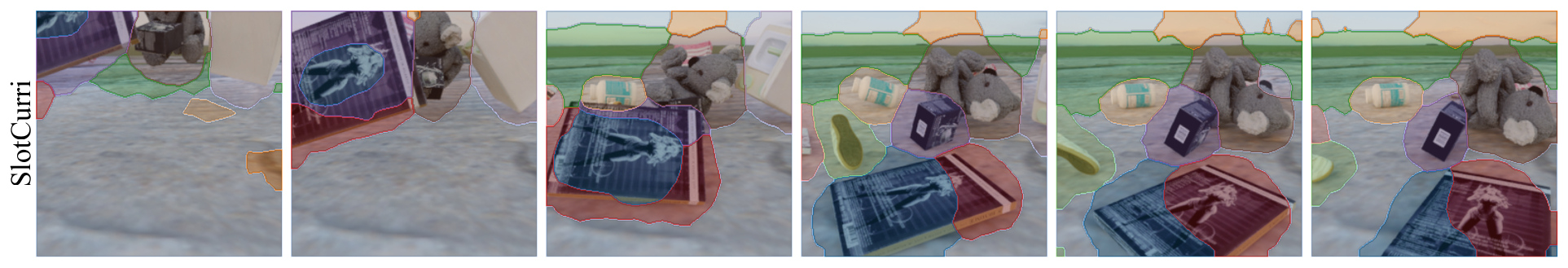}\\ \vspace{-0.2cm}
    \includegraphics[width=1\linewidth]{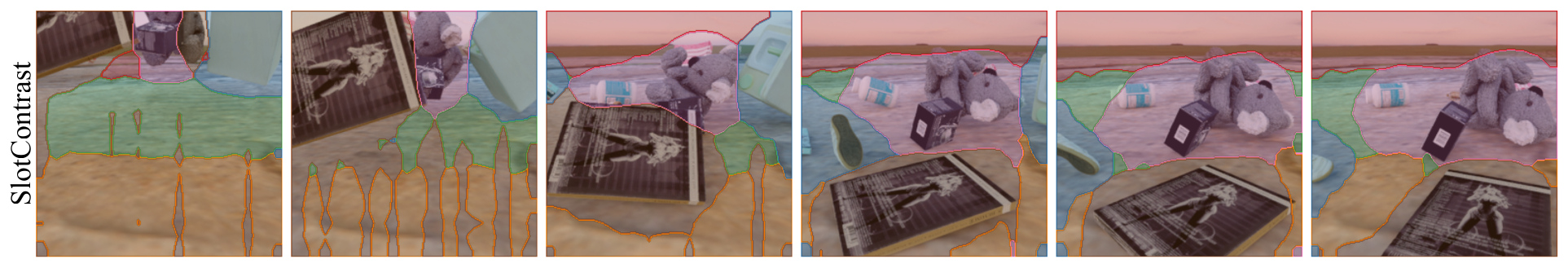}\\ \vspace{-0.2cm}
    \includegraphics[width=1\linewidth]{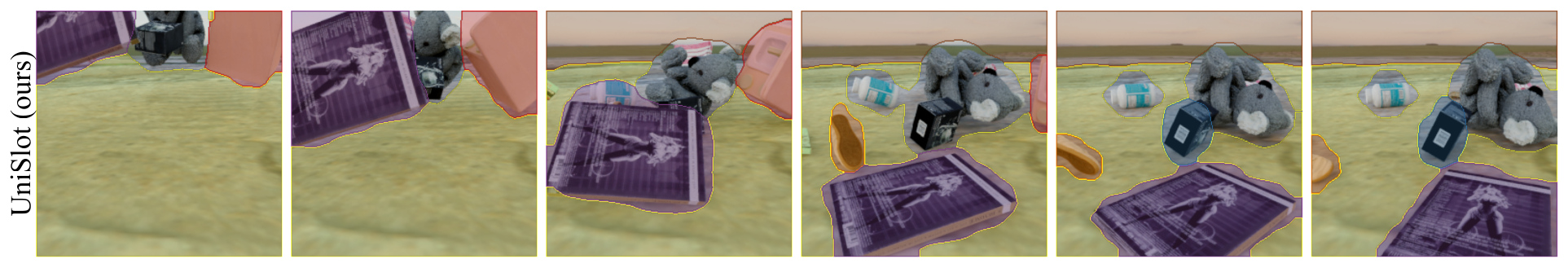}\\ \vspace{0.4cm}
    \includegraphics[width=1\linewidth]{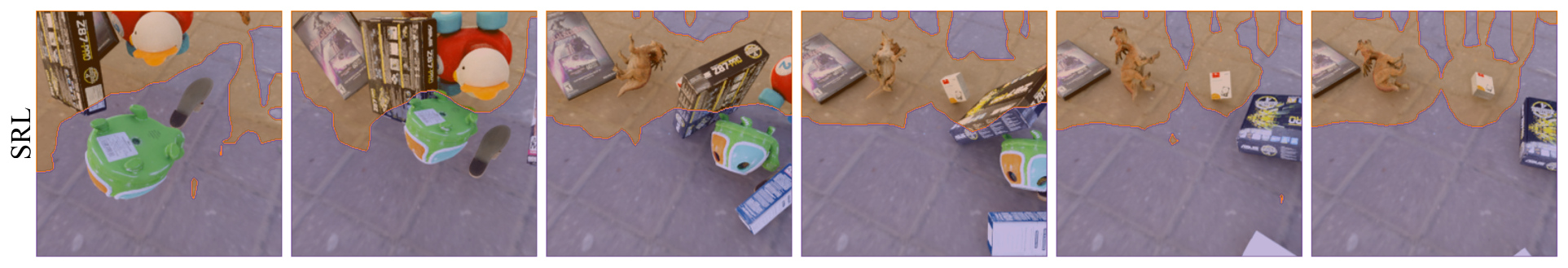}\\ \vspace{-0.2cm}
    \includegraphics[width=1\linewidth]{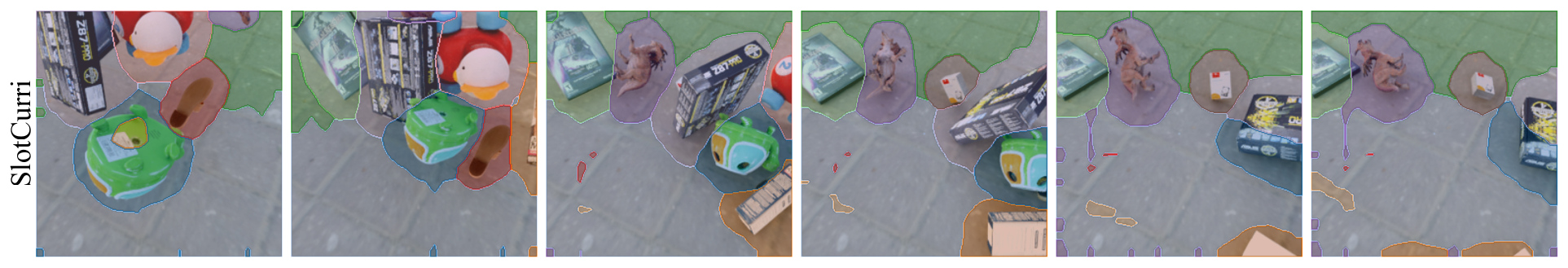}\\ \vspace{-0.2cm}
    \includegraphics[width=1\linewidth]{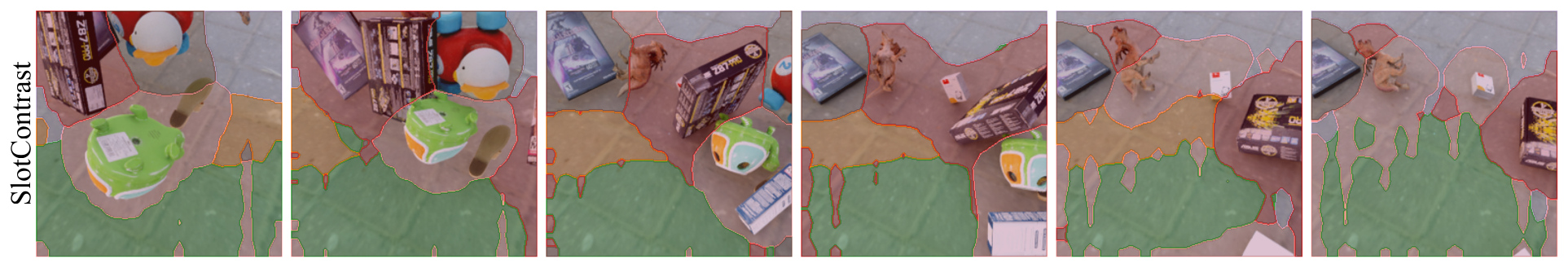}\\ \vspace{-0.2cm}
    \includegraphics[width=1\linewidth]{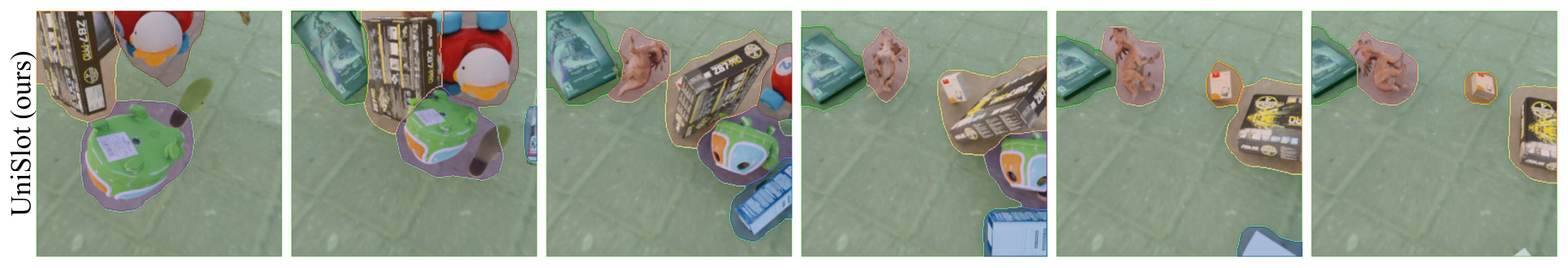}\\
    \label{fig:extra_qualitative_movic}
\end{figure}

\begin{figure}[h]
    \centering
    \caption{Additional qualitative results on CLEVRER. Colors indicate slot assignments and are kept consistent across frames to visualize temporal object tracking.}
    \includegraphics[width=1\linewidth]{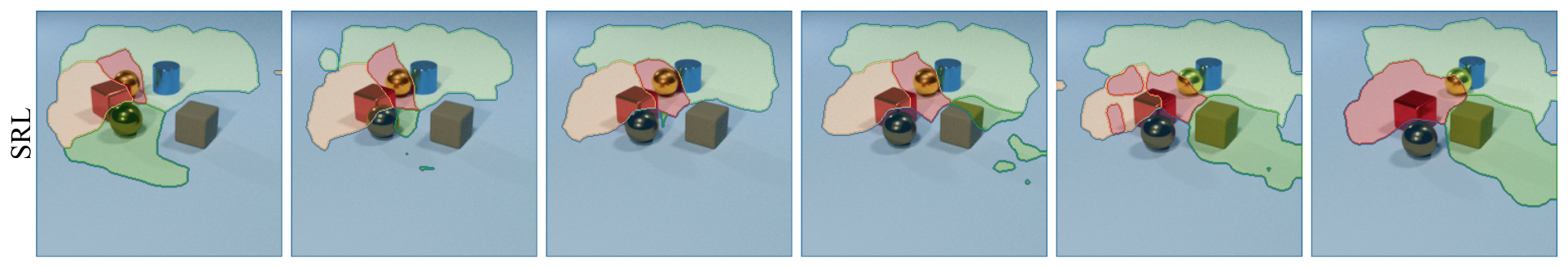}\\ \vspace{-0.2cm}
    \includegraphics[width=1\linewidth]{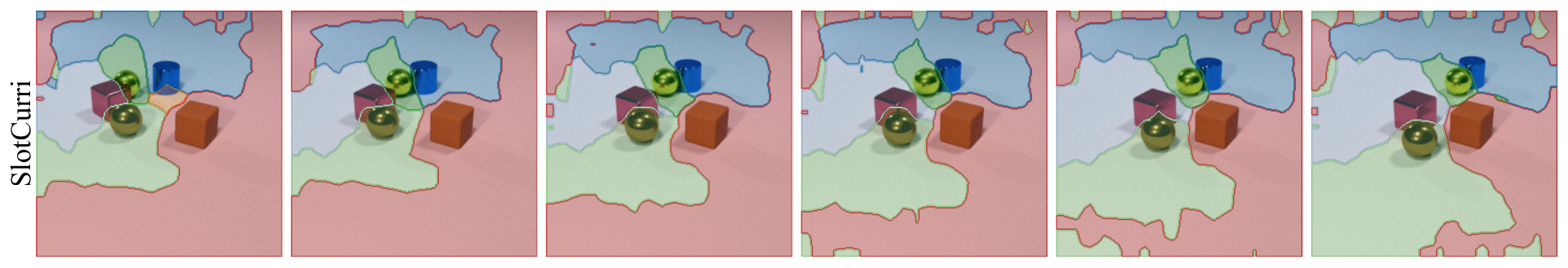}\\ \vspace{-0.2cm}
    \includegraphics[width=1\linewidth]{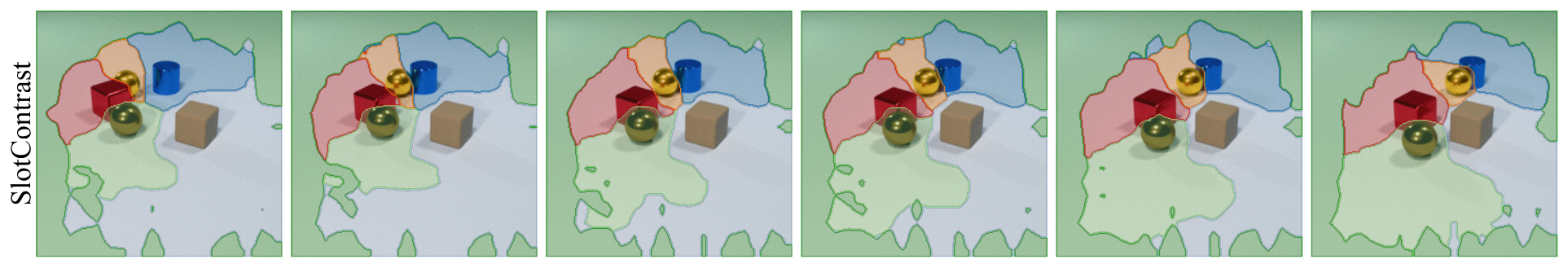}\\ \vspace{-0.2cm}
    \includegraphics[width=1\linewidth]{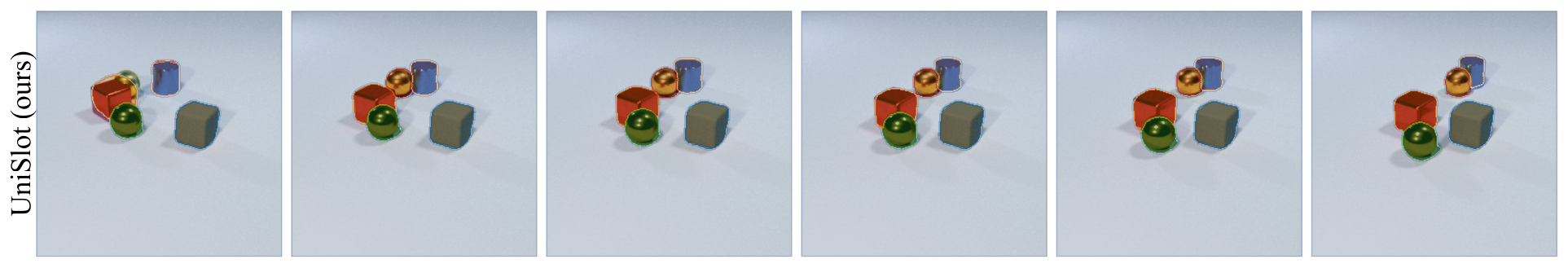}\\ \vspace{0.4cm}
    \includegraphics[width=1\linewidth]{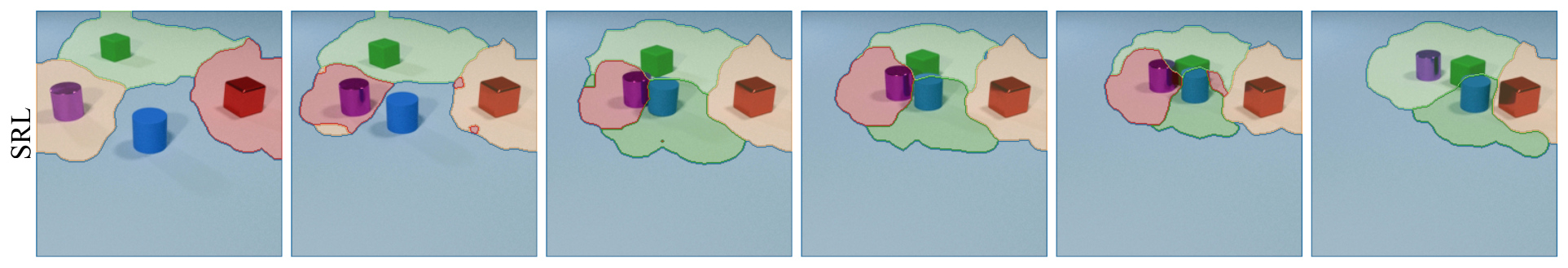}\\ \vspace{-0.2cm}
    \includegraphics[width=1\linewidth]{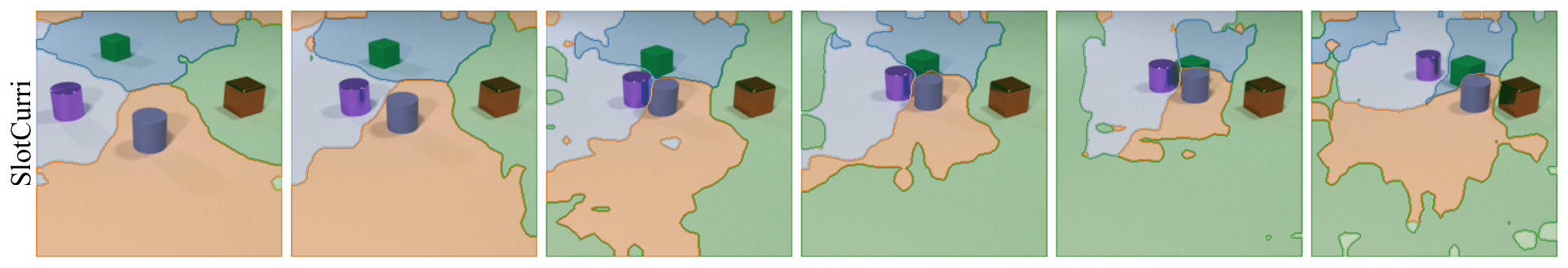}\\ \vspace{-0.2cm}
    \includegraphics[width=1\linewidth]{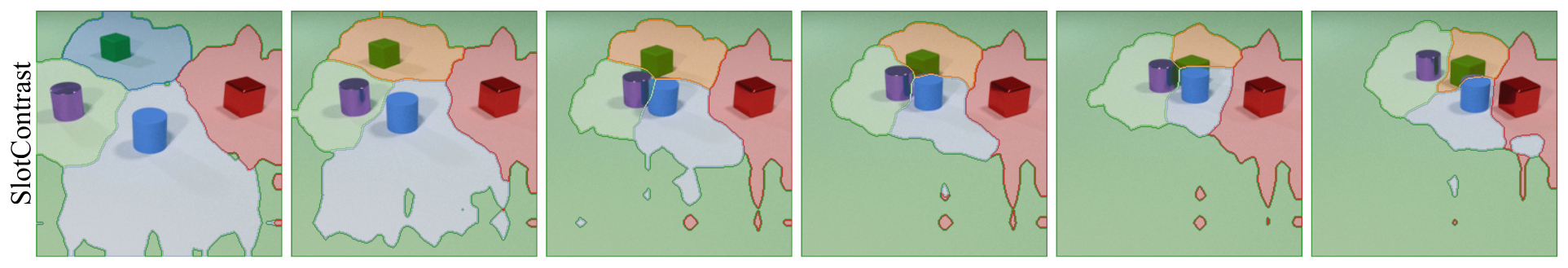}\\ \vspace{-0.2cm}
    \includegraphics[width=1\linewidth]{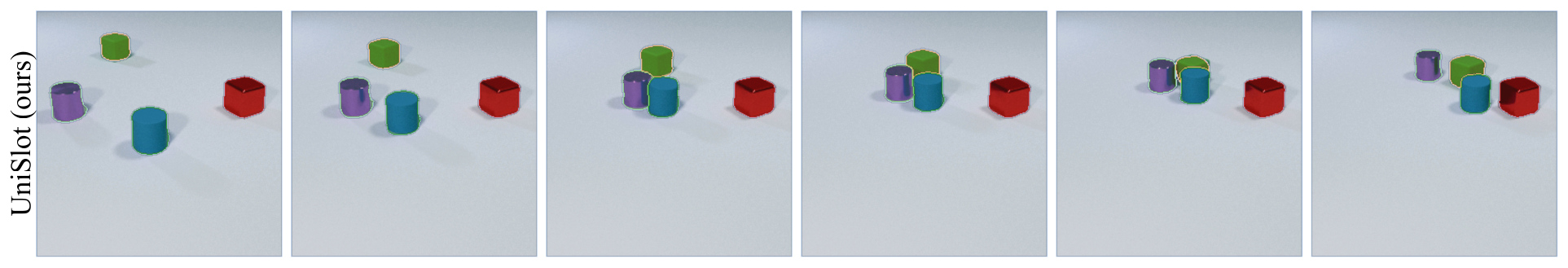}\\
    \label{fig:extra_qualitative_clevrer}
\end{figure}

\end{document}